\documentclass{article}

\usepackage{arxiv}
\usepackage[numbers]{natbib}

\usepackage[utf8]{inputenc} 
\usepackage[T1]{fontenc}    
\usepackage{hyperref}       
\usepackage{url}            
\usepackage{booktabs}       
\usepackage{amsfonts}       
\usepackage{nicefrac}       
\usepackage{microtype}      
\usepackage{lipsum}
\usepackage{graphicx}
\usepackage{caption}
\makeatletter
\def\blfootnote{\gdef\@thefnmark{}\@footnotetext}
\makeatother

\theoremstyle{plain}
\newtheorem{theorem}{Theorem}[section]

\newtheorem{lemma}[theorem]{Lemma}

\theoremstyle{definition}
\newtheorem{definition}[theorem]{Definition}

\theoremstyle{remark}

\title{Generalizing DP-SGD with Shuffling and Batch Clipping}

\author{Marten van Dijk$^{1,2,3}$, 
\textbf{Phuong Ha Nguyen}$^{4}$, Toan N. Nguyen$^{5\dagger}$, \\ \textbf{Lam M. Nguyen}$^{6}$\textbf{,}   \\
$^{1}$ CWI Amsterdam, The Netherlands\\
$^{2}$ Department of Computer Science, Vrije Universiteit Amsterdam, The Netherlands \\
$^{3}$
Department of Electrical and Computer Engineering, University of Connecticut, CT, USA\\
$^{4}$ eBay, CA, USA\\
$^{5}$ Department of Computer Science and Engineering, University of Connecticut, CT, USA \\
$^{6}$ IBM Research, Thomas J. Watson Research Center, Yorktown Heights, NY, USA\\
\\
\texttt{marten.van.dijk@cwi.nl}, \texttt{toan.nguyen@uconn.edu},  \\
\texttt{LamNguyen.MLTD@ibm.com}, \texttt{phuongha.ntu@gmail.com}}

\begin{document}
\maketitle

\blfootnote{$^{\dagger}$ Supported by NSF grant CNS-1413996 “MACS: A Modular
Approach to Cloud Security.”}
\begin{abstract}
Classical differential private DP-SGD implements individual clipping with random subsampling, which forces a mini-batch SGD approach. We provide a general differential private algorithmic framework that goes beyond DP-SGD and allows any possible first order optimizers (e.g., classical SGD and momentum based SGD approaches) in combination with batch clipping, which clips an aggregate of computed gradients rather than summing clipped gradients (as is done in individual clipping). The framework also admits sampling techniques beyond random subsampling such as shuffling. Our DP analysis follows the $f$-DP approach and introduces a new proof technique 
which allows us to derive simple closed form expressions and to also analyse group privacy. In particular, for $E$ epochs work and groups of size $g$, we show a $\sqrt{g E}$ DP dependency for batch clipping with shuffling. 
\end{abstract}

\textcolor{blue}{{\bf Disclaimer:} The main contribution is the new proof technique mentioned in the abstract. As explained in Appendix \ref{sec:adv}, it turns out that we use a stronger adversarial model in our DP analysis compared to the adversarial model used in current literature on the  moment accountant method, $f$-DP, and divergence based DP measures. This leads to weaker DP guarantees because we analyze the stronger adversary who has more capabilities. As a consequence we lose to a significant extent the privacy amplification due to subsampling. Even though the resulting DP guarantees seem promising, they are significantly weaker -- and, in practice, we do not see the very strong adversary we assume in our proofs (in fact, the adversarial model in current literature seems already too strong in practical adversarial scenarios). Even though the new math techniques in this paper are correct, they do not lead to a result that improves over existing work. At best this paper can serve as an example of how careful analysis of hidden assumptions in (existing) proofs can lead to unexpected surprises.}



\section{Introduction}


In order to defend against privacy leakage of local proprietary data during collaborative training of a global model,
\citep{abadi2016deep} introduced  DP-SGD  as it adapts 
Stochastic Gradient Descent (SGD)
with Differential Privacy (DP)\citep{dwork2006calibrating, dwork2011firm,dwork2014algorithmic,dwork2006our}.
This approach allows each client to perform local SGD computations for the data samples $\xi$ that belong to the client's local training data set $d$. 
%
Each client is doing SGD computations for a batch of local data. These recursions together represent a local round and at the end of the local round a local model update (in  the form of an aggregate of computed gradients during the round) is transmitted to the server. The server in turn adds the received local update to its global model -- and once the server receives new updates from (a significant portion of) all  clients, the global model is broadcast to each of the clients. When considering privacy, we are concerned about how much information these  local updates reveal about the used local data sets. 

Depicted in Algorithm \ref{alg:genDPSGD} (a detailed description and explanation is given in Section \ref{sec:alg}), we introduce a general algorithmic framework of which DP-SGD is one example. The main generalization beyond DP-SGD is that the most inner loop can execute any optimization algorithm ${\cal A}$ in order to compute a local round update $U$. 
DP-SGD is realized for $s=1$ with ${\cal A}$ computing a single gradient resulting in (line 21)
\begin{equation}
    U = \sum_{i\in S_b} [ \nabla f(w,\xi_{i})]_C, \label{indclip}
\end{equation}
where $[x]_C= x/\max\{1,\|x\|/C\}$ is the clipping operation and $S_b$ is a batch sampled from  local data set $d$ of size $m$.

We call (\ref{indclip}) the {\em individual clipping} approach and  notice that it implements mini-batch SGD. This is necessary because the DP analysis requires each clipped output to be independent from other clipped outputs. This implies that computation of one clipped output should not involve updating a local model $w$ which is used in a computation of another next clipped output. This means that each of the clipped gradients in the sum are evaluated in the same $w$ (the most recently received global model from the server). 
We notice that the clipping operation introduces `clipping' noise and together with the added Gaussian noise for differential privacy this results in convergence to a final global model with smaller (test) accuracy (than what otherwise, without DP, can be achieved).

Our algorithmic framework allows a much wider choice. In particular,  $m=1$ 
allows for example
\begin{equation}
U = [\sum_{j=1}^s \nabla f(w_j,\xi_{i_j})]_C, \label{batchclip}
\end{equation}
where $S_b=\{i_1,\ldots, i_s\}$ (lines 14, 15) is a batch sampled from the local data set $d$ of size $s$ and where algorithm ${\cal A}$ implements classical SGD according to the SGD recursion
$$ w_{j+1}=w_j-\eta \nabla f(w_j,\xi_{i_j}),$$
where $w_1=w$ is initialized by the most recently received global model from the server (possibly updated with updates send to the central server, see lines 8 and 24 in Algorithm \ref{alg:genDPSGD}) and where $\eta=\eta_{(e-1)\frac{N}{ms}+b}$ is the round step size (for round $b$ in epoch $e$). 

We call (\ref{batchclip})  an example of {\em batch clipping}. Batch clipping in its most general form is
\begin{equation}
    U= [{\cal A}(w,\{\xi_i\}_{i\in S_{b}})]_C.
    \label{batchclipA}
\end{equation}
It allows us to implement classical SGD
and go beyond mini-batch SGD such as  Adam~\cite{kingma2014adam}, AdaGrad~\cite{duchi2011adaptive}, SGD-Momentum \cite{sutskever2013importance}, or RMSProp \cite{zou2018sufficient}in a classical SGD approach without mini-batch. The framework is general in that ${\cal A}$ can implement any optimization algorithm including these
momentum based SGD type algorithms.
Literature 
 and the state-of-the-art $f$-DP framework \citep{dong2021gaussian} only proves DP guarantees for SGD or momentum based SGD with individual clipping and a mini-batch approach as in (\ref{indclip}).


%
%
%
%
%
%

\begin{algorithm}[t]
\caption{
Generalized Framework for DP-SGD}
\label{alg:genDPSGD}
\begin{algorithmic}[1]
\Procedure{DP-SGD-General}{}
    \State $N=$ size training data set $d=\{\xi_i\}_{i=1}^{N}$
    \State $E=$ total number of epochs
    \State $T=$ total number of rounds
    \State diminishing round step size sequence $\{\eta_i\}_{i=1}^{T}$
    \State 
    \State initialize $w$ as the default initial model
    \State {\bf Interrupt Service Routine (ISR)}:  Whenever a new global model $\hat{w}$ is received, computation is interrupted and an ISR is called that replaces $w\leftarrow \hat{w}$ after which computation is resumed
    \State
    \For{$e\in \{1,\ldots, E\}$}
    \State Let $\pi^e$ be a random permutation 
    \State re-index data samples: $\{\xi_i\leftarrow \xi_{\pi^e(i)}\}_{i=1}^N$  
    \State $\{S_{b,h}\}_{b=1,h=1}^{N/(ms),m}\leftarrow$ ${\tt Sample}_{s,m}$ with \State \hspace{0.35cm} $S_{b,h}\subseteq \{1,\ldots, N\}$, 
    \State \hspace{0.35cm} $|S_{b,h}|=s$, $|S_b|=sm$ with $S_b=\bigcup_{h=1}^m S_{b,h}$ 
    \For{$b\in \{1,\ldots, \frac{N}{ms}\}$}
 \State Start of round $(e-1)\frac{N}{ms}+b$:
    \For{$h\in \{1,\ldots m\}$}
    \State $a_h\leftarrow {\cal A}(w,\{\xi_i\}_{i\in S_{b,h}})$
    \EndFor
    \State $U=\sum_{h=1}^m [a_h]_C$
    \State $\bar{U}\leftarrow U+ {\cal N}(0,(2C\sigma)^2{\bf I})$
    \State Transmit $\bar{U}/m$ to central server
    \State Locally update $w\leftarrow w- \eta_{(e-1)\frac{N}{ms}+b} \cdot \bar{U}/m$
    \EndFor
    \EndFor
\EndProcedure
\end{algorithmic}
\end{algorithm}

We have the following main contributions: 
%

\noindent
{\bf General Algorithmic Framework:} Algorithm \ref{alg:genDPSGD} with detailed discussion  in Section \ref{sec:alg} defines our general framework. In the inner loop it allows execution of any optimization algorithm ${\cal A}$ of our choice. This covers classical SGD and momentum based SGD (as discussed above), and can possibly be extended with batch normalization (over $S_b$). Our framework is compatible with asynchronous communication between clients and server. We also notice that the framework allows diminishing step sizes (learning rate), and allows adapting the clipping constant from round to round (this is compatible with DP analysis in general).


\noindent {\bf DP Proof Technique based on a Sampling Induced Distribution:} 
We are the {\em first} to provide a DP analysis of the much wider class of learning algorithms covered by the general algorithmic framework of Algorithm \ref{alg:genDPSGD}.
We introduce a probability distribution $q_E(c)$ induced by the sampling procedure ${\sf Sample}_{s,m}$ and 
we prove a new $f$-DP guarantee  where $f$ is related to a mix of Gaussian trade-off functions $G_{c/\sigma}$ according to distribution $q_E(c)$. 

\noindent {\bf Derivation of DP Guarantees for Group Privacy with Square Root Dependency:} Table \ref{tab:DP} summarizes our results 
as a consequence of applying the theory in  Section \ref{sec:DPanalysis}
(the  proofs and applications of our theory for the tabled cases are in Appendices \ref{sec:subsampling} and \ref{sec:shuffling}). Table \ref{tab:DP} focusses on (general) batch clipping as this allows algorithms beyond SGD with individual clipping, subsampling and a min-batch SGD style approach as in DP-SGD.
Our DP analysis shows that a much wider class of SGD based algorithms have provable DP guarantees. 
We see that in $f$-DP terminology various configurations in the general algorithmic framework provide a $\approx G_{\sqrt{gE}/\sigma}$-DP guarantee with a $\sqrt{gE}$ dependency, where we consider group privacy for groups of size $g$, $E$ is the total number of epochs (measured in data set size $N$) worth of gradient computations, and $\sigma$ characterizes  the added Gaussian DP noise (after normalization with the clipping constant). 

We have a $\sqrt{g}$ dependency for group privacy for all $g\geq 1$ if we use batch clipping and shuffling as a sampling strategy (see Section \ref{sec:alg}). We  implemented batch clipping of a mini-batch according to
\begin{equation}
    U= [\frac{1}{s} \sum_{j=1}^{s}  \nabla f(w,\xi_{i_j})]_C
\label{eq:implementedBC}
\end{equation}
with $S_b=\{i_1,\ldots, i_s\}$ and each gradient evaluated in the same $w$ as an example of batch clipping in (\ref{batchclipA}). Together with shuffling and setting $\sigma=2$ this turns out to achieve about 71.5\% accuracy for  for CIFAR-10 and 98.3\% accuracy for MNIST. This compares to 71.1\% accuracy for CIFAR-10 and 98.4\% accuracy for MNIST for individual clipping with subsampling as in DP-SGD. Which in turn is similar to the accuracies if just mini-batch SGD without DP is implemented. Simulation details are in Appendix \ref{sec:07experiments} (due to space limitation). (For future work we leave improving the accuracy by implementing more advanced learning algorithms that fit the general algorithmic framework.) We conclude that the implemented batch clipping does not degrade accuracy and we conclude that batch clipping with shuffling is a viable candidate with the advantage that we have $\sqrt{g}$ dependency for all $g\geq 1$ for group privacy.

We notice that the shown $\sqrt{g}$ dependency cannot be concluded in a straightforward way from existing results in literature (for various DP measures) that transform individual privacy into a general statement for group privacy. Existing transformations always result in a linear dependency in $g$. This is discussed in related work in Section \ref{sec:rel}.

\begin{table*}[t]
    \centering
{\small 
    \begin{tabular}{|cc|cc|c|c|}
    \hline 
    \parbox[t]{2mm}{\rotatebox[origin=c]{90}{\hspace{1mm} batch clip. (\ref{batchclip}), (\ref{batchclipA}),
    (\ref{eq:implementedBC}) \hspace{1mm}}} &
    \parbox[t]{2mm}{\rotatebox[origin=c]{90}{\hspace{1mm} gen. clipping (\ref{eq:genA})
    \hspace{1mm}}} &
    \parbox[t]{2mm}{\rotatebox[origin=c]{90}{\hspace{1mm} subsampling \hspace{1mm}}} &
    \parbox[t]{2mm}{\rotatebox[origin=c]{90}{\hspace{1mm} shuffling \hspace{1mm}}} &
    \parbox[t]{2mm}{\rotatebox[origin=c]{90}{\hspace{1mm} group privacy \hspace{1mm}}} &
    $h$-DP guarantee 
     \\
    \hline
    \hline
        \vspace{-3mm}   &  &  &  &  &   \\
      x & & x & & $g\geq 1$ & $h$ is in the range $\approx \left[G_{\sqrt{(1+1/\sqrt{2gE})gE}/\sigma}, G_{\sqrt{(1-1/\sqrt{2gE})gE}/\sigma}\right]$,  \\
     &&&&& approximation is for small $e^{-gE}$ and small $s/N$  \\
     &&&&& (see Thm \ref{thm:shuffling} for a tighter lower bound corresponding to Def \ref{def:f}) \\
         \vspace{-2mm}   &  &  &  &  &   \\
      & x & x & & $g=1$ & $h$ is in the range $\approx \left[G_{\sqrt{(1+1/\sqrt{2E})E}/\sigma}, G_{\sqrt{(1-1/\sqrt{2E})E}/\sigma}\right]$, \\
     &&&&& approximation is for small $e^{-E}$  \\
         \vspace{-2mm}   &  &  &  &  &   \\
      & x & & x & $g=1$ & $h=G_{\sqrt{E}/\sigma}$  \\
         \vspace{-2mm}   &  &  &  &  &   \\
      & x & & x & $g\geq 1$ & $h$ is in the range $\approx [G_{\sqrt{gE}/\sigma}, G_0]$, \\    &&&&&    
         approximation is for constant $E$ and  small $g^2ms/(N-g)$ \\
  \vspace{-2mm}   &  &  &  &  &    \\        
      x &  & & x & $g\geq 1$ & $h \in \left[G_{\sqrt{gE}/\sigma}, G_0\right]$  \\
         \vspace{-2mm}   &  &  &  &  &  \\
      x &  & & x & $g\leq s$ & $h \approx G_{\sqrt{gE}/\sigma}$,  \\
     &&&&& approximation is for constant $E$ and
     small $g^2/(N/s-g-g^2)$  \\
    \hline
    \end{tabular}
}
    \caption{
    Trade-off functions $h$ for the mechanism ${\cal M}$ defined by Algorithm \ref{alg:genDPSGD} 
    for all pairs of data sets $d$ and $d'$ 
with $\max \{ |d\setminus d'|, |d'\setminus d|\}\leq g$; 
$h$ cannot be improved beyond the reported ranges closer towards function $G_0(\alpha)= 1-\alpha$ (which represents random guessing of the hypothesis $d$ or $d'$, hence, no privacy leakage). An approximation for a small quantity means that if the quantity tends to $0$, then the approximation becomes tight.} 
    \label{tab:DP}
\end{table*}

\noindent
{\bf Outline:} Section \ref{sec:rel} provides related work on group privacy. Section \ref{sec:alg} defines our general algorithmic framework. The minimal necessary background on $f$-DP is in Section \ref{sec:GDPmain} (with a full description in Appendix \ref{sec:GDP}).
Section \ref{sec:DPanalysis} states the main definitions and main theorems (proved in Appendices \ref{sec:lemmas} and \ref{sec:DPproof}, applied in Appendices  \ref{sec:subsampling} and \ref{sec:shuffling} resulting in Table \ref{tab:DP}, and with additional  discussion 
in Appendices \ref{app:factor} and \ref{sec:adv}).  

\section{Related Work}
\label{sec:rel}

If a mechanism is $(\epsilon,\delta)$-DP, then the literature shows $(g\epsilon,g e^{g-1}\delta)$-DP for groups of size $g$ \citep{DRV10}. Here, we see an exponential dependency in $g$ due to the $g e^{g-1}$ term in the privacy failure probability.
In order to have a better dependency on $g$ and improved adaptive composibility, the notion of Concentrated Differential Privacy (CDP)  \citep{CDP} was introduced.
CDP was re-interpreted and relaxed by using Renyi entropy in \citep{BS15} and its authors followed up with the notion of zero-CDP (zCDP) in \citep{zCDP}. This notion admits
simple interpretable DP guarantees for adaptive composition and group privacy. In particular, we have the general transformation: If a mechanism is $\rho$-zCDP,
then it is $g^2\rho$-zCDP for groups of size $g$.

After the introduction of $\rho$-zCDP, Renyi DP (RDP) was introduced by \citep{RDP}; if a mechanism is $(\omega,\tau)$-RDP,
then it is 
$(\omega/g,g^{\ln 3 /\ln 2}\tau)$-RDP for groups of size $g$.
%
%
%
Combining the ideas that give rise to the zCDP and RDP definitions
leads naturally to the definition of $(\rho,\omega)$-tCDP \citep{tCDP} which relaxes zCDP. We have the general transformation:
$(\rho,\omega)$-tCDP 
implies $(\rho g^2,\omega/g)$-tCDP  for groups of size $g$.

The general transformations for group privacy 
for the divergence based DP measures 
are all tight in that no stronger statement can be given that holds for all mechanisms.
We focus on DP-SGD (just one specific mechanism) and just for DP-SGD we  derive a significantly better dependency on $g$ for group privacy: 
Appendix B in \citep{dong2021gaussian} shows how to infer divergence based DP guarantees from $f$-DP. In particular, $G_{\sqrt{gE}/\sigma}$-DP in Table \ref{tab:DP} implies $(\omega,\frac{gE}{2\sigma^2}\cdot \omega)$-RDP (Renyi differential privacy) for any $\omega> 1$, and this implies
$$\frac{gE}{2\sigma^2}\mbox{-zCDP} \mbox{ and } (\frac{gE}{2\sigma^2}\cdot \omega, \omega)\mbox{-tCDP}.$$
We conclude that we achieve a factor $g$ improvement.

For the $f$-DP framework, \citep{dong2021gaussian} shows the general transformation: If a mechanism is $G_\mu$-DP, then it is $G_{g\mu}$-DP for groups of size $g$. Application of this transformation to our result for individual privacy leads to a linear dependency in $g$ as opposed to the $\sqrt{g}$ dependency proven here.

\section{Algorithmic Framework for Differential Private SGD} \label{sec:alg}


We provide a general algorithmic framework for differential private SGD in Algorithm \ref{alg:genDPSGD}: 




\noindent 
{\bf Asynchronous SGD:} We allow asynchronous communication between clients and the central aggregating server. Each client maintains its own local model. Each client implements an Interrupt Service Routine (ISR) which replaces its local model $w$ with any newly received
global model $\hat{w}$ from the server; the ISR interrupts the computation in lines 10-26 as soon as a new global model is received and resumes computation after updating the local model. In practice asynchronous communication may lead to out of order arrival or dropped global models broadcast by the central server. A limited amount of asynchronous behavior 
leads to provably optimal convergence rates for certain learning tasks \cite{van2020hogwild}.



\noindent 
{\bf Sampling Strategies:}
The whole training process spans $E$ epochs. 
At the start of each epoch ${\tt Sample}_{s,m}$  generates sets $\{S_{b,h}\subseteq \{1,\ldots, N\}\}_{b=1,h=1}^{N/(ms),m}$ where each set $S_{b,h}$ has size $s$; we define $\{S_b=\bigcup_{h=1}^m S_{b,h}\}_{b=1}^{N/(ms)}$ where each set $S_b$ has size $ms$. Here, $b$ indexes the round within an epoch during which the client uses its training data to compute a local update that is transmitted to the server. After multiplying with the appropriate learning rates (step sizes $\eta_i$), the server aggregates received local updates from different clients into the global model maintained by the server. Within round $b$, $h$ indexes independent computations by an algorithm ${\cal A}$ (discussed later). Each computation starts from the client's local model. The $h$-th computation  is based on training data samples from data set $d$ that corresponds to $S_{b,h}$.  

In this paper we analyze and compare two distinct sampling strategies for ${\tt Sample}_{s,m}$:
\begin{itemize}
    \item {\bf Subsampling (SS):} Sets $S_b\subseteq \{1,\ldots, N\}$ with $|S_b|=ms$ are randomly sampled from $\{1,\ldots, N\}$. 
    \item {\bf Shuffling (SH):} The whole index set $\{1,\ldots, N\}$ is shuffled into a random order (according to a random chosen permutation). This is used to  partition data set $d$ into sets $S_b\subseteq \{1,\ldots, N\}$ with $|S_b|=ms$ in a random way. Notice that $d=\bigcup_{b=1}^{N/(ms)} S_b$.
\end{itemize}
The difference between the two strategies is that the $S_b$ are disjoint for shuffling while they may have intersections for subsampling. Once the $S_b$ are sampled,
${\tt Sample}_{s,m}$ selects a random chosen partition in order to split $S_b$ into subsets $S_{b,h}$, $1\leq h\leq m$, with $|S_{b,h}|=s$.




\noindent 
{\bf Update Algorithm:}
A round computes some partial update $a_h\leftarrow {\cal A}(w,\{\xi_i\}_{i\in S_{b,h}})$, 
which only depends on $w$ and the set of training data samples $S_{b,h}$.
Here, $w$ is either equal to the most recent received global model or has been locally updated using some step size in combination with a local noised aggregated update $\bar{U}$ that was also transmitted to the central server. This means that any observer of communication with the central server is also able to compute the used $w$ based on previously observed updates $\bar{U}$. 

Algorithm ${\cal A}$ considers in sequence each of the $s$ training samples in $\{\xi_{i_1},\ldots, \xi_{i_s}\}$, where 
$S_{b,h}=\{i_1,\ldots, i_s\}$ is the random index set produced by ${\tt Sample}_{s,m}$. 
%
%
We notice that our algorithmic framework and DP analysis allow any other optimization algorithm such as momentum based SGD. It may include batch normalization where normalization is done over the  batch $\{\xi_{i_1},\ldots, \xi_{i_s}\}$, and it may also include instance or layer normalization. 
We may even predefine a sequence of algorithms $\{{\cal A}_b\}_{b=1}^{N/(ms)}$ per epoch and use ${\cal A}_b$ in round $b$.

\noindent 
{\bf Clipping:}
To each computed  $a_h$ we apply clipping $x\rightarrow [x]_C=x/\max \{1, \|x\|/C\}$. We aggregate the clipped $[a_h]_C$ in a sum $U$. We add Gaussian noise before it is transmitted to the central server. Each round reveals one noisy update $U$ and we want to bound the aggregated differential privacy leakage. 
%
%
%
%
%
The general formula for $U$ is
\begin{equation}
    U=\sum_{h=1}^m [a_h]_C = \sum_{h=1}^m [{\cal A}(w,\{\xi_i\}_{i\in S_{b,h}})]_C.
    \label{eq:genA}
\end{equation}

\noindent
{\bf Gaussian Noise:} After $U$ is computed according to (\ref{eq:genA}), Gaussian noise ${\cal N}(0,(2C\sigma)^2)$ is added to each entry of vector $U$. The resulting noisy update $\bar{U}$ after averaging by $m$ is sent to the central server.

\noindent
{\bf Some Remarks:} We notice that the framework allows a diminishing step size sequence (learning rate). The DP analysis shows that we may adapt the clipping constant at the start of each round. Rather than computing $U$ as the sum $\sum_{h=1}^m [a_h]_C$, we may compute $\sum_{h=1}^m {\cal B}([a_h]_{C'})$ for some post-processing function/procedure ${\cal B}$ (here, we need to take care that $\|{\cal B}(x)\|\leq C$ for $\|x\|\leq C'$). By revealing a differential private noisy mean and noisy variance of the training data set (due to differential private data normalization pre-processing), algorithm ${\cal A}$ can implement data normalization; in the $f$-DP framework, privacy leakage is now characterized as a trade-off function of the differential private data normalization pre-processing  composed with the trade-off function corresponding to our DP analysis for Algorithm \ref{alg:genDPSGD}. 
Without loss of differential privacy, we may under certain circumstances (by changing sampling to batches $S_b$ with non-fixed probabilistic sizes) use Gaussian noise ${\cal N}(0,(C\sigma)^2{\bf I})$, a factor $2$ less; a discussion is in Appendix \ref{app:factor}.


\section{Background $f$-DP} \label{sec:GDPmain}

Dong et al. \citep{dong2021gaussian} introduced the state-of-the-art DP formulation based on hypothesis testing. From the attacker's perspective, it is natural to formulate the  problem of distinguishing two 
neighboring 
data sets $d$ and $d'$ based on the output of a DP mechanism ${\cal M}$ as a hypothesis testing problem:
$$\mbox{versus } \begin{array}{l}
H_0: \mbox{ the underlying data set is }d \\
H_1: \mbox{ the underlying data set is }d'.
\end{array}
$$
Here, neighboring means that either $|d\setminus d'|=1$ or $|d'\setminus d|=1$. 
More precisely, in the context of mechanism ${\cal M}$, ${\cal M}(d)$ and ${\cal M}(d')$ take as input representations $r$ and $r'$ of data sets $d$ and $d'$ which are `neighbors.' The representations are mappings from a set of indices to data samples with the property that if $r(i)\in d\cap d'$ or $r'(i)\in d\cap d'$, then $r(i)=r'(i)$. This means that the mapping from indices to data samples in $d\cap d'$ is the same for the representation of $d$ and the representation of $d'$. In other words the mapping from indices to data samples for $d$ and $d'$ only differ for indices corresponding to the {\em differentiating} data samples in $(d\setminus d')\cup (d'\setminus d)$. In this sense the two mappings (data set representations) are neighbors.
In our main theorem we will consider the general case $g=\max\{ |d\setminus d'|, |d'\setminus d|\}$ in order to analyse `group privacy.' 

We define the Type I and Type II errors by
$$\alpha_\phi = \mathbb{E}_{o\sim {\cal M}(d)}[\phi(o)]  \mbox{ and } \beta_\phi = 1- \mathbb{E}_{o\sim {\cal M}(d')}[\phi(o)],
$$
where $\phi$ in $[0,1]$ denotes the rejection rule which takes the output of the DP mechanism as input. We flip a coin and reject the null hypothesis with probability $\phi$. The optimal trade-off between Type I and Type II errors is given by the trade-off function
$$ T({\cal M}(d),{\cal M}(d'))(\alpha) = \inf_\phi \{ \beta_\phi \ : \ \alpha_\phi \leq \alpha \},$$ 
for $\alpha \in [0,1]$, where the infimum is taken over all measurable rejection rules $\phi$. If the two hypothesis are fully indistinguishable, then this leads to the trade-off function $1-\alpha$. We say a function $f\in [0,1]\rightarrow [0,1]$ is a trade-off function if and only if it is convex, continuous, non-increasing, at least $0$, and $f(x)\leq 1-x$ for $x\in [0,1]$. 

We define a mechanism ${\cal M}$ to be $f$-DP if $f$ is a trade-off function and
$$
 T({\cal M}(d),{\cal M}(d')) \geq f
$$
for all neighboring $d$ and $d'$.
The $f$-DP framework supersedes all existing other frameworks in that a trade-off function contains all the information needed to derive known DP metrics such as $(\delta,\epsilon)$-DP and divergence based DPs. 



%
\citep{dong2021gaussian} defines Gaussian DP as a special case of $f$-DP where $f$ is a trade-off function
$$G_\mu(\alpha) = T({\cal N}(0,1),{\cal N}(\mu,1))(\alpha) = \Phi( \Phi^{-1}(1-\alpha) - \mu )$$
with $\Phi$ the standard normal cumulative distribution of ${\cal N}(0,1)$. 

The tensor product $f\otimes h$ for trade-off functions $f=T(P,Q)$ and $h=T(P',Q')$ is well-defined by 
$$f\otimes h = T(P\times P',Q\times Q').$$
Let $y_i \leftarrow {\cal M}_i(\texttt{aux},d)$ with $\texttt{aux}=(y_1,\ldots, y_{i-1})$. Theorem 3.2 in \citep{dong2021gaussian} shows that if ${\cal M}_i(\texttt{aux},.)$ is $f_i$-DP for all $\texttt{aux}$, then the composed mechanism ${\cal M}$, which applies ${\cal M}_i$ in sequential order from $i=1$ to $i=T$, is 
$(f_1\otimes \ldots \otimes f_T)$-DP. The tensor product is commutative.
%
As a special case Corollary 3.3 in \citep{dong2021gaussian} states that composition of multiple Gaussian operators $G_{\mu_i}$ results in $G_{\mu}$ where $\mu=\sqrt{\sum_i \mu_i^2}$.


%

Suppose that $d$ and $d'$ do not differ in just one sample, but differ in $g$ samples. \citep{dong2021gaussian} shows that if a mechanism is $G_\mu$-DP, then it is $G_{g\mu}$-DP for groups of size $g$ (the reverse implication is not true in general). 

A full description of $f$-DP (including the subsampling operator $C_{m/N}$ and operator $\circ g$ in Table \ref{tab:DP}) with intuitive explanation is in Appendix \ref{sec:GDP}.

%

\section{DP Analysis}
\label{sec:DPanalysis}



Our main theorems, stated below, are proved in Appendices \ref{sec:lemmas} and \ref{sec:DPproof}. We consider two data sets $d$ and $d'$ that differ in $g$ samples in order to also analyze group privacy. 

The main idea is to bound the sensitivity\footnote{The sensitivity of a value is the Euclidean norm between its evaluation for $d$ and $d'$ respectively.} of $\sum_{h=1}^m [a_h]_C$ in Algorithm \ref{alg:genDPSGD} by $2kC$ where $k$ is equal to the number of $a_h$ whose computation depends on differentiating samples (samples outside $d\cap d'$). The number of rounds $b$ within an epoch that have $k$ values $a_h$ that depend on differentiating samples is denoted by $c_k$. An epoch instance is into some extent characterized by the vector $(c_1,c_2,\ldots,c_g)$ where $c_k$ indicates the number of rounds that have sensitivity equal to $2kC$.

If a round has a sensitivity $2kC$, then the trade-off function of its corresponding mechanism, after adding Gaussian noise ${\cal N}(0,(2C\sigma)^2{\bf I})$, is given by the Gaussian trade-off function
$G_{k/\sigma}$. Therefore, an epoch instance $(c_1,c_2,\ldots,c_g)$ has a Gaussian trade-off function which is the composition of the round trade-off functions $G_{k/\sigma}$: We have
$$ 
G_{1/\sigma}^{\otimes c_1} \otimes G_{2/\sigma}^{\otimes c_2} \otimes \ldots \otimes G_{g/\sigma}^{\otimes c_g}= G_{\sqrt{\sum_{k=1}^g c_k k^2}/\sigma}.
$$
This can also be composed over multiple epochs.
This leads to defining a probability distribution $q_E(c)$ which is the probability that $E$ epoch instances together realize the value $c=\sqrt{\sum_{k=1}^g c_k k^2}$. 
With probability $q_E(c)$, mechanism ${\cal M}$ executes a `sub-mechanism' which has trade-off function $G_{c/\sigma}$. The trade-off function of the overall mechanism ${\cal M}$ is therefore related, but not exactly equal, to the expectation $\sum_c q_E(c)\cdot G_{c/\sigma}$: See Lemma \ref{lem:mech1} in Appendix \ref{sec:lemmas}, we have
$T({\cal M}(d),{\cal M}(d'))\geq f$,  where $f(\alpha)$ equals the trade-off function
\begin{equation}  
\inf_{\{\alpha_c\}} \left\{ \left. \sum_c q_E(c) G_{c/\sigma}(\alpha_c)  \ \right| \ \sum_i q_E(c) \alpha_c=\alpha \right\}. \label{eq:fsum}
\end{equation}

Random variable $c_k$ counts the number of rounds within the $E$ epochs that have sensitivity $2kC$. The vector of random variables $(c_1,\ldots,c_g)$ defines the random variable $c=\sqrt{\sum_{k=1}^g c_k k^2}$. By analyzing its probability distribution $q_E(c)$ we are able to derive upper and lower bounds for $f$ defined in (\ref{eq:fsum}). The theorem provides a solution $f(\alpha)$ for the infinum and provides lower and upper bound functions. 
%
%
%
%
%
This leads to $T({\cal M}(d),{\cal M}(d') \geq f$ 
with upper and lower bounds which, see Table \ref{tab:DP}, can be very close to one another.

We first define/formalize a couple of concepts before stating our main theorems (proofs are in Appendices \ref{sec:lemmas} and \ref{sec:DPproof}).


\begin{definition} \label{def:qEc}
Let ${\cal M}$ be the mechanism  corresponding to our general framework for $E$ epochs which is based on ${\sf Sample}_{s,m}$ with $N$ equal to the size of the data set that is sampled. 
We require
that if ${\tt Sample}_{s,m}$ outputs $\{S^e_{b,h}\}_{b=1,h=1}^{N/(sm),m}$ for the $e$-th epoch, then for each $b$, $\{S^e_{b,h}\}_{h=1}^{m}$ partitions a set of size $ms$ into $m$ subsets $S^e_{b,h}$ of size $s$.
Define
$$ {\cal C} = \left\{ \sqrt{\sum_{k=1}^g c_k k^2} \ : \ \forall_k \ c_k\in \mathbb{N} \right\}.$$
The sampling procedure ${\tt Sample}_{s,m}$ defines a probability distribution $q_E(c)$ over ${\cal C}$ as follows (notice that $q_E(c)$ implicitly depends on $N$):
\begin{eqnarray*}
&& q_E(c) = 
{\tt Pr}_{\{\pi^e\}}\left[ 
\begin{array}{c}
c^2=\sum_{k=1}^g 
c_kk^2 \\
\mbox{with } c_k=\# \{ (b,e) : L_{b,e}=k\} \mbox{ and}\\
 L_{b,e} = |\{ h \ : \ \pi^e(S^e_{b,h}) \cap \{1,\ldots, g\} \neq \emptyset \}| \\
\mbox{ conditioned on } \\
\{ \{S^e_{b,h}\}_{b=1,h=1}^{N/(sm),m} \leftarrow {\tt Sample}_{s,m} \}_{e=1}^E 
\end{array}
\right].
\end{eqnarray*}
\end{definition}

The next definitions define three trade-off functions based on $q_E(c)$ (which, as a consequence, also depend on $N$). The proof of the main theorems in Appendix \ref{sec:DPproof} show that the functions are well-defined.

\begin{definition} \label{def:f}
For distribution $q_E(c)$ over $c\in {\cal C}$ we define for $\alpha\in [0,1]$ function 
$$ f(\alpha)=\sum_{c\in {\cal C}} q_E(c) \cdot \Phi\left(\Lambda(\alpha)\cdot \frac{\sigma}{c} - \frac{c}{2\sigma}\right),$$
where function $\Lambda(\alpha)$, $\alpha\in [0,1]$, is implicitly defined by
    $$1-\alpha = \sum_{c\in {\cal C}} q_E(c) \cdot \Phi\left(\Lambda(\alpha)\cdot \frac{\sigma}{c} + \frac{c}{2\sigma}\right).$$
\end{definition}

\begin{definition}
Let $q_E(c)$ be a distribution over $c\in {\cal C}$. Let $u_{c^{\tt U}}$ and $c^{\tt U}$ be such that
$$
1\geq u_{c^{\tt U}} \geq \sum_{c\in {\cal C}: c<c^{\tt U}} q_E(c).
$$
Notice that $u_{c^{\tt U}}$ bounds the tail of $q_E(c)$ left from $c^{\tt U}$.

For $\alpha\in [0,1]$ we define  functions 
\begin{eqnarray*} \hat{f}^{\tt U}_{c^{\tt U}}(\alpha) &=& u_{c^{\tt U}}+(1-u_{c^{\tt U}})G_{c^{\tt U}/\sigma}(\alpha), \\
f^{\tt U}_{c^{\tt U}}(\alpha) &=& \min \{ \hat{f}^{\tt U}_{c^{\tt U}}, \hat{f}^{\tt U}_{c^{\tt U}}{}^{-1} \}^{**}(\alpha) 
=
\left\{
\begin{array}{ll}
u_{c^{\tt U}}+(1-u_{c^{\tt U}})G_{c^{\tt U}/\sigma}(\alpha),     &  \alpha \in [0,\beta_0], \\
\beta_0+\beta_1-\alpha,     &   \alpha\in [\beta_0,\beta_1], \\
G_{c^{\tt U}/\sigma}(\frac{\alpha -u_{c^{\tt U}} }{1-u_{c^{\tt U}}}),   &   \alpha \in [\beta_1,1],
\end{array}
\right.
\end{eqnarray*}
where
\begin{eqnarray*}
\beta_0 &=&
1- \Phi( \frac{c^{\tt U}}{2\sigma}-\frac{\sigma}{c^{\tt U}} \ln(1-u_{c^{\tt U}})), \\
\beta_1 &=& u_{c^{\tt U}} + (1-u_{c^{\tt U}})\cdot (1- \Phi(\frac{c^{\tt U}}{2\sigma}+\frac{\sigma}{c^{\tt U}} \ln(1-u_{c^{\tt U}}))).
\end{eqnarray*}
\end{definition}

\begin{definition}
Let $q_E(c)$ be a distribution over $c\in {\cal C}$. Let $l_{c^{\tt L}}$ and $c^{\tt L}$ be such that
$$
1\geq l_{c^{\tt L}} \geq \sum_{c\in {\cal C}: c>c^{\tt L}} q_E(c).
$$
Notice that $l_{c^{\tt L}}$ bounds the tail of $q_E(c)$ right from $c^{\tt L}$.

For $\alpha\in [0,1]$ we define  functions 
\begin{eqnarray*} \hat{f}^{\tt L}_{c^{\tt L}}(\alpha) &=& G_{c^{\tt L}/\sigma}(\min \{1,\alpha+l_{c^{\tt L}}\}), \\
f^{\tt L}_{c^{\tt L}}(\alpha) &=& \max \{ \hat{f}^{\tt L}_{c^{\tt L}}, \hat{f}^{\tt L}_{c^{\tt L}}{}^{-1} \}(\alpha) 
=
\left\{
\begin{array}{ll}
G_{c^{\tt L}/\sigma}(\alpha)-l_{c^{\tt L}},     &  \alpha\in [0,\beta],\\
G_{c^{\tt L}/\sigma}(\min \{1,\alpha+l_{c^{\tt L}}\}),     &  \alpha \in [\beta,1],
\end{array}
\right.
\end{eqnarray*}
where
$\beta\in [0,1]$ is implicitly defined as the (unique) solution of 
$$ G_{c^{\tt L}/\sigma}(\beta)-l_{c^{\tt L}} =\beta
$$
(and we notice that $\hat{f}^{\tt L}_{c^{\tt L}}{}^{-1}(\beta) = \beta = \hat{f}^{\tt L}_{c^{\tt L}}(\beta)$).
\end{definition}

\begin{theorem} \label{thm:main}
%
%
If ${\cal M}$ is $h$-DP for all pairs of data sets $d$ and $d'$ 
with $\max \{ |d\setminus d'|, |d'\setminus d|\}\leq g$ and $N=|d\cap d'|+g$,
then 
$$ h\leq f^{\tt U}_{c^{\tt U}} \leq \hat{f}^{\tt U}_{c^{\tt U}}.$$
We notice that $f^{\tt U}_{c^{\tt U}}$ is a symmetric trade-off function.
\end{theorem}

\begin{theorem} \label{thm:main2}
${\cal M}$ is $f$-DP for all pairs of data sets $d$ and $d'$ 
with $\max \{ |d\setminus d'|, |d'\setminus d|\}\leq g$ and $N=|d\cap d'|+g$. Trade-off function $f$ has  lower bound
$$ f\geq f^{\tt L}_{c^{\tt L}}\geq  \hat{f}^{\tt L}_{c^{\tt L}} .$$
Both $f$ and $f^{\tt L}_{c^{\tt L}}$ are symmetric trade-off functions.  For  larger $g$, $f$ becomes smaller (for all $\alpha$).  
%
%
\end{theorem}


Notice that the adversary gets better in hypothesis testing when $f$ becomes smaller for larger $g$.
Therefore, in the worst case, the total number $t=|d\setminus d'|+|d'\setminus d|$ of samples in which $d$ and $d'$ differ are distributed as $|d\setminus d'|=t$ with $|d'\setminus d|=0$ (or vice versa) as this achieves the maximum value for $g$, i.e., $g=t$.








The difficulty of applying Theorems \ref{thm:main} and \ref{thm:main2} is in computing the probability distribution $q_E(c)$ -- the number of possible $c$ may be exponential in $g$ and $E$. Therefore, computing $f$ may be prohibited and only the lower and upper bound functions can possibly be estimated. Here, we remark that in general the lower and upper bounds are not tight together for small $E$; only for larger $E$ the probability distribution $q_E(c)$ will concentrate and   lower and upper bounds exist that  approach one another for large $E$. 




In Appendices \ref{sec:subsampling} and \ref{sec:shuffling} we apply Theorems  \ref{thm:main} and \ref{thm:main2} to the DP analysis of subsampling and shuffling.  Table \ref{tab:DP} in the introduction summarizes the derived bounds\footnote{The table shows upper and lower bounds 
that are independent of $N$ or generally hold for large $N$ (as found in practice). Hence, the DP guarantees for ${\cal M}$ hold for all pairs of data sets $d$ and $d'$ 
with $\max \{ |d\setminus d'|, |d'\setminus d|\}\leq g$ (for the $g$ stated in the table). The condition $N=|d\cap d'|+g$ of the theorem can be discarded.}. 
We notice that shuffling appears to concentrate $q_E(c)$ more compared to subsampling.

\section{Conclusion}
\label{sec:08conclusion}

 

We have introduced a general algorithmic framework for  SGD based learning algorithms with DP guarantees (this includes momentum based classical SGD with batch clipping and shuffling). Our DP guarantees show a non-trivial $\sqrt{g}$ dependency for group privacy. 
\bibliography{references}
\bibliographystyle{plainnat}

\clearpage
\appendix
\onecolumn

\section{Gaussian Differential Privacy}
\label{sec:GDP}

Dong et al. \citep{dong2021gaussian} introduced the state-of-the-art DP formulation based on hypothesis testing. From the attacker's perspective, it is natural to formulate the  problem of distinguishing two 
neighboring 
data sets $d$ and $d'$ based on the output of a DP mechanism ${\cal M}$ as a hypothesis testing problem:
$$H_0: \mbox{ the underlying data set is }d \ \ \ \ \mbox{ versus } \ \ \ \  H_1: \mbox{ the underlying data set is }d' .$$
Here, neighboring means that either $|d\setminus d'|=1$ or $|d'\setminus d|=1$. 
More precisely, in the context of mechanism ${\cal M}$, ${\cal M}(d)$ and ${\cal M}(d')$ take as input representations $r$ and $r'$ of data sets $d$ and $d'$ which are `neighbors.' The representations are mappings from a set of indices to data samples with the property that if $r(i)\in d\cap d'$ or $r'(i)\in d\cap d'$, then $r(i)=r'(i)$. This means that the mapping from indices to data samples in $d\cap d'$ is the same for the representation of $d$ and the representation of $d'$. In other words the mapping from indices to data samples for $d$ and $d'$ only differ for indices corresponding to the differentiating data samples in $(d\setminus d')\cup (d'\setminus d)$. In this sense the two mappings (data set representations) are neighbors.

We define the Type I and Type II errors by
$$\alpha_\phi = \mathbb{E}_{o\sim {\cal M}(d)}[\phi(o)]  \mbox{ and } \beta_\phi = 1- \mathbb{E}_{o\sim {\cal M}(d')}[\phi(o)],
$$
where $\phi$ in $[0,1]$ denotes the rejection rule which takes the output of the DP mechanism as input. We flip a coin and reject the null hypothesis with probability $\phi$. The optimal trade-off between Type I and Type II errors is given by the trade-off function
$$ T({\cal M}(d),{\cal M}(d'))(\alpha) = \inf_\phi \{ \beta_\phi \ : \ \alpha_\phi \leq \alpha \},$$ 
for $\alpha \in [0,1]$, where the infimum is taken over all measurable rejection rules $\phi$. If the two hypotheses are fully indistinguishable, then this leads to the trade-off function $1-\alpha$. We say a function $f\in [0,1]\rightarrow [0,1]$ is a trade-off function if and only if it is convex, continuous, non-increasing, 
and $0\leq f(x)\leq 1-x$ for $x\in [0,1]$. 

We define a mechanism ${\cal M}$ to be $f$-DP if $f$ is a trade-off function and
$$
 T({\cal M}(d),{\cal M}(d')) \geq f
$$
for all neighboring $d$ and $d'$.
Proposition 2.5 in \citep{dong2021gaussian} is an adaptation of a result in \citep{wasserman2010statistical} and states that a mechanism is $(\epsilon,\delta)$-DP if and only if the mechanism is $f_{\epsilon,\delta}$-DP, where
$$f_{\epsilon,\delta}(\alpha) =
\min \{ 0, 1-\delta - e^{\epsilon}\alpha, (1-\delta-\alpha)e^{-\epsilon}\}.
$$
We see that $f$-DP has the $(\epsilon,\delta)$-DP  formulation as a special case. It turns out that the original DP-SGD algorithm can be tightly analysed by using $f$-DP.

\subsection{Gaussian DP}
\label{sec:sens}

%
In order to proceed, \citep{dong2021gaussian} first defines Gaussian DP as another special case of $f$-DP as follows: We define the trade-off function
$$G_\mu(\alpha) = T({\cal N}(0,1),{\cal N}(\mu,1))(\alpha) = \Phi( \Phi^{-1}(1-\alpha) - \mu ),$$
where $\Phi$ is the standard normal cumulative distribution of ${\cal N}(0,1)$. We define a mechanism to be $\mu$-Gaussian DP if it is $G_\mu$-DP. Corollary 2.13 in \citep{dong2021gaussian} shows  that a mechanism is $\mu$-Gaussian DP if and only if it is $(\epsilon, \delta(\epsilon))$-DP for all $\epsilon\geq 0$, where
\begin{equation} \delta(\epsilon) = \Phi(-\frac{\epsilon}{\mu}+\frac{\mu}{2}) - e^{\epsilon} \Phi(-\frac{\epsilon}{\mu}-\frac{\mu}{2}).
\label{eq:gdp}
\end{equation}

Suppose that a mechanism ${\cal M}(d)$ computes some function $u(d)\in \mathbb{R}^n$ and adds Gaussian noise ${\cal N}(0,(c\sigma)^2{\bf I})$, that is,  the mechanism outputs $o\sim u(d)+{\cal N}(0,(c\sigma)^2{\bf I})$. Suppose that $c$ denotes the sensitivity of function $u(\cdot)$, that is, $$\|u(d)-u(d')\|\leq c$$ 
for neighboring $d$ and $d'$; the mechanism corresponding to one round update in Algorithm \ref{alg:genDPSGD} has {\em sensitivity} $c=2C$. After projecting the observed $o$  onto the line that connects $u(d)$ and $u(d')$ and after normalizing by dividing by $c$, we have that differentiating whether $o$ corresponds to $d$ or $d'$ is in the best case for the adversary (i.e., $\|u(d)-u(d')\|=c$) equivalent to differentiating whether a received output is from ${\cal N}(0,\sigma^2)$ or from  ${\cal N}(1,\sigma^2)$. Or, equivalently, from ${\cal N}(0,1)$ or from  ${\cal N}(1/\sigma,1)$. 
This is how the Gaussian trade-off function $G_{\sigma^{-1}}$ comes into the picture. 

\subsection{Subsampling} \label{sec:sample}

%
Besides implementing Gaussian noise,   DP-SGD also uses sub-sampling: For a data set $d$ of $N$ samples, ${\tt Sample}_m(d)$   selects a subset of size $m$ from $d$ uniformly at random. We define convex combinations
$$ f_p(\alpha) = p f(\alpha) + (1-p) (1-\alpha)$$
with corresponding $p$-sampling operator 
$$ C_p(f) = \min \{ f_p, f_p^{-1} \}^{**},
$$
where the conjugate $h^*$ of a function $h$ is defined as
$$ h^*(y) = \sup_x \{ yx -h(x) \}$$ 
and the inverse $h^{-1}$ of a trade-off function $h$ is defined as
\begin{equation}
h^{-1}(\alpha) = \inf\{t\in[0,1] \ | \ h(t)\leq \alpha \} \label{inverse}
\end{equation}
and is itself a trade-off function (as an example, we notice that $G_\mu=G_\mu^{-1}$ and we say $G_\mu$ is symmetric).
Theorem 4.2 in \citep{dong2021gaussian} shows that if a mechanism ${\cal M}$ on data sets of size $N$ is $f$-DP, then the subsampled mechanism ${\cal M}\circ {\tt Sample}_{m}$ is $C_{m/N}(f)$-DP.


The intuition behind operator $C_p$ is as follows. 
First, ${\tt Sample}_m(d)$ samples the differentiating element between $d$ and $d'$ with probability $p$. In this case the computations ${\cal M}\circ {\tt Sample}_{m}(d)$ and  ${\cal M}\circ {\tt Sample}_{m}(d')$ are different and hypothesis testing is possible with trade-off function $f(\alpha)$. With probability $1-p$ no hypothesis testing is possible and we have trade-off function $1-\alpha$. This leads to the convex combination $f_p$. 

Second, we notice if $h=T({\cal M}(d),{\cal M}(d'))$, then $h^{-1}=T({\cal M}(d'),{\cal M}(d))$. Therefore, if ${\cal M}$ is $f$-DP (which holds for all pairs of neighboring data sets, in particular, for the pairs $(d,d')$ and $(d',d)$), then both $h\geq f$ and $h^{-1}\geq f$ and we have a symmetric upper bound $\min\{h,h^{-1}\}\geq f$. Since $f$ is a trade-off function, $f$ is convex and we can compute a tighter upper bound: $f$ is at most the largest convex function $\leq \min\{h,h^{-1}\}$, which is equal to the double conjugate $\min\{h,h^{-1}\}^{**}$. From this we obtain the definition of operator $C_p$.



\subsection{Composition}

%
The tensor product $f\otimes h$ for trade-off functions $f=T(P,Q)$ and $h=T(P',Q')$ is well-defined by 
$$f\otimes h = T(P\times P',Q\times Q').$$
Let $y_i \leftarrow {\cal M}_i(\texttt{aux},d)$ with $\texttt{aux}=(y_1,\ldots, y_{i-1})$. Theorem 3.2 in \citep{dong2021gaussian} shows that if ${\cal M}_i(\texttt{aux},.)$ is $f_i$-DP for all $\texttt{aux}$, then the composed mechanism ${\cal M}$, which applies ${\cal M}_i$ in sequential order from $i=1$ to $i=T$, is 
$(f_1\otimes \ldots \otimes f_T)$-DP.
The tensor product is commutative.

As a special case Corollary 3.3 in \citep{dong2021gaussian} states that composition of multiple Gaussian operators $G_{\mu_i}$ results in $G_{\mu}$ where 
$$\mu=\sqrt{\sum_i \mu_i^2}.$$

\subsection{Tight Analysis DP-SGD}

%
We are now able to formulate the differential privacy guarantee of original DP-SGD 
since it is a composition of  subsampled Gaussian DP mechanisms. Theorem 5.1 in \citep{dong2021gaussian} states that DP-SGD 
as introduced in \cite{abadi2016deep} 
is
$$ C_{m/N}(G_{\sigma^{-1}})^{\otimes T}\mbox{-DP},$$
where $T=(N/m)\cdot E$ is the total number of local rounds. 
Since each of the theorems and results from \citep{dong2021gaussian} enumerated above are exact, we have a tight analysis.
This leads in \citep{zhu2021optimal} to a (tight) differential privacy accountant (using complex characteristic functions for each of the two hypotheses based on taking  Fourier transforms), which can be used by a client to keep track of its current DP guarantee and to understand when to stop  helping the server to learn a global model. Because the accountant is tight, it improves (in general) over the momentum accountant method of \cite{abadi2016deep}.


\subsection{Group Privacy}
\label{sec:groupDP}

%
Theorem 2.14 in \citep{dong2021gaussian} analyzes how privacy degrades if $d$ and $d'$ do not differ in just one sample, but differ in $g$ samples. If a mechanism is $f$-DP, then it is $$[1-(1-f)^{\circ g}]\mbox{-DP}$$ for groups of size $g$ (where $\circ g$ denotes the $g$-fold iterative composition of function $1-f$, where $1$ denotes the constant integer value $1$ and not the identity function, i.e., $(1-f)(\alpha)=1-f(\alpha)$). This is a tight statement in that {\em there exist} $f$ such that the trade-off function for groups of size $g$ cannot be bounded better. In particular, for $f=G_\mu$ we have $G_{g\mu}$-DP for groups of size $g$. 

The intuition behind the $[1-(1-f)^{\circ g}]$-DP result is that the adversary can create a sequence of data sets $d_0=d$, $d_1$, \ldots, $d_{g-1}$, $d_{g}=d'$ such that each two consecutive data sets $d_i$ and $d_{i+1}$ are neighboring. We know that $T({\cal M}(d_i),{\cal M}(d_{i+1}))\geq f$. 
For each rejection rule we may plot a point (in x and y coordinates) $$(\mathbb{E}_{o\sim {\cal M}(d_i)}[\phi(o)], \ \mathbb{E}_{o\sim {\cal M}(d_{i+1})}[\phi(o)]).$$ 
Since $f(\alpha)$ is a lower bound on the Type I vs Type II error curve, the resulting collection of points is upper bounded by the curve $1-f(\alpha)$.
We have that $\alpha=\mathbb{E}_{o\sim {\cal M}(d_i)}[\phi(o)]$ is mapped to $$\mathbb{E}_{o\sim {\cal M}(d_{i+1})}[\phi(o)]\leq 1-f(\alpha)=(1-f)(\alpha).$$ By transitivity, we have that $\alpha=\mathbb{E}_{o\sim {\cal M}(d=d_0)}[\phi(o)]$ is mapped to $$\mathbb{E}_{o\sim {\cal M}(d'=d_{g})}[\phi(o)]\leq (1-f)^{\circ g}(\alpha).$$ This yields the lower bound $$T({\cal M}(d),{\cal M}(d'))\geq 1-(1-f)^{\circ g}$$ on the Type I vs Type II error curve.

Let $\phi[\alpha]$ denote a rejection rule that realizes the mapping from $$\alpha=\mathbb{E}_{o\sim {\cal M}(d_i)}[\phi[\alpha](o)] \ \ \mbox{ to } \ \ (1-f)(\alpha)=\mathbb{E}_{o\sim {\cal M}(d_{i+1})}[\phi[\alpha](o)].$$ Then  
the mapping from $(1-f)^{\circ i}(\alpha)=\mathbb{E}_{o\sim {\cal M}(d_i)}[\phi(o)]$ to $(1-f)^{\circ (i+1)}(\alpha)=\mathbb{E}_{o\sim {\cal M}(d_{i+1})}[\phi(o)]$ is realized by $\phi=\phi[(1-f)^{\circ i}(\alpha)]$.
This shows that the lower bound $1-(1-f)^{\circ g}$ is tight only if we can choose all $\phi[(1-f)^{\circ i}(\alpha)]$ equal to one another.
This is not the case for DP-SGD 
for which it turns out that this lower bound may not be tight at all; rather than a multiplicative factor $g$ as in the mentioned $G_{g\mu}$-DP guarantee we see a $\sqrt{g}$ dependency 
in this paper.
This is done by considering how, due to sub-sampling, the $g$ differentiating samples are distributed across all the rounds within an epoch and how composition of trade-off functions across rounds yields the $\sqrt{g}$ dependency.


\section{Helpful Lemmas}
\label{sec:lemmas}

Our DP analysis depends on the next lemmas which we state first. Let a mechanism ${\cal M}$ be defined as a probabilistic sum of base mechanisms ${\cal M}_i$, that is, 
$${\cal M}=\sum_i q_i {\cal M}_i$$ for some probabilities $q_i$ with $\sum_i q_i=1$. ${\cal M}$ executes mechanism ${\cal M}_i$ with probability $q_i$. Let $f_i$ be a trade-off function for ${\cal M}_i$, that is, 
$$T({\cal M}_i(d),{\cal M}_i(d'))\geq f_i.$$  
Then the following lemmas provide bounds on the trade-off function $T({\cal M}(d),{\cal M}(d'))$.

\begin{lemma} \label{lem:mech1} Let $T({\cal M}_i(d),{\cal M}_i(d'))\geq f_i$ and define
$$ f(\alpha) = \inf_{\{\alpha_i\}} \left\{ \left. \sum_i q_i f_i(\alpha_i)  \ \right| \ \sum_i q_i \alpha_i=\alpha \right\}. $$
Then $f$ is a trade-off function and 
$T({\cal M}(d),{\cal M}(d'))\geq f$. 

In addition, if all $f_i$ are symmetric (i.e., $f_i^{-1}$ as defined in (\ref{inverse}) is equal to $f_i$), then $f$ is symmetric as well. In this case we also have $f_i(f_i(\alpha))=\alpha$ and $f(f(\alpha))=\alpha$ for $\alpha\in [0,1]$.
\end{lemma}

\begin{lemma} \label{lem:mech2}
Suppose that $f_1\leq f_2\leq f_3 \leq \ldots$ with $T({\cal M}_i(d),{\cal M}_i(d'))\geq f_i$. Let
$p_t=\sum_{i< t} q_i$.
Then, we have the lower bound 
$T({\cal M}(d),{\cal M}(d'))(\alpha) \geq f(\alpha) \geq f_t(\min\{1,\alpha+p_t\})$, where $f$ is as defined in Lemma \ref{lem:mech1}.
\end{lemma}

\begin{lemma} \label{lem:mech3}
Suppose that $f_1\leq f_2\leq f_3 \leq \ldots$ with $f_i=T({\cal M}_i(d),{\cal M}_i(d'))$. Let
$p_t=\sum_{i> t} q_i$. Assume there exists a probability distribution $P$ on the real numbers with log concave probability density and there exist real numbers $\{t_i\}$ such that 
$f_i = T (P, t_i + P )$.  Then, we have the upper bound
$p_t+(1-p_t)f_t(\min \{1,\alpha/(1-p_t)\})\geq T({\cal M}(d),{\cal M}(d'))(\alpha)$.
\end{lemma}

\begin{lemma} \label{lem:mech4}
If $f=f_i=T({\cal M}_i(d),{\cal M}_i(d'))$ for all $i$, then  $T({\cal M}(d),{\cal M}(d'))= f$.
\end{lemma}

The next lemma allows us to solve for the infinum in Lemma \ref{lem:mech1}.

\begin{lemma} \label{lem:mech5}
Let 
$$ f(\alpha)=\inf_{\{\alpha_i\}: \sum_i q_i \alpha_i=\alpha} \sum_i q_i f_i(\alpha_i)$$
be the trade-off function of Lemma \ref{lem:mech1} where $f_i=G_{\mu_i}$.
Let $\Lambda(\alpha)$, $\alpha\in [0,1]$, be implicitly defined by the equation
$$1-\alpha = \sum_{i} q_i \Phi(\frac{\Lambda(\alpha)}{\mu_i} + \frac{\mu_i}{2}).$$
Then,
$$ f(\alpha)=\sum_{i} q_i \Phi(\frac{\Lambda(\alpha)}{\mu_i} - \frac{\mu_i}{2}).$$
\end{lemma}



\subsection{Proof of Lemma \ref{lem:mech1}} 
We want to prove a lower bound of the trade-off function of ${\cal M}= \sum_i q_i {\cal M}_i$ in terms of trade-off functions $f_i$ for ${\cal M}_i$:  We define
$$\alpha_{i,\phi} = \mathbb{E}_{o\sim {\cal M}_i(d)}[\phi(o)]  \mbox{ and } \beta_{i,\phi} = 1- \mathbb{E}_{o\sim {\cal M}_i(d')}[\phi(o)]$$
and notice that for ${\cal M}$ we have
$$ \alpha_\phi = \sum_i q_i \alpha_{i,\phi} \mbox{ and } \beta_\phi = \sum_i q_i \beta_{i,\phi}. $$
We derive
\begin{eqnarray}
T({\cal M}(d),{\cal M}(d'))(\alpha) &=& \inf_\phi \{ \beta_\phi \ : \ \alpha_\phi \leq \alpha \} \nonumber \\
&=&
\inf_\phi \{ \sum_i q_i\beta_{i,\phi} \ : \ \sum_i q_i\alpha_{i,\phi} \leq \alpha \} \nonumber \\
&=&
\inf_{\{\alpha_i\}: \sum_i q_i \alpha_i=\alpha} \inf_\phi \{ \sum_i q_i\beta_{i,\phi} \ : \ \forall_i \alpha_{i,\phi} \leq \alpha_i \} \label{eqTinf} \\
&\geq&
\inf_{\{\alpha_i\}: \sum_i q_i \alpha_i=\alpha} \sum_i q_i \inf_\phi \{ \beta_{i,\phi} \ : \ \alpha_{i,\phi} \leq \alpha_i \}  \nonumber \\
&\geq &
\inf_{\{\alpha_i\}: \sum_i q_i \alpha_i=\alpha} \sum_i q_i f_i(\alpha_i). \nonumber 
\end{eqnarray}
Notice that the lower bound is indeed a trade-off function since it is continuous, non-increasing, at least $0$, and $\leq 1-\alpha$ for $\alpha\in [0,1]$, and convexity follows from the next argument: Let $\alpha= p\alpha_0 + (1-p)\alpha_1$. By convexity of $f_i$, $f_i(p\alpha_{0,i}+(1-p)\alpha_{1,i})\leq pf_i(\alpha_{0,i})+(1-p)f_i(\alpha_{1,i})$. Convexity of the lower bound follows from
\begin{eqnarray*}
&& \inf_{\{\alpha_i\}: \sum_i q_i \alpha_i=p\alpha_0+(1-p)\alpha_1} \sum_i q_i f_i(\alpha_i) \\
&=& 
\inf_{\{\alpha_i=p\alpha_{0,i}+(1-p)\alpha_{1,i}\}: \sum_i q_i \alpha_{0,i}=\alpha_0, \sum_i q_i \alpha_{1,i}=\alpha_1
} \sum_i q_i f_i(\alpha_i) \\
&\leq&
\inf_{\{\alpha_i=p\alpha_{0,i}+(1-p)\alpha_{1,i}\}: \sum_i q_i \alpha_{0,i}=\alpha_0, \sum_i q_i \alpha_{1,i}=\alpha_1
} p \sum_i q_i f_i(\alpha_{0,i}) +(1-p) \sum_i q_i f_i(\alpha_{1,i})\\
&= &
p \inf_{\{\alpha_{0,i}\}: \sum_i q_i \alpha_{0,i}=\alpha_0} \sum_i q_i f_i(\alpha_{0,i})
+
(1-p) \inf_{\{\alpha_{1,i}\}: \sum_i q_i \alpha_{1,i}=\alpha_1} \sum_i q_i f_i(\alpha_{1,i}).
\end{eqnarray*}

In addition, if all $f_i$ are symmetric, then
\begin{eqnarray*}
f^{-1}(\alpha) &=& \inf \{ t\in [0,1] \ | \ f(t)\leq \alpha \} \\
&=& 
\inf \{ t\in [0,1] \ | \ 
\inf \{  \sum_i q_i f_i(\alpha_i)  \ | \ \sum_i q_i \alpha_i=t \} \leq \alpha \}.
\end{eqnarray*}
Since all $f_i$ are trade-off functions and therefore non-increasing, increasing $t=\sum_i q_i \alpha_i$ implies that each $\alpha_i$ in the sum $\sum_i q_i f_i(\alpha_i)$ can be increased, hence, $\sum_i q_i f_i(\alpha_i)$ can be decreased. This implies that the infinum of $\sum_i q_i f_i(\alpha_i)$ will be smaller. We want the smallest (infinum) $t$ such that the infinum of $\sum_i q_i f_i(\alpha_i)$ is at most $\alpha$. So, we want the smallest (infinum) $t=\sum_i q_i \alpha_i$ for which $\sum_i q_i f_i(\alpha_i)=\alpha$. This yields
\begin{eqnarray*}
f^{-1}(\alpha) &=& 
\inf \{ t\in [0,1] \ | \ 
\inf \{  \sum_i q_i f_i(\alpha_i)  \ | \ \sum_i q_i \alpha_i=t \} \leq \alpha \} \\
&=& \inf \{  \sum_i q_i \alpha_i  \ | \ \sum_i q_i f_i(\alpha_i)=\alpha \}.
\end{eqnarray*}

Since all $f_i$ are symmetric, we have
$$ f_i(\alpha)=f_i^{-1}(\alpha) =\inf \{ t\in [0,1] \  | \ f(t)\leq \alpha\}. $$
Since $f_i(t)$ is non-increasing, this
implies that, for $\alpha\in [0,1]$,
$f_i(\alpha)=t$ for some $t\in [0,1]$ such that $f_i(t)=\alpha$. This proves $f_i(f_i(\alpha))=\alpha$. And by substituting $\beta_i=f_i(\alpha_i)$ we have
\begin{eqnarray*}
f^{-1}(\alpha) &=& 
\inf \{  \sum_i q_i \alpha_i  \ | \ \sum_i q_i f_i(\alpha_i)=\alpha \} \\
&=& \inf \{  \sum_i q_i f_i(f_i(\alpha_i))  \ | \ \sum_i q_i f_i(\alpha_i)=\alpha \} \\
&=& \inf \{  \sum_i q_i f_i(\beta_i)  \ | \ \sum_i q_i \beta_i=\alpha \} = f(\alpha).
\end{eqnarray*} 
We conclude that $f$ is symmetric. 

The above derivation for $f_i(f_i(\alpha))=\alpha$ also holds for $f$ and we have that $f(f(\alpha))=\alpha$ for $\alpha\in [0,1]$.

\subsection{Proof of Lemma \ref{lem:mech2}}
The lower bound 
follows from Lemma \ref{lem:mech1} with
\begin{eqnarray*}
\sum_{i} q_i f_i(\alpha_i) &\geq& \sum_{i\geq t} q_i f_i(\alpha_i)  \geq 
\sum_{i\geq t} q_i f_t(\alpha_i)
=
\sum_{i\geq t} q_i f_t(\alpha_i) +\sum_{i<t} q_i f_t(1)\\
&\geq&
f_t( \sum_{i\geq t} q_i \alpha_i +\sum_{i<t} q_i)
= 
f_t(\alpha + \sum_{i<t} q_i (1-\alpha_i))
\geq 
f_t(\min\{1,\alpha+p_t\}).
\end{eqnarray*}

\subsection{Proof of Lemma \ref{lem:mech3}}
%
%
%
We refer to Proposition A.3 and the proof of (6) in \cite{dong2021gaussian} from which we conclude that, for all $\hat{\alpha}\in [0,1]$, there exists a rejection rule $\phi^*$ for which $\hat{\alpha}=\textbf{E}_{o\sim P}[\phi^*(o)]$ and, for all $i$,
$$ \beta_{i,\phi^*} =\inf_\phi \{ \beta_{i,\phi} : \alpha_{i,\phi}
\leq \hat{\alpha} \}.
$$
For completeness, it turns out that the rejection rule $\phi^*$ is defined by a threshold mechanism: Reject a sample $o$ if $o\geq t$ with $t=F^{-1}(1-\hat{\alpha})$ 
where $F$ is the cdf of probability distribution $P$.


The upper bound follows from (\ref{eqTinf}) by choosing $\alpha_i=0$ for $i> t$, $\alpha_i= \min\{1,\alpha/(1-p_t)\}$ for $i\leq t$ with $p_t=\sum_{i>t} q_i$, and  choose the rejection rule $\phi^*$ corresponding to $\hat{\alpha}=\min\{1,\alpha/(1-p_t)\}$.
We derive
\begin{eqnarray*}
\inf_\phi \{ \sum_i q_i\beta_{i,\phi} \ : \ \forall_i \alpha_{i,\phi} \leq \alpha_i \} &\leq &
\inf_\phi \{ \sum_{i> t} q_i + \sum_{i\leq t} q_i\beta_{i,\phi} \ : \ \forall_i \alpha_{i,\phi} \leq \alpha_i \} \\
&=&
p_t + 
\inf_\phi \{  \sum_{i\leq t} q_i\beta_{i,\phi} \ : \ \forall_i \alpha_{i,\phi} \leq \alpha_i \} \\
&=&
p_t + 
\inf_\phi \{  \sum_{i\leq t} q_i\beta_{i,\phi} \ : \ \forall_i \alpha_{i,\phi} \leq \min\{1,\alpha/(1-p_t)\} \}\\
&\leq &
p_t + 
\sum_{i\leq t} q_i\beta_{i,\phi^*} \\
&=& 
p_t + 
\sum_{i\leq t} q_if_i(\min\{1,\alpha/(1-p_t)\}) \\
&\leq &
p_t + 
\sum_{i\leq t} q_if_t(\min\{1,\alpha/(1-p_t)\}) \\
& =& 
 p_t + (1-p_t) f_t(\min\{1,\alpha/(1-p_t)\}).
\end{eqnarray*}
\subsection{Proof of Lemma \ref{lem:mech4}}
The lemma is about the special case where $f=f_i$ for all $i$. By taking $t=1$ in the lower bound of the second statement, we have $p_t=\sum_{i<t} q_i =0$ and lower bound $f_t(\alpha)=f(\alpha)$. By taking $t$ equal to the largest index in the upper bound of the third statement, we have $p_t=\sum_{i>t} q_i =0$ and upper bound $f_t(\alpha)=f(\alpha)$ (for the upper bound in this special case we do not need to satisfy $f_i=T(P,t_i+P)$). Combination of these bounds yields $T({\cal M}(d),{\cal M}(d'))=f$.

\subsection{Proof of Lemma \ref{lem:mech5}}

\noindent 
{\bf Proof:}
The infinum can be solved by using a Lagrange multiplier: We define 
$$  \sum_{i} q_i G_{\mu_i}(\alpha_i) - \lambda (\alpha -\sum_{i} q_i \alpha_i)
$$
as a function of all $\alpha_i$ and $\lambda$
and find its stationary points.
We remind the reader that $G_\mu(x)=\Phi(\Phi^{-1}(1-x)-\mu)$ and $\Phi'(x) = e^{-x^2/2}/\sqrt{2\pi}$, hence,
\begin{eqnarray} G'_\mu(x) &=& \Phi'(\Phi^{-1}(1-x)-\mu) \cdot \frac{1}{\Phi'(\Phi^{-1}(1-x))} \cdot (-1) \nonumber \\
&=&
- \frac{e^{-(\Phi^{-1}(1-x)-\mu)^2/2}}{e^{-\Phi^{-1}(1-x)^2/2}}
=- e^{\mu \Phi^{-1}(1-x) -\mu^2/2}.\label{eqder}
\end{eqnarray}
For the stationary point we have the equations
\begin{eqnarray*}
0 &=& -e^{\mu_i \Phi^{-1}(1-\alpha_i) -\mu_i^2/2} + \lambda, \\
\alpha &=& \sum_i q_i \alpha_i.
\end{eqnarray*}
This shows that
$$ \alpha_i = 1-\Phi(\frac{\ln(\lambda)}{\mu_i} + \frac{\mu_i}{2}) $$
and therefore
$$ 1-\alpha = \sum_{i} q_i \Phi(\frac{\ln(\lambda)}{\mu_i} + \frac{\mu_i}{2}).$$
The last equation can be used to solve for $\ln(\lambda)$ (this is possible because $\ln(\lambda)\rightarrow -\infty$ makes the sum zero, while $\ln(\lambda)\rightarrow +\infty$ makes the sum one) with which the different $\alpha_i$ can be computed using the first equation, which in turn allows us to evaluate the trade-off function for the overall mechanism in $\alpha$. Notice that $$G_{\mu_j}(\alpha_j)=
\Phi(\frac{\ln(\lambda)}{\mu_j} - \frac{\mu_j}{2}).
$$
Let $\ln(\lambda)$ be represented as the function value $\Lambda(\alpha)$ and the lemma follows.

\section{Proof DP Analyis}
\label{sec:DPproof}

An adversary, who observes a transmitted noised local update $\bar{U}$, wants to gain information about whether training set $d$ or $d'$ was locally used by the client. 
We analyse  the general situation where sets $d$ and $d'$ differ in a subset of  samples. More precisely, 
$$ d= (d\cap d') + z \mbox{ and } d'=(d\cap d') + z'$$
for some sets $z$ and $z'$. Sets $z$ and $z'$ contain differentiating samples that can be used by the adversary to conclude whether $d$ or $d'$ was used in the computation of $\bar{U}$.

\subsection{Simulator}

%
Suppose that $d$ is the truth.
Let $U$ correspond to a  computation based on a subset $\{\xi_i \in d : i\in S_b\}$. The adversary observes $\bar{U}\leftarrow U+{\cal N}(0,(2C\sigma)^2{\bf I})$. If the used  $\{\xi_i \in d : i\in S_b\}$ is a subset of the intersection $d\cap d'$, then none of the samples $\xi_i$ are differentiating. For each $\xi_i$ there exists a $\xi'_j\in d'$ such that $\xi_i=\xi'_j$. In other words, there exists a simulator with access to $d'$ (but not $d$) who could have produced $U$ (with equal probability). This means that $U$ and as a consequence $\bar{U}$ cannot be used by the adversary to differentiate whether $d$ or $d'$ was used.

Now suppose that $\{\xi_i \in d : i\in S_b\}$ has some differentiating samples in $z$. Now computation of $U$ can be simulated by a simulator with access to $d'$ plus the differentiating samples in $z$. This is because each $\xi_i$ is either in $d\cap d'\subseteq d'$ or in $z$. Suppose that $\{\xi_i \in d : i\in S_b\}$ contains exactly $t$ differentiating samples from $z$. 

If $d'$ were the truth, then the algorithm that produces the local update has no access to $z$. At best it can mimic the simulator for the non-differentiating  $\xi_i\in d\cap d'$ and choose $t$ other samples from $d'$ to replace the $t$ differentiating samples from $z$ (according to some strategy). Let $S'_b$ correspond to the set of samples $\{\xi'_i \in d' : i\in S'_b\}$ chosen by the mimicked simulator. 

In a similar way, the mimicking of the simulator produces a partition $\{S'_{b,h}\}_{h=1}^m$. We have that $\{\xi_i : i\in S_{b,h}\}$ has exactly $t_h$ differentiating samples if and only if $\{\xi_i : i\in S_{b,h}\}$ has exactly $t_h$ differentiating samples. Their non-differentiating samples are the same. Notice that $\sum_{h=1}^m t_h =t$.

The same argument of the mimicking of the simulator also holds at the scale of an epoch which produces $\{S'_b\}_{b=1}^{N/(ms)}$.

The above argument is sound because ${\tt Sample}_{s,m}$ samples from a set of indices $\{1,\ldots, N\}$ and does not sample from the data set $d$ directly. This means that sampling is independent of the data set. And we can formulate our simulator by choosing a suitable index to data sample mapping (permutation). Since any permutation from indices to data samples is equally likely and randomized before each call to ${\tt Sample}_{s,m}$, see Algorithm \ref{alg:genDPSGD}, the way the mimicked  simulator samples from a set of size $N'=|d'|$ cannot be distinguished from how ${\tt Sample}_{s,m}$ samples from a set of size $N=|d|$ if $N=N'$.

For the subsampling strategy, it  matters only very little whether ${\tt Sample}_{s,m}$ samples from a set of size $N$ and the mimicked simulator samples from an index set of  size $N'=|d'|$ with $N\neq N'$. This is because each set $S_b$ is a random subset of size $ms$ out of $N$ indices where in practice $ms\ll N$ and $N'$ is close to $N$ (there are at most $g$ differentiating samples), hence, the difference in frequency of selecting a specific sample by ${\tt Sample}_{s,m}$ or by the mimicked simulator is very small, which we simply {\em assume leads to privacy leakage small enough to be discarded in our DP analysis} (previous work also builds on this assumption even though previous work does not state this assumption explicitly). Similarly, the shuffling strategy for $N\neq N'$ is assumed to suffer sufficiently small privacy leakage that can be discarded. 

Of course, the client should not reveal the exact value of $N=|d|$ of his local data set $d$ (as part of its mechanism), otherwise, the adversary can use this information to figure out whether, for example, $d$ or $d'$ with one extra data sample is being used. 
Notice that in practice, the adversary indeed does not exactly know $N$ or $N'$. The adversary only knows the differentiating samples $z$ and $z'$ and a lot of information about the kind of data samples that are in the intersection $d\cap d'$ without knowing exactly its size.
It is realistic to assume that, besides knowing the differentiating samples, the adversary only knows $|d\cap d'|+noise$ as a-priori information where the noise is large enough to yield a sufficiently strong DP guarantee in itself.

\subsection{Hypothesis Testing}

%
In order to distinguish whether the observed noised update corresponds to $d$ or $d'$, the best case for the adversary is to have knowledge about $S_b$ and $S'_b$ together with the index mappings to $d$ and $d'$ so that it can compute $U$ corresponding to $S_b$ and compute $U'$ corresponding to $S'_b$. In order to be able to do so, we also give the adversary access to the randomness used by algorithm ${\cal A}$ (this includes when and how the local model $w$ is overwritten by the interrupt service routine). This means that the adversary needs to perform hypothesis testing on whether the observed noised update comes from $U+{\cal N}(0,(2C\sigma)^2{\bf I})$ or $U'+{\cal N}(0,(2C\sigma)^2{\bf I})$.

\subsection{Adversarial Model}
\label{app:adversary}

%
Above  defines a strong adversary ${\cal A}_1$ as discussed separately in Appendix \ref{sec:adv}. In practice, the adversary does not know $S_b$ or $S'_b$ and we may treat $U$ and $U'$ as random variables rather than fixed vectors. There may be a plausible assumption on how data samples are generated such that the probability distribution for $U$ and $U'$ can be characterized. If there exists a (round dependent) distribution $\hat{{\cal N}}$ centered around $0$ such that $U\leftarrow u+\hat{{\cal N}}$ and $U'\leftarrow u'+\hat{{\cal N}}$ for fixed vectors $u$ and $u'$ that can be computed by the adversary, then the adversary needs to do hypothesis testing between $u+\hat{{\cal N}}+{\cal N}(0,(2C\sigma)^2)$ and $u'+\hat{{\cal N}}+{\cal N}(0,(2C\sigma)^2)$. This reduces privacy leakage since hypothesis testing becomes less accurate because of the uncertainty added by $\hat{{\cal N}}$. Our strong (worst-case) adversary has no such uncertainty.

\subsection{Gaussian Trade-Off Function}
\label{app:gauss}

%
Let ${\tt rand}$ denote the used randomness by ${\cal A}$. Then, given our strong adversarial model, we may write $a_h={\cal A}({\tt rand};w,\{\xi_i\}_{i\in S_{b,h}})$ in Algorithm \ref{alg:genDPSGD}. We have
$$ U= \sum_{h=1}^m [{\cal A}({\tt rand};w,\{\xi_i\}_{i\in S_{b,h}})]_C \ \ \mbox{ and } \ \ U'= \sum_{h=1}^m [{\cal A}({\tt rand};w,\{\xi'_i\}_{i\in S'_{b,h}})]_C.
$$
For data set $d$, we introduce  parameter
\begin{equation} L(\{S_{b,h}\}_{h=1}^m) = |\{ 1\leq h\leq m \ : \ \{ \xi_i : i\in S_{b,h}\} \cap z \neq \emptyset \}|.\label{eqL}
\end{equation}
Suppose that $L(\{S_{b,h}\}_{h=1}^m)=k$. Then exactly $m-k$ terms in the sums of $U$ and $U'$ coincide. Let $h_1, \ldots, h_k$ be the indices of the terms that do not coincide. Then,
\begin{eqnarray*}
\| U-U'\| &=& \| \sum_{h\in\{h_1,\ldots,h_k\}} [{\cal A}({\tt rand};w,\{\xi_i\}_{i\in S_{b,h}})]_C
-
\sum_{h\in\{h_1,\ldots,h_k\}} [{\cal A}({\tt rand};w,\{\xi'_i\}_{i\in S'_{b,h}})]_C \|\\
&\leq& 2kC.
\end{eqnarray*}
The norm $\|U-U'\|$  is upper bounded by\footnote{The original argument for DP-SGD with $s=v=1$ considers neighboring $d$ and $d'$, that is, $|z\cup z'|=1$. Update $U$ is computed as the sum $U=\sum_{i\in S_b} [\nabla f(w,\xi_i)]_C$. Their DP argument wrongly states that the vector $x= \sum_{i\in S_b\setminus \{n+1\}} [f(w,\xi_i)]_C$, where $n+1$ corresponds to the differentiating sample between $d$ and $d'$, is the same for both $d$ and $d'$ and that 
$U$ for $d$ and $d'$ differs by $[f(w,\xi'_{n+1})]_C$, leading to an upper bound of only $C$. 
However, the samples  chosen from $d$ and $d'$ corresponding to $S_b$  differ in exactly one element and we need the extra factor $2$. This is consistent with our presented analysis. We notice that this was already observed by 
\citep{dong2021gaussian}, see p. 25 bottom paragraph. Appendix \ref{app:factor} has an extended discussion explaining how 
\cite{opacus} implement a slightly different subsampling which is fits the DP-SGD analysis and allows a factor $2$ improvement for $g=1$.} $2kC$. This upper bound is met with equality if all the terms happen to be orthogonal to one another. Without further assumptions on how data samples are generated, we must assume this best case for the adversary. That is, we must assume the adversary can possibly perform a hypothesis test  of $U+{\cal N}(0,(2C\sigma)^2{\bf I})$ versus $U'+{\cal N}(0,(2C\sigma)^2{\bf I})$ where $\|U-U'\|=2kC$. 

After projecting the observed noised local update $\bar{U}$ onto the line that connects $U$ and $U'$ and after normalizing by dividing by $2C$, we have that differentiating whether $\bar{U}$ corresponds to $d$ or $d'$ is equivalent to differentiating whether a received output is from ${\cal N}(0,\sigma^2)$ or from  ${\cal N}(k,\sigma^2)$. Or, equivalently, from ${\cal N}(0,1)$ or from  ${\cal N}(k/\sigma,1)$. 
This is how the Gaussian trade-off function $G_{k/\sigma}$ comes into the picture. 

\subsection{Round Mechanism with Simulator}

%
Let $\bar{{\cal M}}_b$ be the mechanism that produces $\bar{U}$ in round $b$ based on data set $d$ with $L(S_b)=k$. And let $\bar{{\cal M}}^{\tt sim}_b$ be the mechanism that represents the mimicked simulator based on data set $d'$. Then the above argument yields
$$ T(\bar{{\cal M}}_b(d),\bar{{\cal M}}^{\tt sim}_b(d'))= G_{k/\sigma}.$$

\subsection{Sampling Distribution}
\label{app:sampling}

%
We want to analyse privacy leakage over a whole epoch. 
For this reason, we are interested in the joint distribution of $\{L(\{S_{b,h}\}_{h=1}^m)\}_{b=1}^{N/(sm)}$ generated by ${\tt Sample}_{s,m}$. Let $|z|=g$, that is, $d$ contains $g$ differentiating samples.
We define
\begin{equation}
q(c_1,c_2,\ldots, c_g)
=
{\tt Pr}[ \forall_{k=1}^g c_k=\# \{ b : L(\{S_{b,h}\}_{h=1}^m)=k\}  \ | \ \{S_{b,h}\}_{b=1,h=1}^{N/(sm),m} \leftarrow 
{\tt Sample}_{s,m}]. \label{eqq}
\end{equation}
Here, $c_k$ indicates the number of rounds $b$ that have $L(\{S_{b,h}\}_{h=1}^m)=k$ (since the $\{S_{b,h}\}_{h=1}^m$ partition set $S_b$, $0\leq k\leq g$). 


\subsection{Epoch Mechanism with Simulator}

%
Let $(c_1,\ldots, c_g)$ represent an instance of an epoch based on $d$ with $c_k=\# \{ b : L(\{S_{b,h}\}_{h=1}^m)=k\}$ for $1\leq k\leq g$. The probability of such an instance occurring is equal to $q(c_1,c_2,\ldots, c_g)$.
Let $\bar{{\cal M}}_{c_1,\ldots,c_g}$ be the mechanism that produces a sequence of noised local updates $\bar{U}$ for the given epoch instance based on $d$. Let $\bar{{\cal M}}^{\tt sim}_{c_1,\ldots,c_g}$ be the mechanism that represents the mimicked simulator based on $d'$.
Then, by composition over rounds within the epoch instance (where we use its commutative property
),
we conclude that
\begin{equation} T(\bar{{\cal M}}_{c_1,\ldots,c_g}(d),\bar{{\cal M}}^{\tt sim}_{c_1,\ldots,c_g}(d'))= G_{1/\sigma}^{\otimes c_1} \otimes G_{2/\sigma}^{\otimes c_2} \otimes \ldots \otimes G_{g/\sigma}^{\otimes c_g}= G_{\sqrt{\sum_{k=1}^g c_k k^2}/\sigma}. \label{eqG}
\end{equation}

Let ${\cal M}$ be the overall mechanism  which represents one epoch of updates based on $d$ and let ${\cal M}^{\tt sim}$ represent the mimicked simulator.  Then,
\begin{equation} {\cal M} = \sum_{c_1,\ldots, c_g} q(c_1,\ldots,c_g) \cdot
\bar{{\cal M}}_{c_1,\ldots,c_g} \ \ \mbox{ and } \ \  {\cal M}^{\tt sim} = \sum_{c_1,\ldots, c_g} q(c_1,\ldots,c_g) \cdot
\bar{{\cal M}}^{\tt sim}_{c_1,\ldots,c_g}. \label{eqM}
\end{equation}

The final definition of the mimicked simulator ${\cal M}^{\tt sim}$ has in essence oracle access to the randomness used by ${\cal M}$. That is, if ${\cal M}$ chooses to use $\bar{{\cal M}}_{c_1,\dots, c_g}$ with probability $q(c_1,\ldots, c_g)$, then this information is passed on to ${\cal M}^{\tt sim}$ who will choose to use $\bar{M}^{\tt sim}_{c_1,\ldots, c_g}$.
It can do this if we assume the strong adversary with oracle access to the randomness used by ${\cal M}$ for selecting the sets $S_{b,h}$ and executing ${\cal A}$.


\subsection{Final Epoch Mechanism}

%
So far, we assumed $d$ to be the truth. If $d$ is the truth, then
\begin{equation} T({\cal M}(d),{\cal M}(d'))\geq T({\cal M}(d),{\cal M}^{\tt sim}(d')).\label{eqT}
\end{equation}
We have $\geq$ if we consider a weaker adversary without direct knowledge about the exact subset $S_b$ and its subsets $S_{b,h}$ with index mappings 
(for the stronger adversary we have equality).
The exact same analysis can be repeated for $d'$ being the truth. In this case we use parameter $g'=|z'|$ and obtain a similar lower bound as given above. We assume the best case for the adversary, that is, the smaller of the two lower bounds.

The larger $g$ the larger $z$ and probability $q(c_1,\ldots,c_g)$ shifts its mass towards vectors $(c_1,\ldots,c_g)$ with larger $c=\sqrt{\sum_{k=1}^g c_k k^2}$, see (\ref{eqL}) and (\ref{eqq}). This means that  the smaller trade-off functions $G_{c/\sigma}$ (corresponding to the larger $c$) are counted more in  mechanism ${\cal M}$, see (\ref{eqG}) and (\ref{eqM}).  
The best for the adversary is when its hypothesis testing is characterized by the smaller trade-off function ($d$ versus $d'$ being the truth). That is, the one corresponding to the larger of $g$ and $g'$. Therefore, without loss of generality, we assume $g= |z|\geq |z'|$ and equations (\ref{eqG}), (\ref{eqM}), and (\ref{eqT}) provide a lower bound for $T({\cal M}(d),{\cal M}(d'))$.
We notice that for our strong adversary  we have tight bounds in that (\ref{eqG}), (\ref{eqM}), and (\ref{eqT}) can be met with equality. 

Theorem \ref{thm:main} in the main body assumes that the effect of $N\neq N'$ leads to privacy leakage small enough to be discarded in our DP analysis as discussed earlier. The theorem statement is for $N\neq N'$ and  also uses $g=\max\{|z|,|z'|\}$.   In the worst case, the total number $t=|z|+|z'|$ of samples in which $d$ and $d'$ differ are distributed as $|z|=t$ with $|z'|=0$ as this achieves the maximum $g=|z|+|z'|=t$. 
This scenario is met if $d=d'+z$ for a group of $g=|z|$ differentiating samples.

\subsection{Grouping Probability Mass}

%
We apply Lemma \ref{lem:mech1} for (\ref{eqG}) and (\ref{eqM})
to get for $T({\cal M}(d),{\cal M}^{\tt sim}(d'))(\alpha)$ the lower bound  
$$\inf_{\{\alpha_{c_1,\ldots, c_g}\}} 
\left\{ \left. 
\sum_{c_1,\ldots, c_g} q(c_1,\ldots,c_g) \cdot G_{\sqrt{\sum_{k=1}^g c_k k^2}/\sigma}(\alpha_{c_1,\ldots, c_g}) \ \right| \ \sum_{c_1,\ldots, c_g} q(c_1,\ldots,c_g) \cdot \alpha_{c_1,\ldots, c_g}=\alpha \right\}.
$$
Now we may apply Lemma \ref{lem:mech4}
and group probabilities together in the sum of the lower bound that fit the same Gaussian trade-off function:
\begin{eqnarray*}
q(c) &=& \sum_{\{c_k\}_{k=1}^g : c^2=\sum_{k=1}^g c_k k^2} q(c_1,c_2,\ldots, c_g) \\
&=&
{\tt Pr}\left[ \left. c^2=\sum_{k=1}^g \# \{ b : L(\{S_{b,h}\}_{h=1}^m)=k\} \cdot k^2 \ \right| \ \{S_{b,h}\}_{b=1,h=1}^{N/(sm),m} \leftarrow {\tt Sample}_{s,m}\right].
\end{eqnarray*}
This results in
\begin{equation} T({\cal M}(d),{\cal M}^{\tt sim}(d'))(\alpha)\geq f(\alpha)=
 \inf_{\{\alpha_{c}\}_{c\in {\cal C}}}
\left\{  \left. \sum_{c\in {\cal C}} q(c) \cdot G_{c/\sigma}(\alpha_c) \ \right| \ \sum_{c\in {\cal C}} q(c) \cdot \alpha_c=\alpha \right\}
\label{lowG}
\end{equation}
where 
$$ {\cal C} = \left\{ \sqrt{\sum_{k=1}^g c_k k^2} \ : \ \forall_k \ c_k\in \{0,1,\ldots, N/(ms)\} \right\}$$
is the set of all possible values $c$.

Given the previous discussion on $g$, we notice that $f$ gets smaller for larger $g$ (i.e., the adversary gets better at hypothesis testing). 

\subsection{Calculation of the Infinum}

%
The infinum can be calculated using Lemma \ref{lem:mech5}. This yields function $f$ in Theorem \ref{thm:main2} in the main body.

\subsection{Lower and Upper Bounds on $T({\cal M}(d),{\cal M}(d'))$ and $T({\cal M}(d),{\cal M}^{\tt sim}(d'))$}

%
We apply Lemmas \ref{lem:mech2} and \ref{lem:mech3} to equations (\ref{eqG}) and (\ref{eqM}). We use the property that if $c<\hat{c}$, then $ G_{\hat{c}/\sigma}<G_{c/\sigma}$.


For the upper bound (the Gaussian trade-off function fits the required assumption in Lemma \ref{lem:mech3}) we group probabilities together and define
$$
u'_{\hat{c}} = \sum_{c\in {\cal C}: c<\hat{c}} q(c)
$$
Then,
$$ u'_{\hat{c}}+(1-u'_{\hat{c}})G_{\hat{c}/\sigma}(\min \{1,\alpha/(1-u'_{\hat{c}})\}) \geq T({\cal M}(d),{\cal M}^{\tt sim}(d'))(\alpha).$$
Since Gaussian trade-off functions are decreasing in $\alpha$ and are less than $1-\alpha$, we have for $u'_{\hat{c}}\leq u_{\hat{c}}\leq 1$
$$ u'_{\hat{c}}+(1-u'_{\hat{c}})G_{\hat{c}/\sigma}(\min \{1,\alpha/(1-u'_{\hat{c}})\})
\leq 
u'_{\hat{c}}+(1-u'_{\hat{c}})G_{\hat{c}/\sigma}(\alpha)
\leq 
u_{\hat{c}}+(1-u_{\hat{c}})G_{\hat{c}/\sigma}(\alpha).
$$

For the lower bound, we define 
$$ l'_{\hat{c}} = \sum_{c\in {\cal C}: c>\hat{c}} q(c) $$
and (by Lemma \ref{lem:mech2}) we have
$$ T({\cal M}(d),{\cal M}^{\tt sim}(d'))(\alpha) \geq f(\alpha) \geq G_{\hat{c}/\sigma}(\min \{1,\alpha+l'_{\hat{c}}\}).$$
For $l'_{\hat{c}}\leq l_{\hat{c}}\leq 1$ we have
$$G_{\hat{c}/\sigma}(\min \{1,\alpha+l'_{\hat{c}}\})\geq G_{\hat{c}/\sigma}(\min \{1,\alpha+l_{\hat{c}}\}).$$

Summarizing,
\begin{equation} T({\cal M}(d),{\cal M}(d'))\geq T({\cal M}(d),{\cal M}^{\tt sim}(d')), \label{eq:h3}
\end{equation}
which is tight for the strong adversary,
and we have the lower and upper bounds
\begin{equation}  \hat{f}^{\tt U}_{c^{\tt U}} \geq T({\cal M}(d),{\cal M}^{\tt sim}(d')) \geq f \geq \hat{f}^{\tt L}_{c^{\tt L}},
\label{eq:h4}
\end{equation}
where
\begin{eqnarray*} \hat{f}^{\tt U}_{c^{\tt U}}(\alpha) &=& u_{c^{\tt U}}+(1-u_{c^{\tt U}})G_{c^{\tt U}/\sigma}(\alpha), \\
\hat{f}^{\tt L}_{c^{\tt L}}(\alpha)&=&G_{c^{\tt L}/\sigma}(\min \{1,\alpha+l_{c^{\tt L}}\})
\end{eqnarray*}
are trade-off functions (continuous, non-increasing, convex, at least $0$, at most $1-\alpha$ for $\alpha\in [0,1]$).

\subsection{$f$-DP Guarantee}

%
Suppose that ${\cal M}$ is $h$-DP for all pairs of data sets $d$ and $d'$ such that $\max\{ |d\setminus d'|, |d'\setminus d|\}\leq g$. Since 
(\ref{eq:h3}) is tight for the strong adversary, we have
\begin{equation} T({\cal M}(d),{\cal M}(d'))\geq T({\cal M}(d),{\cal M}^{\tt sim}(d'))\geq h. \label{eq:h1}
\end{equation}
Combined with (\ref{eq:h4}) this implies
\begin{equation} \hat{f}^{\tt U}_{c^{\tt U}} \geq T({\cal M}(d),{\cal M}^{\tt sim}(d')) \geq h.
\label{eq:h2}
\end{equation}
By Lemma A.2 in \citep{dong2021gaussian}, 
$$ T({\cal M}(d'),{\cal M}(d))=T({\cal M}(d),{\cal M}(d'))^{-1}.$$
Since $f\geq h$ implies $f^{-1}\geq h^{-1}$ for non-increasing functions $f$ and $h$ (the inverse preserves order), (\ref{eq:h1}) and (\ref{eq:h2}) prove
$$\hat{f}^{\tt U}_{c^{\tt U}}{}^{-1} \geq h^{-1}.$$
We also know that if $h$ is a trade-off function for ${\cal M}$, then, by Proposition 2.4 in \citep{dong2021gaussian}, $\max\{h,h^{-1}\}$ is a trade-off function for ${\cal M}$. Therefore, without loss of generality we may assume that $h$ has this form, hence, $h$ is symmetric and we have
$$\hat{f}^{\tt U}_{c^{\tt U}}{}^{-1} \geq h^{-1}=h.$$
Together with (\ref{eq:h2}) this gives the upper bound
$$ h\leq \min \{ \hat{f}^{\tt U}_{c^{\tt U}}, \hat{f}^{\tt U}_{c^{\tt U}}{}^{-1}\}.$$
Since $h$ is convex, we can improve the upper bound by means of convexification by using the double conjugate. This yields
$$ h\leq \min \{ \hat{f}^{\tt U}_{c^{\tt U}}, \hat{f}^{\tt U}_{c^{\tt U}}{}^{-1}\}^{**}.$$

We also notice that (\ref{eq:h4}) shows that $f$ is a trade-off function for ${\cal M}$. Notice that $f$ is symmetric, see Lemma \ref{lem:mech1}, hence, from (\ref{eq:h4}) we infer that
$$ f=f^{-1}\geq \hat{f}^{\tt L}_{c^{\tt L}}{}^{-1}.$$
This proves
$$ f\geq \max \{ \hat{f}^{\tt L}_{c^{\tt L}}, \hat{f}^{\tt L}_{c^{\tt L}}{}^{-1} \}.$$









\subsection{Explicit Formula for the Lower  Bound}

%
Since the Gaussian trade-off function is symmetric, $G_{c^{\tt L}/\sigma}(G_{c^{\tt L}/\sigma}(x))=x$ for $x\in [0,1]$. Also notice that the Gaussian trade-off function in $x$ is strictly decreasing and is equal to $1$ for $x=0$ and equal to $0$ for $x=1$. From this we obtain that 
$$\hat{f}^{\tt L}_{c^{\tt L}}(t)  = G_{c^{\tt L}/\sigma}(\min \{1,t+l_{c^{\tt L}}\})\leq \alpha
$$
if and only if
$$ \min \{1,t+l_{c^{\tt L}}\}\geq
G_{c^{\tt L}/\sigma}(\alpha)$$
if and only if 
$$
[t\geq 1-l_{c^{\tt L}}] \mbox{ or }
[1-l_{c^{\tt L}}> t\geq G_{c^{\tt L}/\sigma}(\alpha)-l_{c^{\tt L}}].
$$
if and only if
$$ t \geq G_{c^{\tt L}/\sigma}(\alpha) -l_{c^{\tt L}}. 
$$
This shows that
$$
\hat{f}^{\tt L}_{c^{\tt L}}{}^{-1}(\alpha)
= \inf \{ t\in [0,1] \ | \ 
\hat{f}^{\tt L}_{c^{\tt L}}(t) \leq \alpha \}
=\max \{ 0, G_{c^{\tt L}/\sigma}(\alpha) -l_{c^{\tt L}}\}.
$$

We define $\beta\in [0,1]$ as the solution of
$$ 
G_{c^{\tt L}/\sigma}(\beta) -l_{c^{\tt L}} =\beta
$$
(there exists a unique solution because $l_{c^{\tt L}}\leq 1$, $G_{c^{\tt L}/\sigma}$ is convex, and $G_{c^{\tt L}/\sigma}(0)=1$).
Notice that
$$ \hat{f}^{\tt L}_{c^{\tt L}}{}^{-1}(\beta) = \beta = \hat{f}^{\tt L}_{c^{\tt L}}(\beta)$$
(the second equality follows from
$\beta + l_{c^{\tt L}} = G_{c^{\tt L}/\sigma}(\beta) \leq 1$, hence,
$\hat{f}^{\tt L}_{c^{\tt L}}(\beta)=G_{c^{\tt L}/\sigma}(G_{c^{\tt L}/\sigma}(\beta))=\beta$).

For $\alpha\in [0,\beta]$,
$$
\hat{f}^{\tt L}_{c^{\tt L}}(\alpha)=G_{c^{\tt L}/\sigma}(\alpha +l_{c^{\tt L}})
\leq 
G_{c^{\tt L}/\sigma}(\alpha) -l_{c^{\tt L}},
$$
where the inequality follows from the observation that the inequality is met with equality for $\alpha=\beta$ and the derivative of $G_{c^{\tt L}/\sigma}(\alpha +l_{c^{\tt L}})$ is at least the derivative of $G_{c^{\tt L}/\sigma}(\alpha)$ as a function of $\alpha$ due to the convexity of $G_{c^{\tt L}/\sigma}$.
Since $G_{c^{\tt L}/\sigma}(\alpha) -l_{c^{\tt L}}\geq G_{c^{\tt L}/\sigma}(\alpha +l_{c^{\tt L}})\geq 0$, we have $\hat{f}^{\tt L}_{c^{\tt L}}(\alpha)=G_{c^{\tt L}/\sigma}(\alpha) -l_{c^{\tt L}}$.
We conclude that for $\alpha\in [0,\beta]$,
$$ \max\{ \hat{f}^{\tt L}_{c^{\tt L}}, \hat{f}^{\tt L}_{c^{\tt L}}{}^{-1} \} (\alpha) = G_{c^{\tt L}/\sigma}(\alpha) -l_{c^{\tt L}}.
$$

By a similar argument, we have that for $\alpha\in [\beta,1]$,
$$ \max\{ \hat{f}^{\tt L}_{c^{\tt L}}, \hat{f}^{\tt L}_{c^{\tt L}}{}^{-1} \} (\alpha) = G_{c^{\tt L}/\sigma}(\min \{1, \alpha +l_{c^{\tt L}}\}).
$$
Notice that for $\alpha\in [\beta,1-l_{c^{\tt L}}]$ there exists an $\alpha'\in [0,\beta]$ such that $\alpha=  G_{c^{\tt L}/\sigma}(\alpha') -l_{c^{\tt L}}$, hence, $G_{c^{\tt L}/\sigma}(\min \{1, \alpha +l_{c^{\tt L}}\})=G_{c^{\tt L}/\sigma}(G_{c^{\tt L}/\sigma}(\alpha'))=\alpha'$. This confirms the symmetry of the curve $\max\{ \hat{f}^{\tt L}_{c^{\tt L}}, \hat{f}^{\tt L}_{c^{\tt L}}{}^{-1} \}(\alpha)$ around the diagonal $\alpha \rightarrow \alpha$. 

\subsection{Explicit Formula for the Upper  Bound}

%
By following the kind of argument we used for improving the lower bound, we observe that
$$
\hat{f}^{\tt U}_{c^{\tt U}}(t) = u_{c^{\tt U}}+(1-u_{c^{\tt U}})G_{c^{\tt U}/\sigma}(t) \leq \alpha
$$
if and only if
$$ t \geq G_{c^{\tt U}/\sigma}(\frac{\alpha -u_{c^{\tt U}} }{1-u_{c^{\tt U}}}).
$$
This shows that
$$
\hat{f}^{\tt U}_{c^{\tt U}}{}^{-1}(\alpha)
= \inf \{ t\in [0,1] \ | \ 
\hat{f}^{\tt U}_{c^{\tt U}}(t) \leq \alpha \}
= \min \{ 1, G_{c^{\tt U}/\sigma}(\frac{\alpha -u_{c^{\tt U}} }{1-u_{c^{\tt U}}}) \}.
$$

We define $\beta\in [0,1]$ as the solution of
$$ 
G_{c^{\tt U}/\sigma}(\frac{\beta -u_{c^{\tt U}} }{1-u_{c^{\tt U}}}) =\beta
$$
(there exists a unique solution because  $G_{c^{\tt L}/\sigma}$ is convex, and $G_{c^{\tt L}/\sigma}(0)=1$ and $G_{c^{\tt L}/\sigma}(1)=0$).
Notice that
$$ \hat{f}^{\tt U}_{c^{\tt U}}{}^{-1}(\beta) = \beta = \hat{f}^{\tt U}_{c^{\tt U}}(\beta)$$
(the second equality follows from
$G_{c^{\tt L}/\sigma}(\beta)=G_{c^{\tt L}/\sigma}(G_{c^{\tt U}/\sigma}(\frac{\beta -u_{c^{\tt U}} }{1-u_{c^{\tt U}}}))=\frac{\beta -u_{c^{\tt U}} }{1-u_{c^{\tt U}}}$ by the symmetry of $G_{c^{\tt U}/\sigma}$).


For $\alpha\in [0,\beta]$,
\begin{eqnarray*}
\hat{f}^{\tt U}_{c^{\tt U}}(\alpha) &=& u_{c^{\tt U}}+(1-u_{c^{\tt U}})G_{c^{\tt U}/\sigma}(\alpha) \\
&\leq&
G_{c^{\tt U}/\sigma}(\frac{\max\{0,\alpha -u_{c^{\tt U}}\} }{1-u_{c^{\tt U}}}) \\
&=&
\min \{1, G_{c^{\tt U}/\sigma}(\frac{\alpha -u_{c^{\tt U}} }{1-u_{c^{\tt U}}})\}
=\hat{f}^{\tt U}_{c^{\tt U}}{}^{-1}(\alpha),
\end{eqnarray*}
where the inequality follows from the following observation: Since $G_{c^{\tt U}/\sigma}$ is symmetric, 
the curve $y=G_{c^{\tt U}/\sigma}(\frac{x -u_{c^{\tt U}} }{1-u_{c^{\tt U}}})$ for $x\in [u_{c^{\tt U}} ,1]$ is the same as the curve $x=u_{c^{\tt U}}+(1-u_{c^{\tt U}})G_{c^{\tt U}/\sigma}(y)$ for $y\in [0,1]$, hence, it is equal to the curve $y=u_{c^{\tt U}}+(1-u_{c^{\tt U}})G_{c^{\tt U}/\sigma}(x)$ mirrored around the diagonal $y=x$. The inequality follows from the fact that above the diagonal the mirrored curve is larger than the curve itself (due to the addition of $u$).

We conclude that for $\alpha\in [0,\beta]$,
$$ \min\{ \hat{f}^{\tt U}_{c^{\tt U}}, \hat{f}^{\tt U}_{c^{\tt U}}{}^{-1} \} (\alpha) = u_{c^{\tt U}}+(1-u_{c^{\tt U}})G_{c^{\tt U}/\sigma}(\alpha)
$$
and  for $\alpha\in [\beta,1]$,
$$ \min\{ \hat{f}^{\tt U}_{c^{\tt U}}, \hat{f}^{\tt U}_{c^{\tt U}}{}^{-1} \} (\alpha) = G_{c^{\tt U}/\sigma}(\frac{\alpha -u_{c^{\tt U}} }{1-u_{c^{\tt U}}}).
$$

Convexification by means of the double conjugate computes $\beta_0\leq \beta \leq \beta_1$ such that $\bar{f}^{\tt U}_{c^{\tt U}}$ defined as $\min\{ \hat{f}^{\tt U}_{c^{\tt U}}, \hat{f}^{\tt U}_{c^{\tt U}}{}^{-1} \} (\alpha)$ has derivative $-1$ for $\alpha=\beta_0$ and $\alpha=\beta_1$. Next the upper bound is improved by drawing the line from $(\beta_0,\bar{f}^{\tt U}_{c^{\tt U}}(\beta_0))$ to $(\beta_1,\bar{f}^{\tt U}_{c^{\tt U}}(\beta_1))$.
See (\ref{eqder}), $G'_\mu(x)=
- e^{\mu \Phi^{-1}(1-x) -\mu^2/2}$. Let $\mu = c^{\tt U}/\sigma$. This gives
\begin{eqnarray*}
-1&=&-(1-u_{c^{\tt U}})e^{\mu \Phi^{-1}(1-\beta_0) -\mu^2/2} \\
-1&=&-e^{\mu \Phi^{-1}(1-\frac{\beta_1 -u_{c^{\tt U}} }{1-u_{c^{\tt U}}}) -\mu^2/2}/(1-u_{c^{\tt U}})
\end{eqnarray*}
with the solutions
\begin{eqnarray*}
\beta_0 &=&
1- \Phi( \frac{\mu}{2}-\frac{1}{\mu} \ln(1-u_{c^{\tt U}})), \\
\beta_1 &=& u_{c^{\tt U}} + (1-u_{c^{\tt U}})\cdot (1- \Phi(\frac{\mu}{2}+\frac{1}{\mu} \ln(1-u_{c^{\tt U}}))).
\end{eqnarray*}
Notice that
\begin{eqnarray*}
\bar{f}^{\tt U}_{c^{\tt U}}(\beta_0) &=&
u_{c^{\tt U}}+(1-u_{c^{\tt U}})G_{c^{\tt U}/\sigma}(\beta_0)
= u_{c^{\tt U}}+(1-u_{c^{\tt U}})\Phi(-\frac{\mu}{2}-\frac{1}{\mu} \ln(1-u_{c^{\tt U}}))=\beta_1, \\
\bar{f}^{\tt U}_{c^{\tt U}}(\beta_1) &=&
G_{c^{\tt U}/\sigma}(\frac{\beta_1 -u_{c^{\tt U}} }{1-u_{c^{\tt U}}})=\beta_0.
\end{eqnarray*}
This proves 
$$
\min\{ \hat{f}^{\tt U}_{c^{\tt U}}, \hat{f}^{\tt U}_{c^{\tt U}}{}^{-1} \}^{**} (\alpha) =
\left\{
\begin{array}{ll}
u_{c^{\tt U}}+(1-u_{c^{\tt U}})G_{c^{\tt U}/\sigma}(\alpha)     & \mbox{ for } \alpha \in [0,\beta_0], \\
\beta_0+\beta_1-\alpha     &  \mbox{ for } \alpha\in [\beta_0,\beta_1], \\
G_{c^{\tt U}/\sigma}(\frac{\alpha -u_{c^{\tt U}} }{1-u_{c^{\tt U}}})   & \mbox{ for }  \alpha \in [\beta_1,1].
\end{array}
\right.
$$





\subsection{Multiple Epochs}

%
For a single epoch, we can use the equivalent formulation
$$
 L(\{S_{b,h}\}_{h=1}^m) = |\{ 1\leq h\leq m \ : \ S_{b,h} \cap \{1,\ldots, g\} \neq \emptyset \}|$$
 with
$$
q(c) =
{\tt Pr}_\pi\left[ \left. c^2=\sum_{k=1}^g \# \{ b : L(\{\pi(S_{b,h})\}_{h=1}^m)=k\} \cdot k^2 \ \right| \ \{S_{b,h}\}_{b=1,h=1}^{N/(sm),m} \leftarrow {\tt Sample}_{s,m}\right],
$$
where $\pi$ is a random permutation of $\{1,\ldots, N\}$. This formulation is independent of the actual data set $d$ and makes computing $q(c)$ a combinatorics problem. 

All our analysis can be generalized to multiple epochs. Of course, we can use the composition tensor. But an alternative is to take equation (\ref{eqG}) and have the $c_i$ do their counting over multiple epochs. This leads to the more general definition
$$
q_E(c) =
{\tt Pr}_{\{\pi^e\}}\left[ \left. c^2=\sum_{k=1}^g \# \{ (b,e) : L(\{\pi^e(S^e_{b,h})\}_{h=1}^m)=k\} \cdot k^2 \ \right| \ \{ \{S^e_{b,h}\}_{b=1,h=1}^{N/(sm),m} \leftarrow {\tt Sample}_{s,m} \}_{e=1}^E \right]
$$
and all derivations continue as before. We have summarized our results in Theorems \ref{thm:main} and \ref{thm:main2} in the main body.


\section{Probabilistic Filtered Batches: Towards Another Factor 2 Improvement}
\label{app:factor}

Our algorithmic framework uses static $m$, $s$, and $v$ leading to sets $S_b$ with size $ms$ for each round $b$. We propose a probabilistic filtering of these sets $S_b$. This will give new sets $S'_b$ with sizes $|S'_b|$ coming from some probability distribution, rather than being constant. As we will see, this may amplify the resulting differential privacy by a factor $2$ in a slightly less strong adversarial model.

The main idea is to precompute all $S_{b,h}$ and keep each of the subsets with probability $1/2$. This leads to variable sized subsets $$S_b=\bigcup_{h\in {\cal F}_b} S_{b,h},$$
where ${\cal F}_b\subseteq \{1,\ldots,m\}$ with ${\tt Pr}[h\in {\cal F}_b]=1/2$. Rather than implementing a for loop over $h\in \{1,\ldots, m\}$, we implement a for loop over $h\in {\cal F}_b$ in Algorithm \ref{alg:genDPSGD}. We also replace the for loop over $b\in \{1,\ldots, \frac{N}{ms}\}$ by a for loop over
$$ b\in \{ j\in \{1,\ldots, \frac{N}{ms}\} \ | \ |{\cal F}_b|\geq \tau \}
$$
for some threshold $\tau$. For example, when using batch clipping with $m=1$, we only want to keep the rounds with non-empty updates, hence, we set $\tau=1$ (in this case we want to make sure that the stream of updates transmitted to the server correspond to transmission times that follow a memoryless distribution, otherwise the observer can figure out which rounds discard their update). 

Notice that the noised update has the form
$$ \bar{U}_b= \sum_{h\in {\cal F}_b} [a_h]_C + {\cal N}(0,(2C\sigma)^2 {\bf I}) \mbox{ with } |{\cal F}_b|\geq \tau.
$$
(We may decide to divide by $|{\cal F}_b|$ rather than $\mathbb{E}[|{\cal F}_b|]=m/2$ before transmitting to the server.)

In the proof of Theorems \ref{thm:main} and \ref{thm:main2} in Appendix \ref{app:gauss} we compute a sensitivity $2kC$ coming from a bound 
$\|U_b-U'_b\|\leq 2kC$.
Here, the strong adversary replays algorithm ${\cal A}$ for the same randomness ${\sf rand}$ and the same non-differentiating samples in $d\cap d'$ indexed by subsets $S_{b,h}$. Those subsets $S_{b_h}$ that do not include differentiating samples lead to cancellation of terms in the sums that define $U_b$ and $U'_b$. We only keep the $k$ terms in $U_b$ and the $k$ terms in $U'_b$ based on the $S_{b,h}$ that do contain differentiating samples. This leads to the $2kC$ upper bound.




When we include the proposed filtering ${\cal F}_b$, we do not maintain the strong adversarial model which would provide the adversary with the knowledge of all the used coin flips that led to ${\cal F}_b$. We only give the adversary knowledge of the subset ${\cal S}$ of indices $h\in {\cal F}_b$ for which $S_{b,h}$ does not contain any differentiating samples. This means that the adversary only knows set ${\cal S}$ with the knowledge that there exists a set ${\cal F}_b$ 
such that ${\cal S}\subseteq {\cal F}_b$
and ${\cal F}_b\setminus {\cal S}$ corresponds to indices $h$ for which $S_{b,h}$  contains  differentiating sample(s). 

The adversary knows that if $k$ subsets $S_{b,h}$ have differentiating samples, then the size $|{\cal F}_b \setminus {\cal S}|=k'$ with probability ${k \choose k'}/2^k$. Only the terms related to the $k'$ indices in ${\cal F}_b \setminus {\cal S}$ are kept in the sums of $U_b$ and $U'_b$: We have $\|U_b-U'_b\|\leq 2k'C$ with probability ${k \choose k'}/2^k$. In expectation the norm $\|U_b-U'_b\|$ is bounded by $2(k/2)C=kC$.

After adding Gaussian noise, the adversary projects the observed noised local update $\bar{U}_b$ onto the line that connects $U_b$ and $U'_b$ and after normalizing by dividing by $2C$, we have that differentiating whether $\bar{U}_b$ corresponds to $d$ or $d'$ is equivalent to differentiating whether a received output is from ${\cal N}(0,\sigma^2)$ or from ${\cal N}(k',\sigma^2)$ where ${\tt Pr}[k']={k \choose k'}/2^k$ for $0\leq k'\leq k$. Or equivalently, from ${\cal N}(0,1)$ or from ${\cal N}(k'/\sigma,1)$ where ${\tt Pr}[k']={k \choose k'}/2^k$. 

The latter corresponds to a round mechanism ${\cal M}$ composed of submechanisms ${\cal M}_{k'}$:
$$ {\cal M} = \sum_{k'=0}^k ({k \choose k'}/2^k) \cdot {\cal M}_{k'},$$
where ${\cal M}_{k'}$ is $G_{k'/\sigma}$-DP. We can use the lemmas in Appendix \ref{sec:lemmas} to analyze bounds on the DP guarantee for ${\cal M}$, however, this will not lead to a tight expression. 

A better approach is to simply transform distribution $q(c_1,\ldots,c_g)$ as defined in (\ref{eqq}). If we have $c_k$ rounds each with $k$ differentiating samples, then the binomial distribution ${\tt Pr}[k']={k \choose k'}/2^k$ redistributes these $c_k$ rounds among the $c_1$, $c_2$, \ldots, $c_k$. The transformed $q(c_1,\ldots,c_g)$ can be used to compute $q_E(c)$ and the analysis leading to Table \ref{tab:DP} can be repeated. 

In expectation, the sensitivity is $kC$ and the adversary differentiates between whether a received output is from ${\cal N}(0,\sigma^2)$ or from ${\cal N}(k/2,\sigma^2)$ (since $\mathbb{E}[k']=k/2$) and we have the Gaussian trade-off function $G_{k/(2\sigma)}$. This seems to indicate a factor 2 improvement (at best); it is as if $\sigma$ is increased by a factor 2. We leave a detailed analysis for future work. 


For individual clipping with subsampling  for $g=1$, the package 
\cite{opacus} actually implements subsampling using its own filtering technique: For each round $b$, $S_b$ is populated with samples drawn from a data set of size $N$ where each sample has probability $m/N$ of being included in $S_b$. This yields a distribution where $|S_b|$ is equal to $m$ in expectation. The adversary does not know whether the differentiating sample is included in $S_b$ or not and this leads to differentiating between $U$ computed on a set $S_b$ and $U'$ computed in a set $S_b$ minus the differentiating sample. This gives the bound $\|U-U'\|\leq C$ rather than $2C$ and we have the factor 2 improvement. This corresponds to DP-SGD \cite{abadi2016deep} which adds ${\cal N}(0,(C\sigma)^2 {\bf I})$ noise without the factor 2 -- so the DP analysis in \cite{abadi2016deep}  is compatible with the probabilistic filtering implemented in Opacus but misses out on a factor 2 penalty for the original subsampling discussed in \cite{abadi2016deep} and further analysed in this paper. Also, notice that for $g\geq 1$, it is not immediately clear how to prove the factor $2$ improvement  as we will still want to see a $\sqrt{g}$ dependency in the resulting DP guarantee (although the above intuition seems to also hold for general $g$).


In the above argument, we  {\em assume} that the adversary does not learn the actually used mini-batch size otherwise we will again need the factor $2$ in our analysis. For large $m$ and $N$, it seems reasonable to assume that the adversary cannot gain significant knowledge about the used mini-batch size from an observed scaled noised update $\bar{U}/m$ (which includes noise ${\cal N}(0,(2C\sigma/m){\bf I})$ as a function of the mini-batch size $m$). Nevertheless, in the Opacus setting the strong adversary in the DP analysis is made slightly weaker.


In this paper we stay on the safe side where we do not attempt a factor 2 improvement by means of probabilistic filtering. This is consistent with the $f$-DP framework introduced in \cite{dong2021gaussian} which avoids discussing a factor 2 improvement.

\section{Subsampling}
\label{sec:subsampling}

In this section we use our main theorems to prove several lemmas which together can be summarized by the next theorem.

\begin{theorem}[{\bf Subsampling}]
Let $h$ be a trade-off function such  that the more general formula (\ref{eq:gen}) with subsampling in Algorithm \ref{alg:genDPSGD}  yields a mechanism ${\cal M}$ which is $h$-DP 
for all pairs of data sets $d$ and $d'$ 
with $\max \{ |d\setminus d'|, |d'\setminus d|\}\leq g$ and $N=|d\cap d'|+g$.

{\bf [Case $g=1$:]} If we restrict ourselves to $g=1$, i.e., $d$ and $d'$ are neighboring data sets, then there exists an $h$ such that ${\cal M}$ is $h$-DP and
satisfies the  lower  bound:
$$G_{\sqrt{(1+1/\sqrt{2E})E}/\sigma}(\min\{1,\alpha+e^{-E}\}) \leq h(\alpha).$$
Furthermore, for all $h$ with the property that ${\cal M}$ is $h$-DP, we have the  upper bound
$$h(\alpha)\leq e^{-E} + (1-e^{-E}) G_{\sqrt{(1-1/\sqrt{2E})E}/\sigma}(\alpha).$$


{\bf [General $g$, individual clipping:]} Let 
$$c^{\tt L}=\sqrt{\beta \min\{m,g\} gE}$$
with $\beta=e^{N/(N-g-m)}+\gamma
\approx e+\gamma$
for some $\gamma>0$, and define 
$$l_{c^{\tt L}}= e^{-\gamma gE}.$$ Then, for
individual clipping (\ref{indclip}) and general $g\geq 1$,
mechanism ${\cal M}$ is $h$-DP for 
$h=f^{\tt L}_{c^{\tt L}}$ with 
$$h(\alpha)= f^{\tt L}_{c^{\tt L}}(\alpha)\geq  \hat{f}^{\tt L}_{c^{\tt L}}(\alpha)=G_{c^{\tt L}/\sigma}(\min \{1,\alpha+l_{c^{\tt L}}\})=
G_{\sqrt{\beta \min\{m,g\}gE}/\sigma}(\min \{1,\alpha+e^{-\gamma gE}\}).$$ 

{\bf [General $g$, batch clipping:]}
Let 
$$ q = \frac{N}{s} \cdot (1-{N-g \choose s}/{N \choose s}) \approx g$$
for small $s/N\ll 1$. Then, for
batch clipping (\ref{batchclip}) and general $g\geq 1$, 
mechanism ${\cal M}$ is $h$-DP for some trade-off function $h$ that
satisfies the  lower  bound
$$ G_{\sqrt{(1+1/\sqrt{2qE})qE}/\sigma}(\min\{1,\alpha+e^{-qE}\}) \leq h(\alpha).$$
Furthermore, for all $h$ with the property that ${\cal M}$ is $h$-DP, we have the  upper bound
$$ h(\alpha) \leq  e^{-qE} + (1-e^{-qE}) G_{\sqrt{(1-1/\sqrt{2qE})qE}/\sigma}(\alpha).$$
\end{theorem}

\subsection{Complexity of $q_E(c)$}

%
We first consider individual clipping with subsampling, that is, $s=v=1$. Since $s=v=1$, each set $S^e_{b,h}$ is a singleton set and the union $S^e_b=\bigcup_{h=1}^m S^e_{b,h}$ is a set of size $m$. We have
$$L(\{\pi^e(S^e_{b,h})\}_{h=1}^m)=|\pi^e(S^e_b)\cap \{1,\ldots, g\}|.$$
Due to the random permutation $\pi^e$,  $\pi^e(S^e_{b})$ is randomly subsampled from the set of indices $\{1,\ldots, N\}$. Therefore, for $k\leq m$,
$$ 
{\tt Pr}[L(\{\pi^e(S^e_{b,h})\}_{h=1}^m)=k]=
{N-g \choose m-k}{g \choose k}/{N \choose m}.
$$
We denote this probability by $q_k$.
For $m<k\leq g$, it is straightforward to see that ${\tt Pr}[L(\{\pi^e(S^e_{b,h})\}_{h=1}^m)=k]=0$ and we define $q_k=0$ for $m<k\leq g$. 

Given this notation, the joint probability of having 
$$\{c_k=\# \{ (b,e) : L(\{\pi^e(S^e_{b,h})\}_{h=1}^m)=k\}\}_{k=1}^g$$
is equal to
\begin{equation} q_E(c_1,\ldots, c_g) = {(N/m)\cdot E \choose c_1,c_2,\ldots, c_g} (1-\sum_{k=1}^g q_k)^{(N/m)\cdot E - \sum_{k=1}^g c_k}\prod_{k=1}^g q_k^{c_k}.\label{complex}
\end{equation}
We can use these probabilities in the definition of trade-off function $f$, but this becomes too complex. Even the upper and lower bounds will require some fine tuned calculus in order to get tight expressions.

For the general setting of formula (\ref{eq:gen}) we need to consider $s>1$. This means that the probabilities $q_k$ become more complex. Now $ms$ samples are selected out of the data set with $N$ samples and grouped in $m$ subsets $S^e_{b,h}$  of size $s$. The total number of possibilities is
$${N \choose ms} {ms \choose s, s, \ldots, s}.
$$
Suppose that
$k'$ of the $ms$ samples belong to the $g$  differentiating samples -- there are 
$${N-g \choose ms-k'}{g \choose k'}$$
combinations. Suppose that $k'_h$ differentiating samples are put in the subset $S^e_{b,h}$ of size $s$. Then, $k'=\sum_{h=1}^m k'_h$ with $0\leq k'_h\leq s$.
This gives
$$ {k' \choose k'_1, \ldots, k'_m}
$$
combinations. Notice that $s-h'_h$ non-differentiating samples are put in $S^e_{b,h}$. This gives
$${ms-k' \choose (s-k'_1), \ldots, (s-k'_m)}
$$
combinations. If $k=\# \{ h \ | \ k_h\neq 0\}$, then we contribute to $q_k$. Therefore, let 
$$ {\cal K}_k=\left\{ (k'_1,\ldots, k'_m) \ | \ k=\# \{ h \ | \ k_h\neq 0\} \mbox{ and }  \forall_{h=1}^m 0\leq k'_h\leq s  \right\}.
$$

Combining all previous expressions yields
$$
q_k = \frac{\sum_{(k'_1,\ldots,k'_m)\in {\cal K}_k} {N-g \choose ms-\sum_{h=1}^m k'_h}{g \choose \sum_{h=1}^m k'_h}  {\sum_{h=1}^m k'_h \choose k'_1, \ldots, k'_m} {ms-\sum_{h=1}^m k'_h \choose (s-k'_1), \ldots, (s-k'_m)}}{ {N \choose ms} {ms \choose s, s, \ldots, s}},
$$
a considerably more complex expression for $q_k$ to work with. 



\subsection{Individual Clipping for $g=1$}

We start by analyzing individual clipping for the simplest case $g=1$. We will show that we can derive simple to interpret tight formulas that 
resemble the Gaussian trade-off function $G_{\sqrt{E}/\sigma}$. In the case $g=1$ we only need to consider $q_1=m/N$. In $q_E(c_1,\ldots,c_g)$ notation this gives $$q_E(c_1)={(N/m)\cdot E \choose c_1}(1-m/N)^{(N/m)\cdot E-c_1}(m/N)^{c_1}.$$
Notice that the theorem uses $q_E(c)$ which equals the above $q_E(c_1)$ evaluated in $c_1=c^2$, that is, $c=\sqrt{c_1}$.
From Hoeffding's inequality we obtain
\begin{eqnarray*}
\sum_{c'_1\leq c_1} q_E(c'_1) &\leq & \exp(-2 (E-c_1)^2)
\mbox{ for } c_1\leq E, \\
\sum_{c'_1\geq c_1} q_E(c'_1) &\leq & \exp(-2 (E-c_1)^2)
\mbox{ for } c_1\geq E.
\end{eqnarray*}
In the upper and lower bounds of Theorem \ref{thm:main} we choose $(c^{\tt U})^2=(1-1/\sqrt{2E})E$ which gives $u_{c^{\tt U}}= e^{-E}$ as upper bound on the first sum and we choose $(c^{\tt L})^2=(1+1/\sqrt{2E})E$ which gives $l_{c^{\tt L}}= e^{-E}$ as upper bound on the second sum. 

From this we conclude that
\begin{eqnarray*}
\hat{f}^{\tt U}_{c^{\tt U}}(\alpha) &= & e^{-E} + (1-e^{-E}) G_{\sqrt{(1-1/\sqrt{2E})E}/\sigma}(\alpha), \\
\hat{f}^{\tt L}_{c^{\tt L}}(\alpha) &=& G_{\sqrt{(1+1/\sqrt{2E})E}/\sigma}(\min\{1,\alpha+e^{-E}\}).
\end{eqnarray*}
For larger $E$, the upper and lower bound converge to $G_{\sqrt{E}/\sigma}(\alpha)$. 
We recall that for the strong adversary $h=C_{m/N}(G_{1/\sigma})^{\otimes (N/m)\cdot E}$ is a tight trade-off function for ${\cal M}$ with $g=1$.





We notice that the above analysis for $g=1$ also holds for the mechanism based on the general formula  (\ref{eq:gen}) (by replacing $m$ with $sm$).



\begin{lemma} \label{lem:indsub}
Let $h$ be a trade-off function such that individual clipping (\ref{indclip}) with subsampling in Algorithm \ref{alg:genDPSGD} (i.e., DP-SGD) yields a mechanism ${\cal M}$ which is 
$h$-DP 
for all pairs of data sets $d$ and $d'$ 
with $\max \{ |d\setminus d'|, |d'\setminus d|\}\leq 1$ and $N=|d\cap d'|+1$ (we have $g=1$). 
Then $h$ satisfies
\begin{eqnarray*}
&& G_{\sqrt{(1+1/\sqrt{2E})E}/\sigma}(\min\{1,\alpha+e^{-E}\}) 
\leq 
h(\alpha)
\leq  e^{-E} + (1-e^{-E}) G_{\sqrt{(1-1/\sqrt{2E})E}/\sigma}(\alpha).
\end{eqnarray*}
\end{lemma}



The lemma can be improved by using the lower and upper bounds derived from the symmetric trade-off functions $f^{\tt L}_{c^{\tt L}}$ and $f^{\tt U}_{c^{\tt U}}$. However, the weaker bounds in the current lemma already capture intuitions stated in the next corollaries. 
The latter property comes from the fact that in expectation a single epoch will only leak privacy related to one $G_{1/\sigma}$ instance. This is because there are $N/m$ rounds and each round has probability $m/N$ to leak privacy according to $G_{1/\sigma}$ (subsampling with probability $m/N$ and composition over $N/m$ rounds cancels one another). Composition over $E$ epochs will lead to $G_{1/\sigma}^{\otimes E}=G_{\sqrt{E}/\sigma}$ in expectation. The resulting independence of the total number of rounds gives the engineer the freedom to tune parameter $m$ for achieving the best accuracy.

\subsection{General Clipping for $g=1$}

%
We notice that the analysis leading to Lemma \ref{lem:indsub} for $g=1$ also holds for the mechanism based on the general formula  (\ref{eq:gen}) (by replacing $m$ with $sm$):

\begin{lemma} \label{lem:genclipI}
The more general formula (\ref{eq:gen}) with subsampling in Algorithm \ref{alg:genDPSGD}  yields a mechanism ${\cal M}$ which is $h$-DP 
for all pairs of data sets $d$ and $d'$ 
with $\max \{ |d\setminus d'|, |d'\setminus d|\}\leq 1$ and $N=|d\cap d'|+1$ (we have $g=1$) for some trade-off function $h$ satisfying the  lower  bound:
$$G_{\sqrt{(1+1/\sqrt{2E})E}/\sigma}(\min\{1,\alpha+e^{-E}\}) \leq h(\alpha).$$
Furthermore, for all $\bar{h}$ with the property that ${\cal M}$ is $\bar{h}$-DP, we have the  upper bound
$$\bar{h}(\alpha)\leq e^{-E} + (1-e^{-E}) G_{\sqrt{(1-1/\sqrt{2E})E}/\sigma}(\alpha).$$
This shows into what extent the lower bound is tight.
\end{lemma}


\subsection{Lower Bound for Individual Clipping for $g$}

In order to get a 
lower bound on the trade-off function for individual clipping ($s=v=1$) with subsampling that holds for arbitrary $g$, we derive and use upper bounds on the $q_k$ (this puts more weight towards larger $k$, hence, larger $c$ in $q_E(c)$, which in turn favors the smaller Gaussian trade-off functions $G_{c/\sigma}$). 

We first consider $g>m$. 
Notice that $q_k=0$ for $m<k\leq g$. In the worst-case we see scaling with $m$ in a single round leading to trade-off function $G_{m/\sigma}$ (see the group privacy discussion in Section \ref{sec:GDP}). 
In expectation there are at most $g$ rounds that have differentiating samples. Therefore, a coarse worst-case analysis should lead to a lower bound of $G_{m/\sigma}^{\otimes g}$ for the trade-off function for a single epoch.
composition over $E$ epochs gives the lower bound trade-off function $G_{m\sqrt{g}\sqrt{E}/\sigma}$. As we will see, a precise analysis can improve this to a $\sqrt{mg}$ rather than a $m\sqrt{g}$ dependency:


For individual clipping we have
\begin{eqnarray*}
q_{k+1} &=& {N-g \choose m-k-1}{g \choose k+1}/{N \choose m}  \\
&=&
{N-g \choose m-k}\frac{m-k}{N-g-m+k+1}{g \choose k}\frac{g-k}{k+1}/{N \choose m} \\
&=& q_k \frac{m-k}{N-g-m+k+1}\frac{g-k}{k+1}  \\
&\leq& q_k \frac{g}{N-g-m}\frac{m-k}{k+1}.
\end{eqnarray*}
and
\begin{eqnarray*}
q_1 &=& {N-g \choose m-1}{g \choose 1}/{N \choose m} \\
&=& \frac{gm}{N}{N-g \choose m-1}/{N-1 \choose m-1} \leq  \frac{gm}{N}\leq \frac{g}{N-g-m}\frac{m}{1}.
\end{eqnarray*}
By induction in $k$,
$$ q_k \leq (\frac{g}{N-g-m} )^k {m \choose k}.$$
We substitute this in (\ref{complex}) and obtain
\begin{eqnarray*}
q_E(c_1,\ldots, c_m, 0, ...,0) 
&\leq& {(N/m)\cdot E \choose c_1,c_2,\ldots, c_m} \prod_{k=1}^m q_k^{c_k} \\
&\leq & {(N/m)\cdot E \choose c_1,c_2,\ldots, c_m} (\frac{g}{N-g-m} )^{\sum_{k=1}^m c_k\cdot k} \prod_{k=1}^m {m \choose k}^{c_k}.
\end{eqnarray*}

Let
$$c^{\tt L}=\sqrt{\beta mgE}
\mbox{ with } \beta=e^{N/(N-g-m)}+\gamma
\approx e+\gamma,$$
for some $\gamma>0$.
For $c=\sqrt{\sum_{k=1}^m c_k\cdot k^2}> c^{\tt L}$, we have
$$
\sum_{k=1}^m c_k\cdot k \geq \sum_{k=1}^m c_k\cdot k^2/m\geq \beta mgE/m=
\beta gE.
$$
Therefore, if $c>c^{\tt L}$, then
\begin{eqnarray*}
(\frac{g}{N-g-m} )^{\sum_{k=1}^m c_k\cdot k} &=&
(\frac{g\beta }{(N-g-m)} )^{\sum_{k=1}^m c_k\cdot k} (1/\beta)^{\sum_{k=1}^m c_k\cdot k} \\
&\leq &
(\frac{g\beta }{(N-g-m)} )^{\sum_{k=1}^m c_k\cdot k}
(1/\beta)^{\beta gE}.
\end{eqnarray*}
This proves that
\begin{eqnarray*}
&& \sum_{c_1,c_2,\ldots c_m : c=\sqrt{\sum_{k=1}^m c_k\cdot k^2}> c^{\tt L}} q_E(c_1,\ldots, c_m, 0, ...,0) \\
&\leq& \sum_{c_1,c_2,\ldots, c_m} {(N/m)\cdot E \choose c_1,c_2,\ldots, c_m} (\frac{g\beta }{(N-g-m)} )^{\sum_{k=1}^m c_k\cdot k} (1/\beta)^{\beta gE} \prod_{k=1}^m {m \choose k}^{c_k} \\
&=&
(1/\beta)^{\beta gE}  \sum_{c_1,c_2,\ldots, c_m} {(N/m)\cdot E \choose c_1,c_2,\ldots, c_m}  \prod_{k=1}^m ({m \choose k}(\frac{g\beta }{(N-g-m)} )^k)^{c_k} \\
&=&
(1/\beta)^{\beta gE}  (1+ \sum_{k=1}^m {m \choose k} (\frac{g\beta }{(N-g-m)} )^k)^{(N/m)\cdot E} \\
&=& (1/\beta)^{\beta gE} ((1+\frac{g\beta}{(N-g-m)})^m)^{(N/m)\cdot E} \\
&=& (1/\beta)^{\beta gE} (1+\frac{g\beta}{(N-g-m)})^{NE} \\
&\leq &  (1/\beta)^{\beta gE} e^{(g\beta) (N/(N-g-m)) E} \\
&=& ((e^{(N/(N-g-m)}/\beta)^{\beta gE}
= (1-\gamma/\beta)^{\beta gE}
\leq e^{-\gamma gE},
\end{eqnarray*}
where the two last inequalities follow from $(1+x/y)^y\leq e^x$ for all real valued $x>-y$ (by definition $-\gamma >-\beta$).

Notice that the above analysis can be generalized to $g\leq m$: We substitute $m$ by $g$ in the definition of probability $q_E(\dots)$,
$$ q_E(c_1,\ldots, c_g) 
\leq {(N/m)\cdot E \choose c_1,c_2,\ldots, c_g} \prod_{k=1}^g q_k^{c_k},
$$
and we use $c^{\tt L}=\sqrt{\beta g^2 E}$ such that we can repeat the argument $\sum_{k=1}^g c_k\cdot k \geq \sum_{k=1}^g c_k\cdot k^2/g\geq \beta g^2 E/g=\beta gE$ as before. We obtain a
 linear $g$ dependency, which is also what the group privacy analysis with operator $\circ$ would give us. It remains an open problem to improve the analysis for $g\leq m$.
 

Application of Theorem \ref{thm:main2} gives the following lemma:

\begin{lemma} \label{lem:genlow}
Let 
$$c^{\tt L}=\sqrt{\beta \min\{m,g\} gE}$$
with $\beta=e^{N/(N-g-m)}+\gamma
\approx e+\gamma$
for some $\gamma>0$, and define 
$$l_{c^{\tt L}}= e^{-\gamma gE}.$$ Then,
individual clipping (\ref{indclip}) with subsampling in Algorithm \ref{alg:genDPSGD} (i.e., DP-SGD) yields a mechanism ${\cal M}$ which is $h$-DP for 
$h=f^{\tt L}_{c^{\tt L}}$ with 
$$h(\alpha)= f^{\tt L}_{c^{\tt L}}(\alpha)\geq  \hat{f}^{\tt L}_{c^{\tt L}}(\alpha)=G_{c^{\tt L}/\sigma}(\min \{1,\alpha+l_{c^{\tt L}}\})=
G_{\sqrt{\beta \min\{m,g\}gE}/\sigma}(\min \{1,\alpha+e^{-\gamma gE}\})$$ 
for all pairs of data sets $d$ and $d'$ 
with $\max \{ |d\setminus d'|, |d'\setminus d|\}\leq g$ and $N=|d\cap d'|+g$.
\end{lemma}

\subsection{Batch Clipping for $g$}

%
We consider batch clipping with $m=1$ and $v=1$. This gives $N/s$ rounds within an epoch and each round computes on a batch of $s$ data samples. We have 
$$L(\{\pi^e(S^e_{b,h})\}_{h=1}^1)= \left\{ 
\begin{array}{ll}
1, & \mbox{ if } |\pi^e(S^e_b)\cap \{1,\ldots, g\}|\neq 0, \\
0, & \mbox{ if } |\pi^e(S^e_b)\cap \{1,\ldots, g\}|= 0.
\end{array} 
\right.
$$
We have
\begin{eqnarray*} 
{\tt Pr}[L(\{\pi^e(S^e_{b,h})\}_{h=1}^1)=1] &=& 1- 
{N-g \choose s}/{N \choose s}
=1-\prod_{j=0}^{g-1} \frac{N-s-j}{N-j} \\
&\approx& 1-(1-s/N)^g \approx gs/N.
\end{eqnarray*}
We denote this probability by $q_1$ and define $q_2=\ldots=q_g=0$.
Given this notation, the  probability of having 
$$c_1=\# \{ (b,e) : L(\{\pi^e(S^e_{b,h})\}_{h=1}^1)=1\}$$
is equal to
$$ q_E(c_1,c_2=0,\ldots, c_g=0) = {(N/s)\cdot E \choose c_1} (1-q_1)^{(N/s)\cdot E -  c_1} q_1^{c_1}.
$$
We repeat the same type of calculus as before. From Hoeffding's inequality we obtain
\begin{eqnarray*}
\sum_{c'_1\leq c_1} q_E(c'_1,0,\ldots,0) &\leq & \exp(-2 ((N/s)E q_1-c_1)^2)
\mbox{ for } c_1\leq (N/s)E q_1, \\
\sum_{c'_1\geq c_1} q_E(c'_1,0,\ldots,0) &\leq & \exp(-2 ((N/s)E q_1-c_1)^2)
\mbox{ for } c_1\geq (N/s)E q_1.
\end{eqnarray*}
In the upper and lower bounds of Theorem \ref{thm:main} we choose $(c^{\tt U})^2=(1-1/\sqrt{2(N/s)E q_1})(N/s)E q_1$ which gives $u_{c^{\tt U}}= e^{-(N/s)E q_1}$ and we choose $(c^{\tt L})^2=(1+1/\sqrt{2(N/s)E q_1})(N/s)E q_1$ which gives $l_{c^{\tt L}}= e^{-(N/s)E q_1}$. Notice that $q_1\approx gs/N$ which makes $(N/s)E q_1\approx gE$.






\begin{lemma} \label{lem:batchsub}
Let 
$$ q = \frac{N}{s} \cdot (1-{N-g \choose s}/{N \choose s}) \approx g$$
for small $s/N\ll 1$. Then,
batch clipping with subsampling in Algorithm \ref{alg:genDPSGD} yields a mechanism ${\cal M}$ which is $h$-DP 
for all pairs of data sets $d$ and $d'$ 
with $\max \{ |d\setminus d'|, |d'\setminus d|\}\leq g$ and $N=|d\cap d'|+g$ for some trade-off function $h$ satisfying the  lower  bound:
$$ G_{\sqrt{(1+1/\sqrt{2qE})qE}/\sigma}(\min\{1,\alpha+e^{-qE}\}) \leq h(\alpha).$$
Furthermore, for all $\bar{h}$ with the property that ${\cal M}$ is $\bar{h}$-DP, we have the  upper bound
$$ \bar{h}(\alpha) \leq  e^{-qE} + (1-e^{-qE}) G_{\sqrt{(1-1/\sqrt{2qE})qE}/\sigma}(\alpha).$$
This shows into what extent the lower bound is tight.
\end{lemma}







\section{Shuffling}
\label{sec:shuffling}

In this section we use our main theorems to prove several lemmas which together can be summarized by the next theorem.

\begin{theorem}[{\bf Shuffling}] \label{thm:shuffling}
Let $h$ be a trade-off function such  that the more general formula (\ref{eq:gen}) with shuffling in Algorithm \ref{alg:genDPSGD}  yields a mechanism ${\cal M}$ which is $h$-DP 
for all pairs of data sets $d$ and $d'$ 
with $\max \{ |d\setminus d'|, |d'\setminus d|\}\leq g$ and $N=|d\cap d'|+g$.

[{\bf Case $g=1$:}] If we restrict ourselves to $g=1$, i.e., $d$ and $d'$  are neighboring data sets, then ${\cal M}$ is $h$-DP for
$$ h=G_{\sqrt{E}/\sigma}$$
(lower bound). Furthermore, this is tight and cannot be improved (upper bound).

[{\bf Case $g\geq 1$:}]
Let $c^{\tt L}=\sqrt{g}$ and $l_{c^{\tt L}}=g^2ms/(N-g)$.
Then,  mechanism ${\cal M}$ is $h^{\otimes E}$-DP for 
$h=f^{\tt L}_{c^{\tt L}}$ with 
$$h= f^{\tt L}_{c^{\tt L}}\geq  \hat{f}^{\tt L}_{c^{\tt L}}=G_{c^{\tt L}/\sigma}(\min \{1,\alpha+l_{c^{\tt L}}\})=
G_{\sqrt{g}/\sigma}(\min \{1,\alpha+\frac{g^2ms}{N-g}\})$$ 
and
$$h^{\otimes E}= f^{\tt L}_{c^{\tt L}}{}^{\otimes E}\geq 
(\alpha \rightarrow G_{\sqrt{g}/\sigma}(\min \{1,\alpha+\frac{g^2ms}{N-g}\}))^{\otimes E}\approx G_{\sqrt{gE}/\sigma},$$
where the approximation is for constant $E$ and  small $g^2ms/(N-g)$.

[{\bf General $g$, batch clipping:}]
For batch clipping (\ref{batchclip}) and general $g\geq 1$, mechanism ${\cal M}$ is $h$-DP for 
$$ h= G_{\sqrt{gE}/\sigma}
$$
(lower bound). For $g\leq s$, mechanism ${\cal M}$ is $h$-DP for
$h=f^{\otimes E}$ with
$$ h= f^{\otimes E} \geq G_{\sqrt{gE}/\sigma},$$
$f$ is defined as the symmetric trade-off function 
$$ f(\alpha)=\sum_{j=1}^g q_j \cdot \Phi(\Lambda(\alpha)\cdot \frac{\sigma}{\sqrt{j}} - \frac{\sqrt{j}}{2\sigma}) \mbox{ with } 1-\alpha = \sum_{j=1}^g q_j \cdot \Phi(\Lambda(\alpha)\cdot \frac{\sigma}{\sqrt{j}} + \frac{\sqrt{j}}{2\sigma}),$$
where $q_j={N/s \choose j} {g-1 \choose j-1}/ {N/s-1+g \choose g}$.

Furthermore, let 
$c^{\tt U}=\sqrt{g}$ and $u_{c^{\tt U}}=g^2/(N/s-g-g^2)$.
If $g\leq s$, then for all $h$ with the property that ${\cal M}$ is $h$-DP, we have the  upper bound
$$ h \leq  f^{\tt U}_{c^{\tt U}}{}^{\otimes E} \leq \hat{f}^{\tt U}_{c^{\tt U}}{}^{\otimes E},
$$
where
$$ \hat{f}^{\tt U}_{c^{\tt U}}{}^{\otimes E}= (u_{c^{\tt U}} +(1-u_{c^{\tt U}}) G_{c^{\tt U}/\sigma})^{\otimes E}
\approx
G_{\sqrt{g}/\sigma}^{\otimes E} = G_{\sqrt{gE}/\sigma}
$$
for constant $E$ and small $g^2/(N/s-g-g^2)$.
\end{theorem}

\subsection{General Clipping for $g=1$}





Consider a single epoch.
For $g=1$, we have that exactly one round in the epoch has the differentiating sample. We do not even need to use our main theorems and can immediately conclude that the trade-off function is equal to $G_{1/\sigma}$ for each epoch. This is tight for the strong adversary. Composing this over $E$ rounds gives $G_{\sqrt{E}/\sigma}$ and fits Lemma \ref{lem:genclipI}.

\begin{lemma} \label{lem:indshuf}
The more general formula (\ref{eq:gen}) with shuffling in Algorithm \ref{alg:genDPSGD}  yields a mechanism ${\cal M}$ which is $G_{\sqrt{E}/\sigma}$-DP  for all pairs of data sets $d$ and $d'$ 
with $\max \{ |d\setminus d'|, |d'\setminus d|\}\leq 1$ and $N=|d\cap d'|+1$ (we have $g=1$). This is tight.
%
\end{lemma}

\subsection{General clipping for $g\geq 1$}
\label{app:genlowshufQ}


We will now consider the general case $g\geq 1$ for a single epoch $e=1$  and prove a lower bound based on Theorem \ref{thm:main2}. Let $c^{\tt L}=\sqrt{g}$.
Notice that if each of the $g$ differentiating samples are separated by at least $ms-1$ non-differentiating samples in the permutation that defines the shuffling of all data samples used for the epoch, then each subset $\pi^{e}(S_b)$ has at most one differentiating sample.\footnote{We may assume that $S_b=\{(b-1)ms+1,\ldots, bms\}$.}
This means that there are exactly $g$ rounds that each have one differentiating sample, hence, $c_1=g$ and $c_2=\ldots=c_g=0$. 

The probability that each of the $g$ differentiating samples are separated by at least $ms-1$ non-differentiating samples is at least the number of combinations for distributing $N-gms$ non-differentiating samples and $g$ groups of one differentiating sample followed by $ms-1$ non-differentiating samples, divided by the total number of ways $g$ differentiating samples can be distributed among $N$ samples: We have
\begin{eqnarray*}
q_{E=1}(c_1=g,0,\ldots,0) &\geq & {N-gms+g \choose g}/{N \choose g}
\\
&=&\frac{N-gms+g}{N}\cdot \ldots \cdot \frac{N-gms+1}{N-g+1} \\
&\geq& (\frac{N-gms+1}{N-g+1})^g =(1-g(ms-1)/(N-g+1))^g \\
&\geq& 1-g^2(ms-1)/(N-g+1) \geq 1-g^2ms/(N-g).
\end{eqnarray*}
We derive
$$\sum_{c>c^{\tt L}} q_E(c) = 1-\sum_{c\leq c^{\tt L}=\sqrt{ag}} q_E(c)\leq 1-q_{E=1}(c_1=g,0,\ldots,0)\leq g^2ms/(N-g).
$$
This shows that we can apply Theorem \ref{thm:main2} for $c^{\tt L}=\sqrt{g}$ and $l_{c^{\tt L}}=g^2ms/(N-g)$. We obtain the following lemma, where we use composition over $E$ epochs.

\begin{lemma} \label{lem:genlowshufQ}
Let 
$$c^{\tt L}=\sqrt{g} \mbox{ and } l_{c^{\tt L}}=g^2ms/(N-g).$$
Then, the more
general formula (\ref{eq:gen}) with shuffling in Algorithm \ref{alg:genDPSGD} for one epoch $E=1$ yields a mechanism ${\cal M}$ which is $h$-DP for 
$h=f^{\tt L}_{c^{\tt L}}$ with 
$$h= f^{\tt L}_{c^{\tt L}}\geq  \hat{f}^{\tt L}_{c^{\tt L}}=G_{c^{\tt L}/\sigma}(\min \{1,\alpha+l_{c^{\tt L}}\})=
G_{\sqrt{g}/\sigma}(\min \{1,\alpha+\frac{g^2ms}{N-g}\})$$ 
for all pairs of data sets $d$ and $d'$ 
with $\max \{ |d\setminus d'|, |d'\setminus d|\}\leq g$ and $N=|d\cap d'|+g$. By using composition over $E$ epochs this yields the lower bound
$$h^{\otimes E}= f^{\tt L}_{c^{\tt L}}{}^{\otimes E}\geq 
(\alpha \rightarrow G_{\sqrt{g}/\sigma}(\min \{1,\alpha+\frac{g^2ms}{N-g}\}))^{\otimes E}\approx G_{\sqrt{gE}/\sigma},$$
where the approximation is for constant $E$ and small $g^2ms/(N-g)$.
\end{lemma}


\subsection{Batch Clipping for $g$}

We consider a single epoch, $E=1$. Since $m=1$, we only need to consider values $c_1$ (values $c_2=\ldots=c_g=0$). We assume $g\leq s$. Then,
the total number of possible distributions of $g$ differentiating samples over $N/s$ rounds is a ball-in-bins problem and equals ${N/s-1+g \choose g}$ (since each round has a sufficient number of `slots' $s$ to host as many as $g$ balls). 
The number of ways to choose $j$ rounds in which  differentiating samples will be allocated is equal to ${N/s \choose j}$. This allocates already 1 differentiating sample in each of the $j$ rounds; $j$ differentiating samples in total. The remaining $g-j$ samples can be freely distributed over the $j$ rounds and this gives another ${j-1+(g-j) \choose g-j}={g-1 \choose j-1}$  possibilities (again a balls-in-bins problem). For $1\leq j\leq g$, we have
$$q_{E=1}(c_1=j,0,\ldots,0) =
{N/s \choose j} {g-1 \choose j-1}/ {N/s-1+g \choose g}.$$
We will denote this probability by $q_j$.
This allows us to compute trade-off function $f$ in Theorem \ref{thm:main2} by setting
$$q_E(\sqrt{j}) = q_{E=1}(j,0,\ldots,0)=q_j.$$ We have
$$ f(\alpha)=\sum_{j=1}^g q_j \cdot \Phi\left(\Lambda(\alpha)\cdot \frac{\sigma}{\sqrt{j}} - \frac{\sqrt{j}}{2\sigma}\right),$$
where function $\Lambda(\alpha)$ is implicitly defined by
    $$1-\alpha = \sum_{j=1}^g q_j \cdot \Phi\left(\Lambda(\alpha)\cdot \frac{\sigma}{\sqrt{j}} + \frac{\sqrt{j}}{2\sigma}\right).$$
    
Clearly, $f(\alpha)$ is lower bounded by replacing all $\sqrt{j}$ by the larger $\sqrt{g}$, since this always assumes the worst-case where all $g$ differentiating samples are distributed over different round, hence, composition of $G_{1/\sigma}$ happens $g$ times. This gives
$$f \geq G_{\sqrt{g}/\sigma}$$
and over $E$ epochs this implies the lower bounds $f^{\otimes E}\geq G_{\sqrt{g}/\sigma}^{\otimes E}=G_{\sqrt{g E}/\sigma}$. 

Notice that this is exactly the lower bound that follows from Theorem \ref{thm:main2} for $c^{\tt L}=\sqrt{g}$ and $l_{c^{\tt L}}=\sum_{c>c^{\tt L}} q_E(c) =0$.
Also notice that the lower bound holds for $g>s$. This is because the lower bound assumes the worst-case where all $g$ differentiating samples are distributed over different rounds and this is not restricted by the size of a round (we still have $l_{c^{\tt L}}=0$).


If we look at the expression for  $q_j=q_{E=1}(c_1=j,0,\ldots,0)$ more carefully, then we observe that
\begin{eqnarray*}
q_g &=& {N/s \choose g} / {N/s-1+g \choose g} \\
&=&  \frac{N/s \cdot (N/s-1) \cdot \ldots \cdot (N/s-g+1)}{(N/s+g-1) \cdot (N/s-1+g-1) \cdot \ldots \cdot (N/s)} \\ &\geq& (\frac{N/s-g+1}{N/s})^g  =(1-(g-1)/(N/s))^g \approx 1-(g-1)g/(N/s),
\end{eqnarray*}
in other words its probability mass is concentrated in $c_1=g$. This can be used to extract an upper bound from Theorem \ref{thm:main} by setting $c^{\tt U}=\sqrt{g}$ ($=c^{\tt L})$ and upper bound the tail $\sum_{c<c^{\tt U}}q_E(c)$. 

For $g\leq s$ we derive
\begin{eqnarray*} q_{j+1} &=& {N/s \choose j+1} {g-1 \choose j}/ {N/s-1+g \choose g} \\
&=&
{N/s \choose j} \frac{N/s-j}{j+1} {g-1 \choose j-1} \frac{g-j}{j}/ {N/s-1+g \choose g} \\
&=& q_j \frac{N/s-j}{j+1}\frac{g-j}{j}.
\end{eqnarray*}
We have
$$ q_j = q_{j+1} \frac{j+1}{N/s-j}\frac{j}{g-j}\leq 
q_{j+1} \frac{g^2}{N/s-g}.
$$
This shows that for $E=1$,
\begin{eqnarray*}
\sum_{c<c^{\tt U}} q_{E}(c)
&=& \sum_{j=1}^{g-1} q_j
\leq \sum_{j=1}^{g-1} q_g (\frac{g^2}{N/s-g})^{g-j}
\leq \sum_{j=1}^{g-1}  (\frac{g^2}{N/s-g})^{g-j}
= \sum_{j=1}^{g-1}  (\frac{g^2}{N/s-g})^{j} \\
&\leq & \frac{g^2}{N/s-g} / (1- \frac{g^2}{N/s-g}) =\frac{g^2}{N/s-g-g^2}.
\end{eqnarray*}
We conclude that Theorem \ref{thm:main} can be applied for $c^{\tt U}=\sqrt{g}$ and $u_{c^{\tt U}}=g^2/(N/s-g-g^2)$ for $g\leq s$ and $E=1$. This yields an upper bound on the trade-off function for $E=1$ that can be composed $E$ times in order to achieve an upper bound for general $E$.




\begin{lemma} \label{lem:batchshuf}
Batch clipping with shuffling in Algorithm \ref{alg:genDPSGD} yields a mechanism ${\cal M}$ which is $h$-DP 
for all pairs of data sets $d$ and $d'$ with with $\max \{ |d\setminus d'|, |d'\setminus d|\}\leq g$ and $N=|d\cap d'|+g$ 
with 
$$ h= G_{\sqrt{gE}/\sigma}.
$$
For $g\leq s$, mechanism ${\cal M}$ is $h$-DP for
$h=f^{\otimes E}$ with
%
$$ h= f^{\otimes E} \geq G_{\sqrt{gE}/\sigma}$$
for trade-off function $f$ defined by 
$$ f(\alpha)=\sum_{j=1}^g q_j \cdot \Phi(\Lambda(\alpha)\cdot \frac{\sigma}{\sqrt{j}} - \frac{\sqrt{j}}{2\sigma}) \mbox{ with } 1-\alpha = \sum_{j=1}^g q_j \cdot \Phi(\Lambda(\alpha)\cdot \frac{\sigma}{\sqrt{j}} + \frac{\sqrt{j}}{2\sigma}),$$
where $q_j={N/s \choose j} {g-1 \choose j-1}/ {N/s-1+g \choose g}$. 

Furthermore, let 
$c^{\tt U}=\sqrt{g}$ and $u_{c^{\tt U}}=g^2/(N/s-g-g^2)$.
If $g\leq s$, then for all $\bar{h}$ with the property that ${\cal M}$ is $\bar{h}$-DP, we have the  upper bound
$$ \bar{h} \leq  f^{\tt U}_{c^{\tt U}}{}^{\otimes E} \leq \hat{f}^{\tt U}_{c^{\tt U}}{}^{\otimes E},
$$
where
$$ \hat{f}^{\tt U}_{c^{\tt U}}{}^{\otimes E}= (u_{c^{\tt U}} +(1-u_{c^{\tt U}}) G_{c^{\tt U}/\sigma})^{\otimes E}
\approx
G_{\sqrt{g}/\sigma}^{\otimes E} = G_{\sqrt{gE}/\sigma}
$$
for constant $E$ and small $g^2/(N/s-g-g^2)$.
This shows into what extent the lower bound is tight.
\end{lemma}

\section{Strong Adversarial Model} \label{sec:adv}

We assume an adversary who knows the differentiating samples in $d\setminus d'$ and $d'\setminus d$,  but who a-priori (before mechanism ${\cal M}$ is executed) may only know (besides say a 99\% characterization of $d
\cap d'$) an estimate of the number of samples in the intersection of $d$ and $d'$, i.e., the adversary knows $|d\cap d'|+noise$  where the noise is large enough to yield a `sufficiently strong' DP guarantee with respect to the size of the used data set ($d$ or $d'$). Since ${\cal M}$ does not directly reveal the size of the used data set, we assume (as in prior literature) that the effect of $N=|d|\neq N'=|d'|$ contributes at most a very small amount of privacy leakage, sufficiently small to be discarded in our DP analysis: That is, we may as well assume $N=N'$ in our DP analysis.


In this setting of $N=N'$ the  DP analysis here and in prior work  considers an adversary who can mimic mechanism ${\cal M}\circ {\sf Sample}_{s,m}$ in that it can replay into large extent how ${\sf Sample}_{s,m}$ samples the used data set ($d$ or $d'$): 
We say a round has $k$ differentiating data samples if
${\sf Sample}_{s,m}$ sampled a subset of indices which contains exactly $k$ indices of  differentiating data samples from $(d\setminus d')\cup (d'\setminus d)$. 
The adversary knows how ${\sf Sample}_{s,m}$ operates and can derive a joint probability distribution $\mathbb{P}$ of the number of differentiating data samples for each round within the sequence of rounds that define the series of epochs during which updates are computed. 
We consider two types of strong adversaries in our proofs when bounding trade-off functions:

\noindent
\textbf{Adversary ${\cal A}_0$:}
Adversary ${\cal A}_0$, used in prior work\footnote{Prior work provides DP analysis for DP-SGD with individual clipping characterized by $s=1$ in the notation of this paper.}, does not know the exact instance drawn from $\mathbb{P}$ but  is, in the DP proof, given the ability to realize for each round the trade-off function $f_k(\alpha)$ that corresponds to hypothesis testing between  ${\cal M}\circ {\tt Sample}_{s,m}(d)$ and  ${\cal M}\circ {\tt Sample}_{s,m}(d')$ if ${\tt Sample}_{s,m}$ has selected $k$ differentiating samples in that round. 
Adversary ${\cal A}_0$ in the DP analysis that characterizes $f_k(\alpha)$ is given  knowledge about the mapping from indices to values in $d$ or $d'$. Here (as discussed in Appendix \ref{sec:GDP}), the mapping from indices to values in $d\cap d'$ is the same for the mapping from indices to values in $d$ and the mapping from indices to values in $d'$. 
Furthermore, the adversary  
can replay how ${\sf Sample}_{s,m}$  samples each subset $S_b$ of $sm$ indices from\footnote{By assuming $N=N'$ in the DP analysis, knowledge of how ${\sf Sample}_{s,m}$ samples a subset of indices  cannot be used to differentiate the hypotheses of $d$ versus $d'$ based on their sizes (since the index set corresponding to $d$ is exactly the same as the index set corresponding to $d'$).} $\{1,\ldots, N=N'\}$, and it knows all the randomness used by ${\cal M}$ before ${\cal M}$ adds Gaussian noise for differential privacy (this includes when and how the interrupt service routine overwrites the local model). This strong adversary represents a worst-case scenario for the `defender' when analyzing the differential privacy of a single round. 
For DP-SGD (with $s=1$) this analysis for neighboring data sets 
leads to the argument of Section \ref{sec:sample} where with probability $p$ (i.e., $k=1$) the adversary can achieve trade-off function $f(\alpha)$ and with probability $1-p$ (i.e., $k=0$) can achieve trade-off function $1-\alpha$ leading ultimately to operator $C_p$. This in turn leads to the trade-off function  $C_{m/N}(G_{\sigma^{-1}})^{\otimes T}$ with $p=m/N$,
which is {\em tight for adversary ${\cal A}_0$}. 
We notice that adversary ${\cal A}_0$ is used in DP analysis of current literature including the moment accountant method of \cite{abadi2016deep} for analysing $(\epsilon,\delta)$-DP and analysis of divergence based DP measures.

In the DP analysis  adversary ${\cal A}_0$ is given knowledge about the number $k$ of differentiating samples when analysing a single round. That is, it is given an instance of $\mathbb{P}$ projected on a single round. We notice that in expectation the sensitivity (see Section \ref{sec:sens}) of a single round as observed by adversary ${\cal A}_0$ for neighboring data sets is equal to $(1-p)\cdot 0 + p\cdot 2C=(m/N)\cdot 2C$ and this gives rise to an `expected' trade-off function $G_{1/(\sigma N/m)}$. Composition over $T=c^2 (N/m)^2$ rounds gives $G_{1/\sigma}$. This leads us to believe that $C_{m/N}(G_{\sigma^{-1}})^{\otimes T}$ converges to $G_{c\cdot h(\sigma)}$ for constant $m$ and $T=c^2 (N/m)^2 \rightarrow \infty$ (or, equivalently, $\sqrt{T}\cdot m/N =c$ with $T\rightarrow \infty$ and $N\rightarrow \infty$) where $h(\sigma)$ is some function that only depends on $\sigma$. This intuition is confirmed by Corollary 5.4 in \cite{dong2021gaussian}, which characterizes
$$h(\sigma) = \sqrt{2(e^{\sigma^{-2}}\Phi(3\sigma^{-1}/2)+3\Phi(-\sigma^{-1}/2)-2)}
= \sigma^{-1} \cdot \sqrt{1-\frac{5}{8} \frac{\sigma^{-1}}{\sqrt{2\pi}}+O(\sigma^{-2})}.
$$

\noindent
\textbf{Adversary ${\cal A}_1$:}
We define the second type of adversary ${\cal A}_1$  as one who has knowledge about a full instance of $\mathbb{P}$, not just a projection of $\mathbb{P}$ on a single round as for adversary ${\cal A}_0$. This allows a DP analysis that into some extent computes a convex combination of trade-off functions that each characterize all the rounds together as described by an instance of $\mathbb{P}$. This gives adversary ${\cal A}_1$, which we  use in the DP proof in Appendix \ref{app:adversary},  more information and the resulting DP guarantees should be weaker compared to the analysis\footnote{Currently, this has only been done  for DP-SGD since we can only analyse the general algorithmic framework for ${\cal A}_1$.} based on adversary ${\cal A}_0$ (because  adversary ${\cal A}_1$ considers a more worst-case leakage scenario). 

Let us also consider DP-SGD with $s=1$. Then,
since each epoch has $N/m$ rounds and since $m/N$ is equal to the probability that ${\tt Sample}_{s,m}$ selects a differentiating sample in a round when considering neighboring data sets, we have that a single epoch of $N/m$ rounds has in expectation exactly one round with one differentiating sample while all other rounds only use non-differentiating samples. This is a composition of $G_{\sigma^{-1}}$ with $N/m-1$ times $G_0$. We have that the trade-off function for $T=c^2 \cdot (N/m)$ rounds has in expectation a trade-off function $G_{c/\sigma}$. This is confirmed by the $G_{\sqrt{E}/\sigma}$-DP type guarantees in Table \ref{tab:DP} where the total number of gradient computations is equal to $c^2 \cdot (N/m)$ times $m$, which is equal to $E=c^2$ epochs of size $N$. This shows 
$G_{c\cdot h(\sigma)}$ for $T=c^2\cdot (N/m)$
rounds with $h(\sigma)=\sigma^{-1}$. Compared to  adversary ${\cal A}_0$ this is indeed a weaker statement (weaker DP guarantee):

For ${\cal A}_0$ we may choose $T=E\cdot N/m$ (since $NE=mT$) for a constant $E$ and $m$ (both relatively small like 50 or in the 100s in practice) and have $T\rightarrow \infty$ for $N\rightarrow \infty$. This means that for rather large $N$ (which often happens in practice) we may conclude that DP-SGD is approximately\footnote{If $N$ and $T$ get larger, then on one hand more rounds means more updates and thus more moments for privacy leakage, but on the other hand a larger $N$ means that the subsampling will amplify the privacy guarantee more. It is therefore unclear how, for a small constant $c$, convergence to $G_{c\cdot h(\sigma)}$ takes place. We do not know when $N$ is large enough to make this a close approximation. E.g., if we translate the $G_{c\cdot h(\sigma)}$-DP guarantee into $(\epsilon,\delta)$-DP, then it seems that this may lead to a  rather strong statement which we  may not yet have confirmed in practice (with a precise DP accountant based on the $f$-DP framework). We leave this question of how DP guarantees for ${\cal A}_0$ and ${\cal A}_1$ exactly compare  for future work.} $G_{c\cdot h(\sigma)}$-DP for quite small $c=\sqrt{T}\cdot m/N = \sqrt{E\cdot m/N}$ and $h(\sigma)$ slightly smaller than $1/\sigma$. This is closer to the ideal diagonal trade-off function $\alpha \rightarrow 1-\alpha$ when compared to $\approx G_{\sqrt{E}/\sigma}$-DP for DP-SGD in the ${\cal A}_1$ adversarial model. We stress that the reason for using ${\cal A}_1$ is that it allows us to prove DP guarantees for our general algorithmic framework for the first time and allows us to prove the new $\sqrt{g}$ dependency for group privacy.

\noindent
\textbf{Concluding Remarks:}
We are the {\em first} to provide a DP analysis of the much wider class of learning algorithms covered by the general algorithmic framework of Algorithm \ref{alg:genDPSGD}.
Our main insight for being able to prove a DP guarantee is the introduction of a {\em stronger} adversary, denoted by ${\cal A}_1$ leading to Table \ref{tab:DP}, compared to the adversarial model, denoted by ${\cal A}_0$ which is used in  DP proofs in current literature and leads to Table \ref{tab:DPSGD} which summarizes the state of the art $f$-DP analysis of DP-SGD for ${\cal A}_0$. In both ${\cal A}_0$ and ${\cal A}_1$ the adversary knows how ${\sf Sample}_{m}$ operates and can derive a joint probability distribution $\mathbb{P}$ of the number of differentiating data samples
for each round within the sequence of rounds that define the series of epochs during which updates are computed. DP analysis based on ${\cal A}_0$ considers each round separately and analysis of a single round assumes a strong adversary who knows the instance drawn from the projection of $\mathbb{P}$ on that round (analysis of the whole sequence of rounds follows from composition of the derived DP guarantees of single rounds). Our DP analysis based on ${\cal A}_1$ deals with the whole sequence of rounds at once and the resulting analysis assumes a  stronger adversary who knows the instance drawn from $\mathbb{P}$. 


\begin{table*}[t]
    \centering
{\small 
    \begin{tabular}{|c|c|c|c|c|}
    \hline 
    \parbox[t]{2mm}{\rotatebox[origin=c]{90}{\hspace{1mm} ind. clipping (\ref{indclip}) \hspace{1mm}}} &
    \parbox[t]{2mm}{\rotatebox[origin=c]{90}{\hspace{1mm} subsampling \hspace{1mm}}} &
    \parbox[t]{2mm}{\rotatebox[origin=c]{90}{\hspace{1mm} group privacy \hspace{1mm}}} &
    $h$-DP guarantee 
     & 
    \parbox[t]{2mm}{\rotatebox[origin=c]{90}{\hspace{1mm} adversarial model \hspace{1mm}}}  \\
    \hline
    \hline
         \vspace{-3mm} &      &  & &  \\
    x &   x &  $g=1$ & $h=C_{m/N}(G_{1/\sigma})^{\otimes (N/m)\cdot E}$ from \cite{dong2021gaussian},
    & ${\cal A}_0$ \\
    &&& the {\em original DP-SGD setting} \cite{abadi2016deep} & \\
         \vspace{-3mm} &      &  & &  \\
    x &   x & $g\geq 1$ & $h=1-(1-f)^{\circ g}$ with 
    & ${\cal A}_0$ \\
    &&& $f= C_{m/N}(G_{1/\sigma})^{\otimes (N/m)\cdot E}$, see \cite{dong2021gaussian} & \\
    \hline
    \hline
        \vspace{-3mm} &      &  & &  \\
    x &   x &  $g\geq 1$ & $h$ is in the range  $\approx \left[G_{\sqrt{(e+\gamma) \min\{m,g\} g E}/\sigma}, G_0\right]$, & ${\cal A}_1$ \\
    &&& approximation is for small $e^{-\gamma g E}$, see Appendix \ref{sec:subsampling}  & \\

    \hline
    \end{tabular}
}
    \caption{
    Trade-off functions $h$ for the mechanism ${\cal M}$ defined by Algorithm \ref{alg:genDPSGD} in the ${\cal A}_0$ or ${\cal A}_1$ adversarial model for all pairs of data sets $d$ and $d'$ 
with $\max \{ |d\setminus d'|, |d'\setminus d|\}\leq g$; 
$h$ cannot be improved beyond the reported ranges closer towards function $G_0(\alpha)=1-\alpha$ (which represents random guessing of the hypothesis $d$ or $d'$, hence, no privacy leakage). An approximation for a small quantity means that if the quantity tends to $0$, then the approximation becomes tight.} 
    \label{tab:DPSGD}
\end{table*}

The stronger ${\cal A}_1$ allows us to use more "structure" in our DP proofs.
We introduce a probability distribution $q_E(c)$ induced by the sampling procedure ${\sf Sample}_{s,m}$. Together with the knowledge of an instance of $\mathbb{P}$ given to our stronger adversary in our DP analysis, 
we prove a new $f$-DP guarantee  where $f$ is related to a mix of Gaussian trade-off functions $G_{c/\sigma}$ according to distribution $q_E(c)$.


Since ${\cal A}_1$ is a stronger adversary (more capabilities) compared to ${\cal A}_0$,  DP guarantees that can be proven under ${\cal A}_1$ also hold for ${\cal A}_0$;  we expect that if in future work one is able  to prove DP guarantees by assuming the weaker ${\cal A}_0$ for the general framework, then the resulting guarantees will be stronger.
We notice that  ${\cal A}_1$ and also ${\cal A}_0$ are not realistic adversaries, instead they are theoretical constructs that allow us to prove DP guarantees. 
A practical reconstruction or membership attack generally implements an adversary who is much weaker, i.e., has less capabilities, compared to either ${\cal A}_0$ or ${\cal A}_1$. It remains an open problem to mathematically define a weaker practical adversary for which a DP proof can be formulated that yields  stronger DP guarantees  for the algorithms in our general framework. 

\textcolor{blue}{{\bf Additional Remarks:} First, we notice that the state-of-the-art $f$-DP analysis not only holds for individual clipping but also for the generalized batch clipping discussed in this paper. This is because its analysis does not depend on what $f$ computes, except that after clipping its sensitivity is bounded to $2C$. Hence, the individually clipped gradient computation can represent a clipped `batch-gradient' computation. }

\textcolor{blue}{Second, for groups of size $g$, we have in $f$-DP the trade-off function $$\approx G_{\sqrt{Em/N}\cdot h(\sigma)\cdot g}= G_{c\cdot h(\sigma)\cdot g}$$ for  $E=c^2N/m$ (or equivalently $T=c^2(N/m)^2$) with large $N$.
Our analysis shows $$G_{\sqrt{(e+\gamma)mgE}/\sigma}$$ for $g\geq m$. Notice that $h(\sigma)\approx 1/\sigma$ for large $\sigma$. Therefore, only if $\sqrt{Em/N}g>\sqrt{(e+\gamma)mgE}$, then our analysis in the stronger adversarial model provides an improved DP guarantee. But this requires, $g>(e+\gamma)N$ while $g$ is at most $N$. We conclude that the $\sqrt{g}$ dependency does not carry over into the weaker (current state-of-the-art) adversary ${\cal A}_0$.}
\section{Experiments}
\label{sec:07experiments}




\begin{figure*}[th!]
\centering
\subfigure[CIFAR10,$SS$,$IC$]{\includegraphics[width=0.32\textwidth]{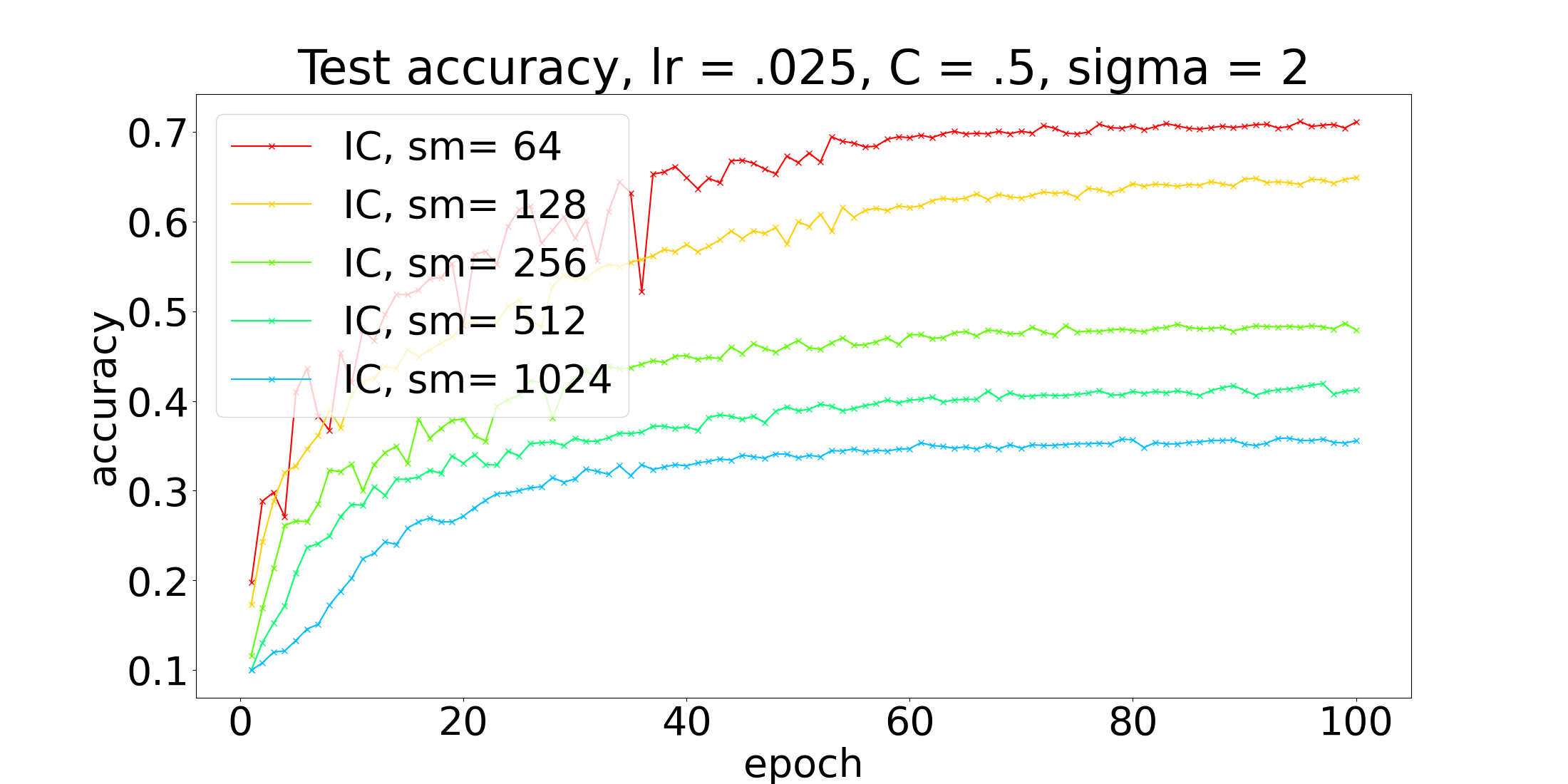}}
\subfigure[CIFAR10,$SS$,$BC$]{\includegraphics[width=0.32\textwidth]{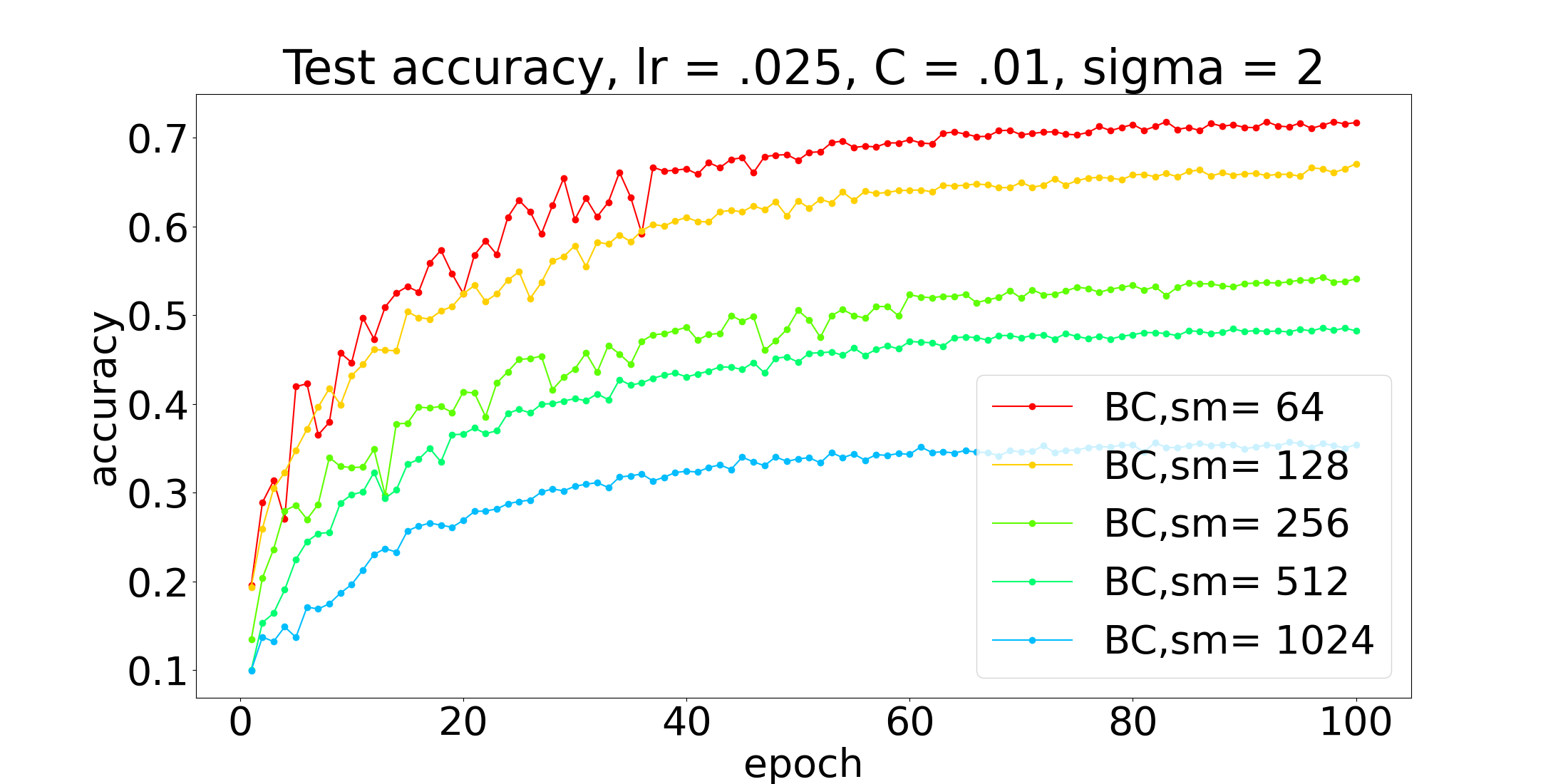}}
\subfigure[CIFAR10,$SS$,$MC$]{\includegraphics[width=0.32\textwidth]{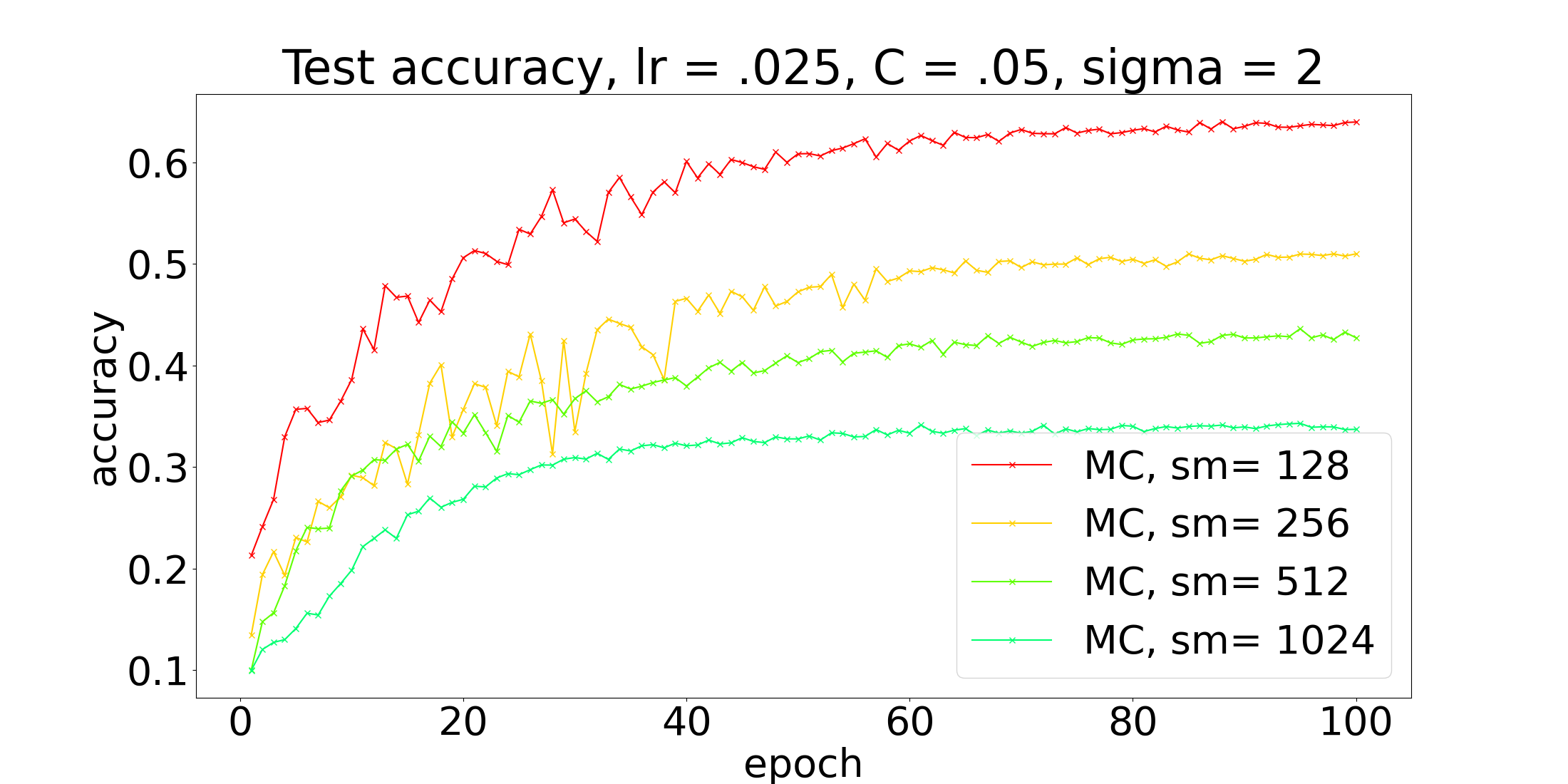}}\\
\subfigure[CIFAR10,$SH$,$IC$]{ \includegraphics[width=0.32\textwidth]{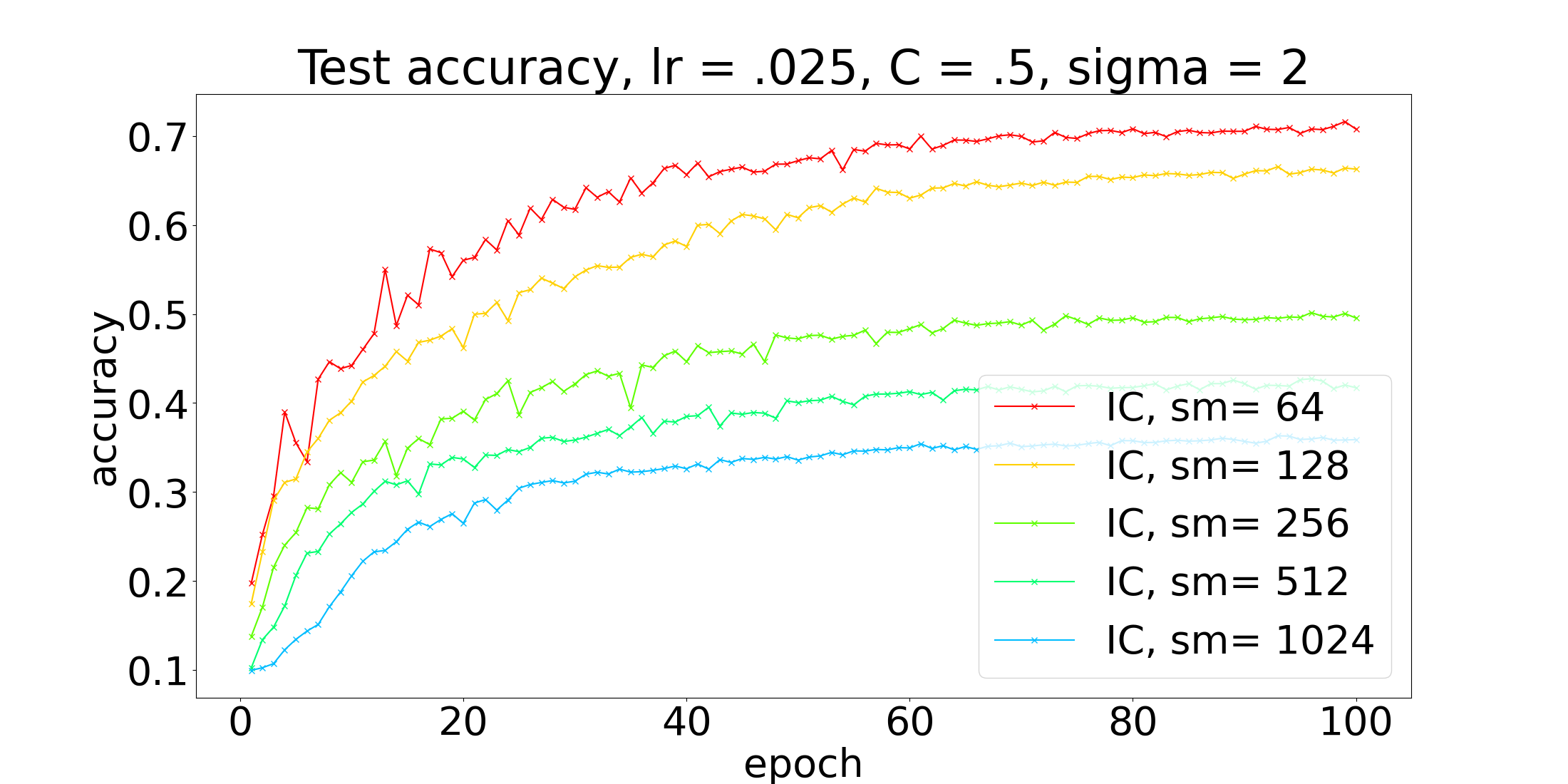}}
\subfigure[CIFAR10,$SH$,$BC$]{\includegraphics[width=0.32\textwidth]{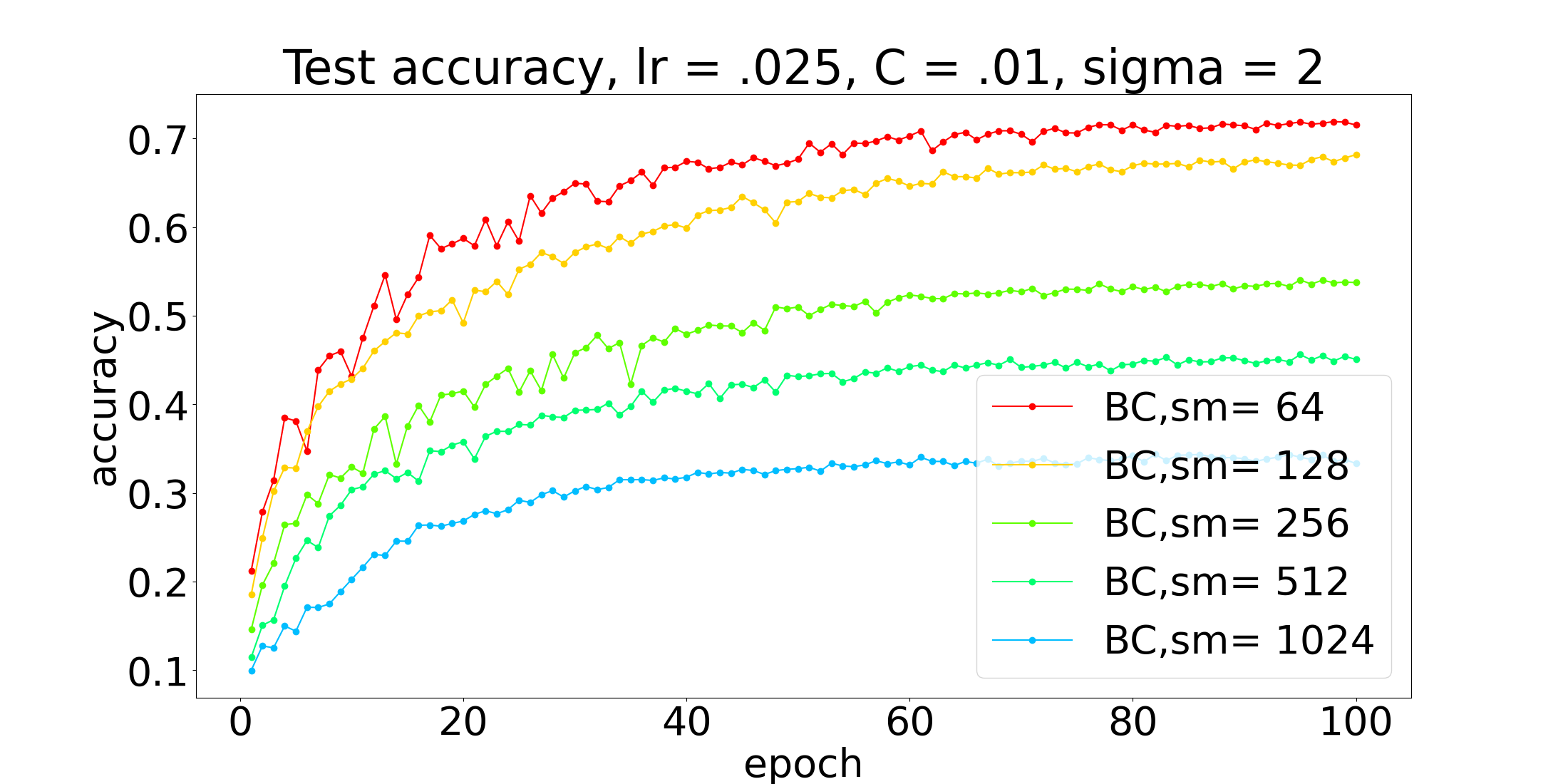}}
\subfigure[CIFAR10,$SH$,$MC$]{\includegraphics[width=0.32\textwidth]{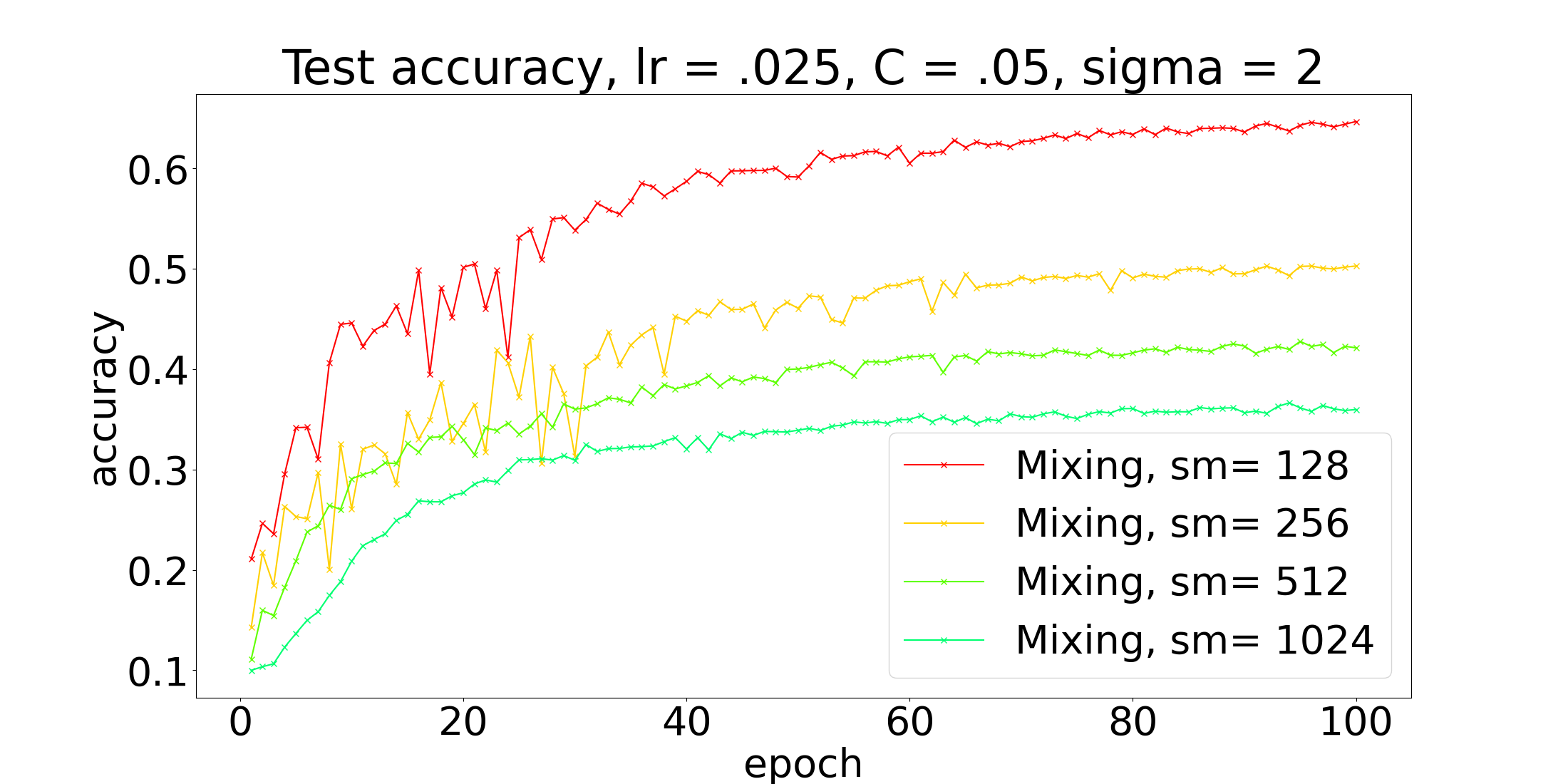}}\\
\subfigure[MNIST,$SS$,$IC$]{\includegraphics[width=0.32\textwidth]{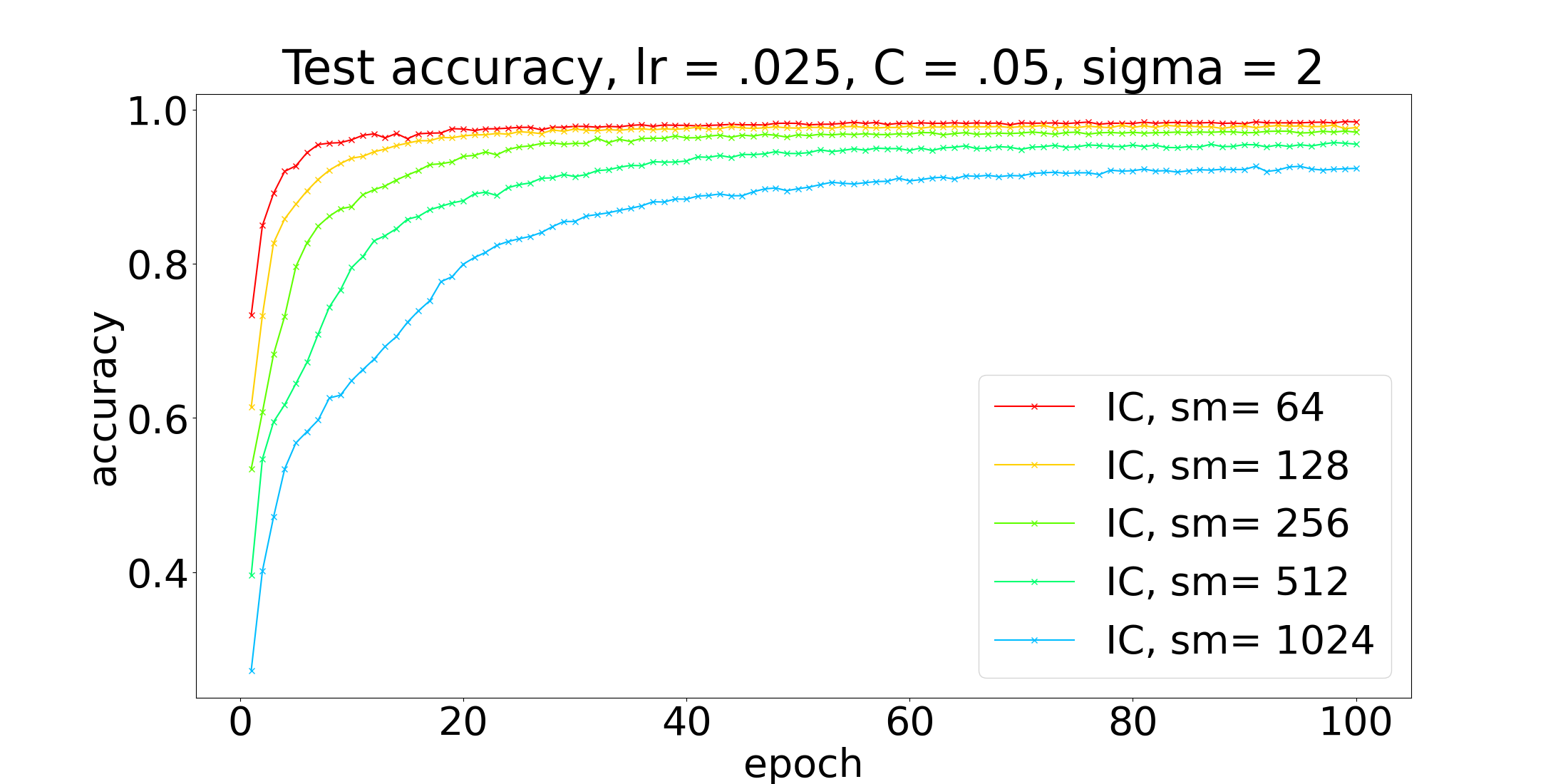}}
\subfigure[MNIST,$SS$,$BC$]{\includegraphics[width=0.32\textwidth]{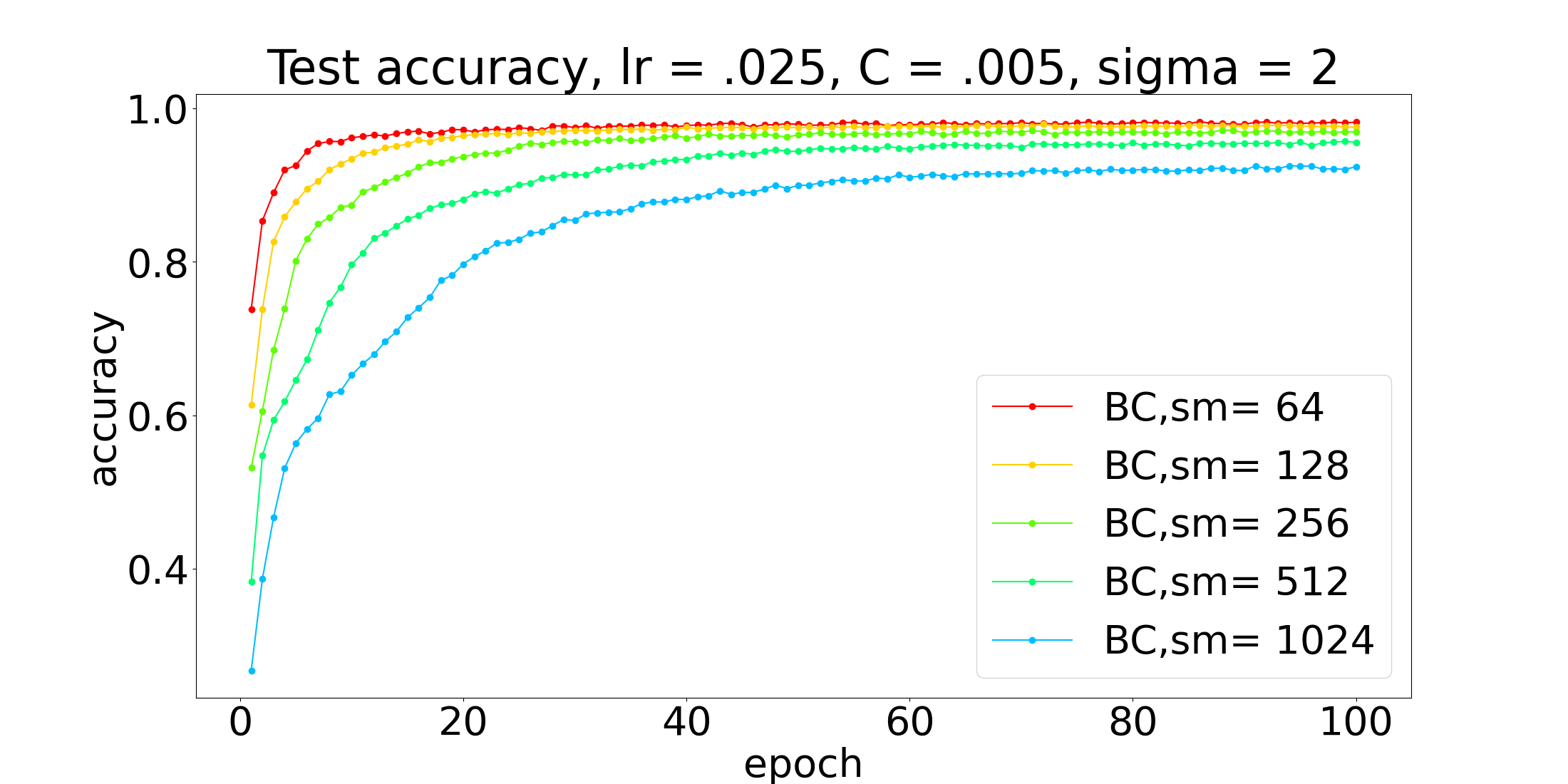}}
\subfigure[MNIST,$SS$,$MC$]{\includegraphics[width=0.32\textwidth]{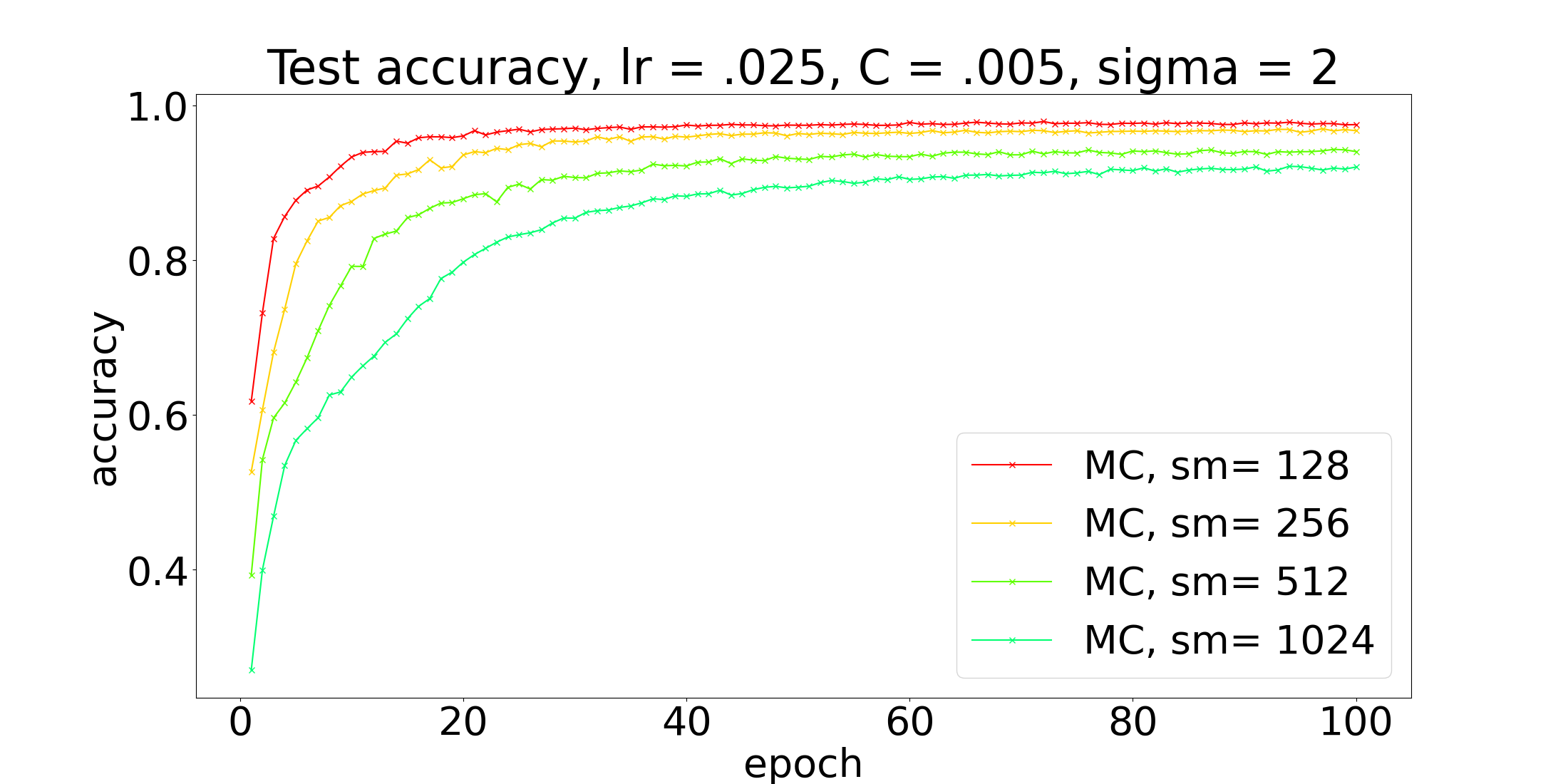}}\\
\subfigure[MNIST,$SH$,$IC$]{\includegraphics[width=0.32\textwidth]{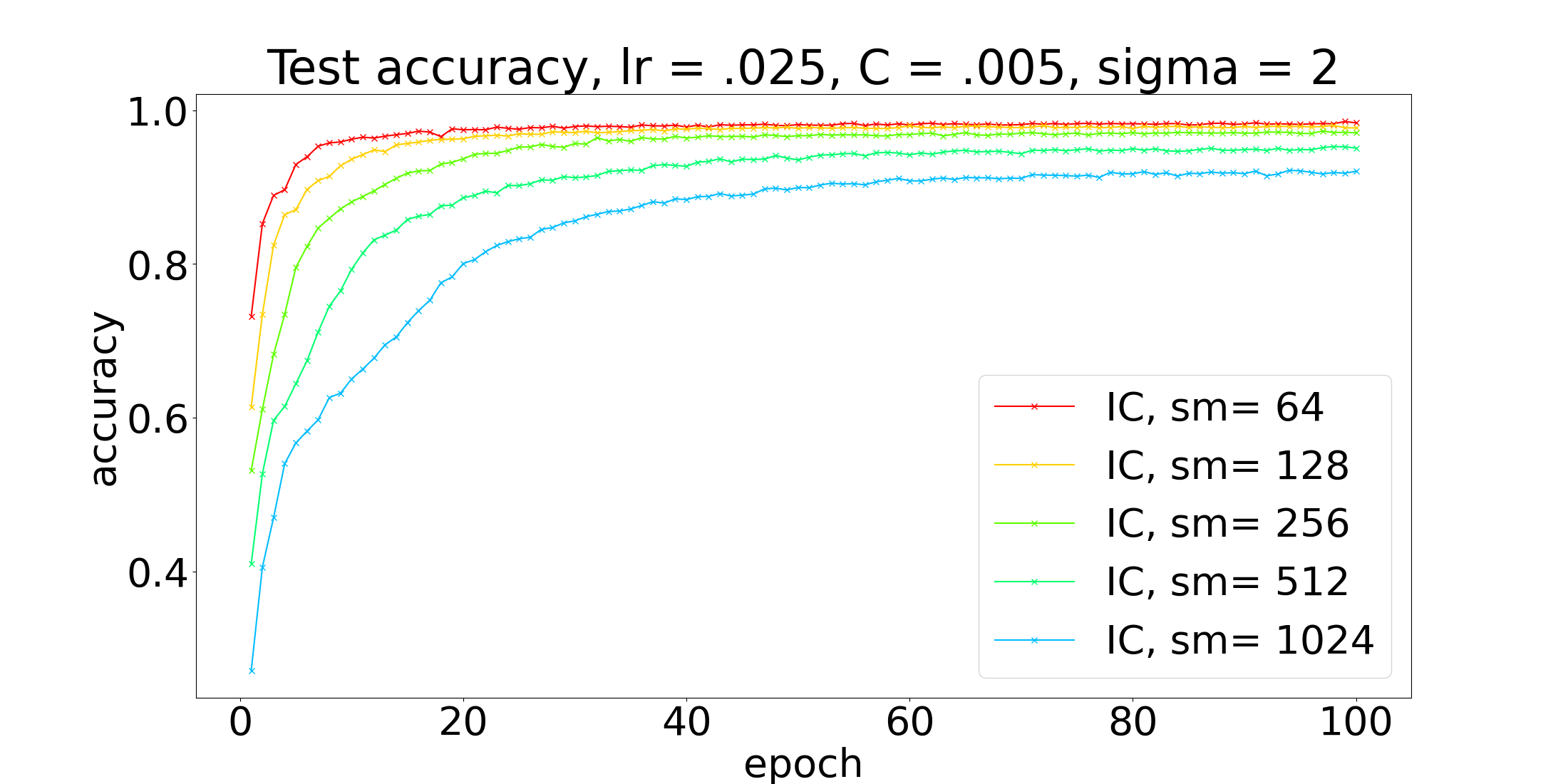}}
\subfigure[MNIST,$SH$,$BC$]{\includegraphics[width=0.32\textwidth]{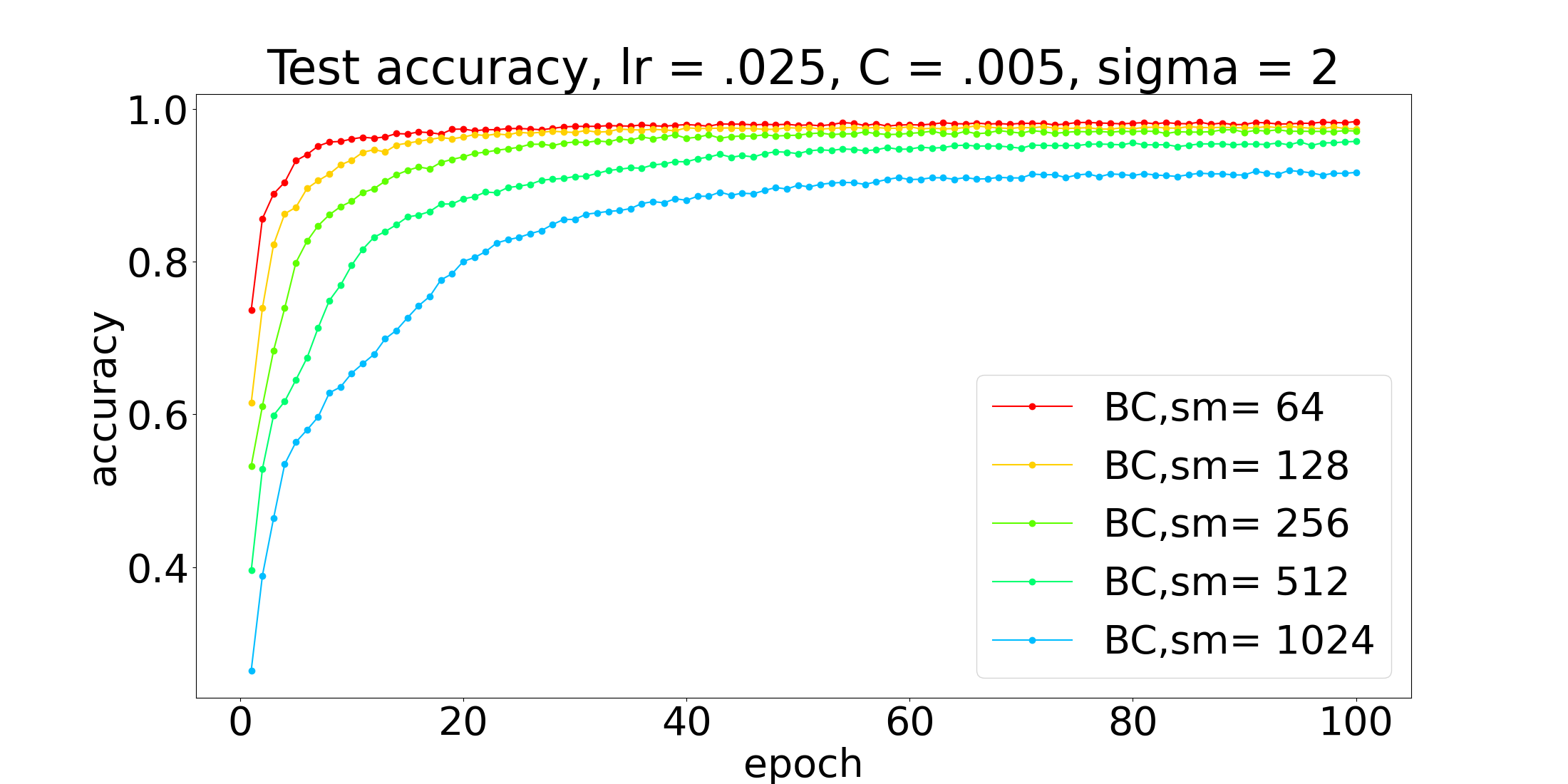}}
\subfigure[MNIST,$SH$,$MC$]{\includegraphics[width=0.32\textwidth]{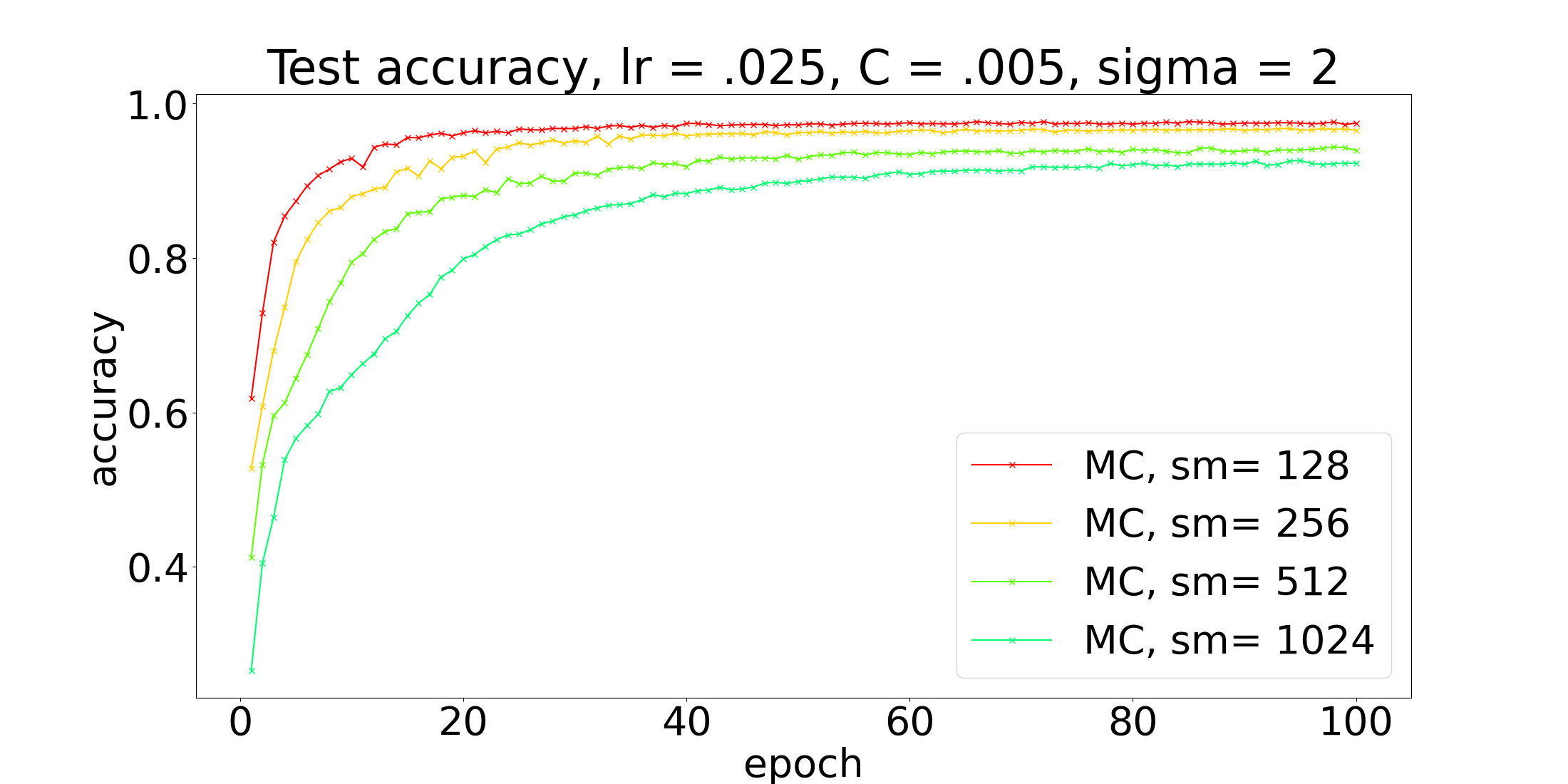}}
\caption{CIFAR10 and MNIST testing accuracy for batch size $|S_b|= sm= 64,128,256,512,1028$ where $SS$ and $SH$ denote SubSampling and SHuffling, and where $IC$, $BC$ and $MC$ denote 
Individual Clipping ($s=1$), Batch Clipping ($m=1$) and Mixed Clipping, respectively. For MC, we set $s=64$ and $m=2,4,8$ and $16$. The used learning rate (step size) is $\mu=0.025$ (denoted by 'lr'), the used noise is given by $\sigma=2$ (denoted by 'sigma'), and the used clipping constant $C$ is fine-tuned for each separate setting. }
\label{fig:CIFAR10Accuracy}
\end{figure*}

In this section we provide a study on extending the mini-batch SGD approach of DP-SGD to include batch clipping as is allowed by the DP analysis of our general framework. We demonstrate that  Algorithm~\ref{alg:genDPSGD} can produce high accuracy models for different setups -- in our experiments, batch clipping yields accuracies comparable to DP-SGD with individual clipping, see Table \ref{tab:main_result}. 

\begin{table}[ht!]
\centering
\scalebox{0.8}{
\begin{tabular}{|c|c|c|c|}
\hline
 Sampling & Clipping  & CIFAR-10 & MNIST \\ \hline \hline
 $SS$& $IC$  & $71.09\% / 68.94\%$ & $98.41\% / 98.31\%$\\ \hline 
 $SS$& $BC$  & $71.7\% / 68.94\%$ & $98.21\% / 98.31\%$ \\ \hline 
 $SS$& $MC$ & $63.97\% / 68.94\%$  &    $97.5\% / 98.31\%$\\ \hline \hline 
 $SH$& $IC$ &  $70.77\% / 70.14\%$ & $98.38\% / 98.4\%$\\ \hline 
 $SH$& $BC$ & $71.52\% / 70.14\%$ & $98.33\% / 98.4\%$\\ \hline 
 $SH$& $MC$ & $64.68\% / 70.14\%$ & $96.97\% / 98.4\%$\\ \hline 
\end{tabular}
}
\caption{The best results among the different setups of Figure \ref{fig:CIFAR10Accuracy} for CIFAR-10 and MNIST with $\sigma=2$ and learning rate $\eta = 0.025$ compared to mini-batch SGD without DP.}
\label{tab:main_result}
\end{table}

Table \ref{tab:main_result} with Figure \ref{fig:CIFAR10Accuracy} depict accuracies for CIFAR-10~\cite{Krizhevskycifar10} and MNIST~\cite{lecun-mnisthandwrittendigit-2010} where we compute updates
\begin{equation}
U = \sum_{h=1}^m [\frac{1}{s} \sum_{i\in S_{b,h}} \nabla f(w,\xi_{i})]_C \label{eq:gen}
\end{equation}
with $s=1$ for Individual Clipping (IC) (this is DP-SGD), $m=1$ for pure Batch Clipping (BC), and values for $s$ and $m$ in between these two extremes denoted by Mixed Clipping (MC). We implement mini-batch SGD and therefore we use the same $w$ in our gradient computations in (\ref{eq:gen}).\footnote{Notice that division by $m$ is taken care of in lines 23 and 24 of Algorithm \ref{alg:genDPSGD} and division by $s$ is done in (\ref{eq:gen}) itself.} The initial learning rate (step size) is $\mu=0.025$ 
and, for each new epoch, the step size is decreased by a factor $\gamma=0.9$ if there is a decrease in testing accuracy between the last two consecutive epochs.
%
%
In our settings we use the same $\sigma=2$ and compare experiments for each benchmark for the same  $E=100$ number of epochs gradient computations.

For each different setup, we report the best result among our grid-search in Table~\ref{tab:main_result}. For the image classification model of MNIST, we use LeNet-5 \cite{Lecun1998lenet5} which yields $98.31\%$ and $98.38\%$ testing accuracy for mini-batch SGD without DP for subsampling and shuffling. For our demonstration of being able to achieve similar high accuracy models, we use the lightweight Convolutional Neural Network (CNN) in Table \ref{tab:CIFAR10convNN} for CIFAR10. This has both fast training and good accuracy; $68.94\%$ and $70.14\%$ for mini-batch SGD without DP for subsampling and shuffling. 

\begin{table}[h]
\label{tab:CIFAR10convNN}
\centering
\scalebox{0.8}{
\begin{tabular}{|c|c|c|c|c|}
\hline
 Operation Layer & \#Filters  & Kernel size & Stride & Padding \\ \hline 
 \parbox[c]{3cm}{\vspace{1mm} \centering $Conv2D$+$ReLu$ \vspace{1mm}}& $32$ & $3 \times 3$ & $1 \times 1$ & $1 \times 1$\\ \hline 
 $AvgPool2d$& $1$  & $2 \times 2$ & $2 \times 2$ & $-$\\ \hline 
\parbox[c]{3cm}{\vspace{1mm} \centering $Conv2D$+$ReLu$\vspace{1mm}}& $64$ & $3 \times 3$ & $1 \times 1$ & $1 \times 1$\\ \hline 
 $AvgPool2d$& $1$  & $2 \times 2$ & $2 \times 2$ & $-$\\ \hline 
 \parbox[c]{3cm}{\vspace{1mm} \centering $Conv2D$+$ReLu$\vspace{1mm} }& $64$ & $3 \times 3$ & $1 \times 1$ & $1 \times 1$\\ \hline 
 $AvgPool2d$& $1$  & $2 \times 2$ & $2 \times 2$ & $-$\\ \hline 
 \parbox[c]{3cm}{\vspace{1mm} \centering $Conv2D$+$ReLu$ \vspace{1mm}}& $128$ & $3 \times 3$ & $1 \times 1$ & $1 \times 1$\\ \hline 
 $AdaptiveAvgPool2d$& $-$  & $-$ & $-$ & $-$\\ \hline 
 $Softmax$& $-$  & $-$ & $-$ & $-$\\ \hline 
\end{tabular}
}
\caption{CNN architecture for CIFAR-10 dataset}
\end{table}

This section reports an initial study and future work is needed to explore the more general algorithmic freedom in (algorithm ${\cal A}$ of) our framework for improving accuracy in specific application scenarios.

\end{document}